\documentclass[10pt,twocolumn,letterpaper]{article}
\usepackage{cvpr}

\usepackage{microtype}

\usepackage{bm}
\usepackage{fontawesome}
\usepackage{bbding}

\usepackage{tabularx}
\usepackage{longtable}
\usepackage{multirow}
\usepackage{makecell}
\usepackage{colortbl}

\usepackage{float}
\usepackage{placeins}

\usepackage{dashrule}

\usepackage{tcolorbox}
\tcbuselibrary{breakable}
\newcommand{\VarSty}[1]{\textnormal{\ttfamily\color{blue!90!black}#1}\unskip}

\usepackage{titletoc}

\definecolor{cvprblue}{rgb}{0.21,0.49,0.74}
\usepackage[pagebackref,breaklinks,colorlinks,allcolors=cvprblue]{hyperref}
\def\confName{CVPR}
\def\confYear{2027}

\title{ManiVid: Unified and Explainable Manipulation Analysis of Videos via MLLMs}

\author{%
Hengrui Kang\textsuperscript{\rm 1,2*},
Zhonghao Yan\textsuperscript{\rm 4*},
Yuxuan Yang\textsuperscript{\rm 4*},
Ruoyan Jing\textsuperscript{\rm 4},
Yuncheng Guo\textsuperscript{\rm 5},
Hao Chen\textsuperscript{\rm 6}, \\
Kongming Liang\textsuperscript{\rm 4},
Zhanyu Ma\textsuperscript{\rm 4},
Conghui He\textsuperscript{\rm 2},
Weijia Li\textsuperscript{\rm 3\textdagger}
\\[1mm]
\normalsize \textsuperscript{\rm 1} Shanghai Jiao Tong University \quad
\normalsize \textsuperscript{\rm 2} Shanghai Artificial Intelligence Laboratory \quad 
\normalsize \textsuperscript{\rm 3} Tsinghua University \\
\normalsize \textsuperscript{\rm 4} Beijing University of Posts and Telecommunications \quad 
\normalsize \textsuperscript{\rm 5} Fudan University \quad 
\normalsize \textsuperscript{\rm 6} The Chinese University of Hong Kong
}

\makeatletter
\patchcmd{\@maketitle}
  {\vskip .375in}
  {\vspace*{-.10in}}
  {}
  {\PackageWarningNoLine{main}{CVPR title patch failed}}
\makeatother

\begin{document}
\makeatletter
\twocolumn[{%
  \@twocolumnfalse
  \maketitle
  \vspace{-6.5mm}
\begin{minipage}{\textwidth}
    \centering
    \includegraphics[width=\linewidth]{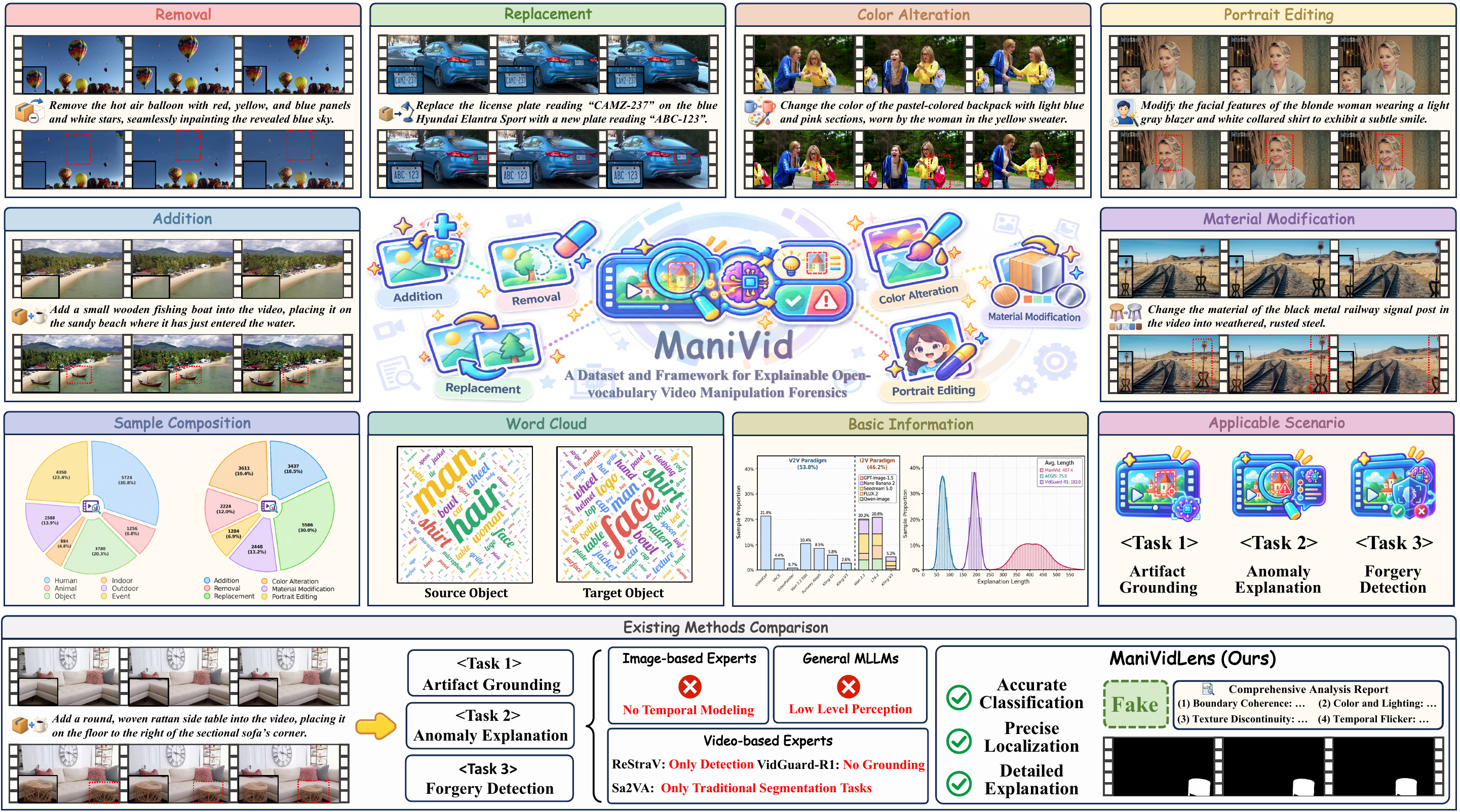}
    \vspace{-6mm}
    \captionsetup{hypcap=false}
    \captionof{figure}{\textbf{Overview of ManiVid.} ManiVid unifies forgery detection, artifact grounding, and anomaly explanation. The \textbf{ManiVid-38K} dataset provides paired, open-vocabulary manipulations across six types with authenticity labels, pixel-level masks, and anomaly explanations. \textbf{ManiVidBench} provides 2K videos for evaluation, and \textbf{ManiVidLens} offers an end-to-end solution.}
    \label{fig:teaser}
    \vspace{7pt} 
\end{minipage}

}]
\makeatother
\begin{abstract}
Rapid advances in AI-generated video (AIGV) have increased the risks posed by deceptive video manipulation. Unlike fully synthetic videos, manipulated videos retain most source content and alter only localized regions, making forensic analysis particularly challenging.
Existing video forgery research faces two limitations in data and methodology: (1) high-quality datasets and benchmarks tailored to manipulated videos remain scarce, and (2) multimodal large language models (MLLMs) extend forgery analysis beyond binary classification but struggle to exploit low-level forensic cues and to provide precise pixel-level grounding.
To address both, we introduce \textbf{ManiVid}, a unified forensic analysis task covering forgery detection, artifact grounding, and anomaly explanation for manipulated videos. We construct \textbf{ManiVid-38K}, the first dataset to combine paired, open-vocabulary localized manipulations of general videos with authenticity labels, forgery masks, and anomaly explanations. It comprises about 19K manually verified real--fake video pairs, mostly at 1080P resolution, produced by 15 recent generation models under two paradigms. We sample 1K pairs for \textbf{ManiVidBench}, balanced across six manipulation types and generation models for fair evaluation.
We further propose \textbf{ManiVidLens}, a unified framework for explainable video forgery analysis.
Its \textit{Forensic Evidence Router} supplies shared low-level forensic evidence to both multimodal reasoning and video segmentation, and its \textit{Prompt Distill Module} converts grounding states into semantic and geometric prompts while distilling spatial priors for mask decoding and full-video propagation.
ManiVidLens outperforms the strongest competing methods in artifact grounding (0.407 vs.\ 0.336 mIoU; 0.422 vs.\ 0.348 J\&F) and anomaly explanation (0.593 vs.\ 0.252 ROUGE-L; 0.796 vs.\ 0.710 CSS). 
Its forgery detection remains comparable to dedicated classifiers (0.920 Acc; 0.919 F1).
Together, our dataset and framework enable accurate and interpretable forensic analysis of manipulated videos.
The code, model, and dataset will be released.

\end{abstract}

\section{Introduction}

Recent advances in AI-generated video (AIGV)~\citep{team2025kling, wan2025wan, hacohen2026ltx} have enabled increasingly realistic, temporally coherent, and visually compelling videos through text-to-video (T2V) and image-to-video (I2V) generation.
Their widespread adoption on social media has lowered the barrier to customized content creation, but it has also raised serious concerns such as fraud and online misinformation.

To mitigate these risks, researchers have developed a range of video forgery detection methods.
Early approaches apply image detectors to individual frames and rely primarily on spatial forgery traces~\citep{vahdati2024beyond}.
More recent video detectors jointly model spatial artifacts and temporal inconsistencies to predict authenticity~\citep{chen2026demamba, interno2026ai, zhang2026physics, zheng2025d3}.
Emerging approaches employ multimodal large language models (MLLMs) to complement binary judgments with anomaly explanations~\citep{park2025vidguard}.
Despite this, two critical limitations remain:

\textbf{(\emph{i}) \textit{Limited Attention and Data for Video Manipulation.}}
Unlike fully synthetic videos generated from text or images, manipulated videos preserve most source content and alter only localized regions, which makes their edits subtle and hard to distinguish from authentic content. Yet research predominantly targets fully synthetic videos~\citep{bai2024ai,chen2026demamba}, and studies of video manipulation remain limited to specific scenarios such as face swapping~\citep{dolhansky2020deepfake}.
Building large-scale, high-quality, and diverse datasets for general video manipulation analysis poses three challenges. \textbf{(1) Quality:} edits must follow instructions and preserve unrelated content, requiring careful verification; \textbf{(2) Diversity:} generating diverse instructions that remain consistent with each source video is difficult at scale; \textbf{(3) Scalability:} video editing requires more computation and time than image editing. 
These challenges hinder large-scale benchmarks for general video manipulation that jointly provide paired open-vocabulary manipulations, pixel-level masks, and anomaly explanations.

\textbf{(\emph{ii}) \textit{Insufficient Interpretability in Existing Detection Methods.}}
Most video forgery detectors output only authenticity labels, without localizing manipulations or explaining the evidence. Recent systems use MLLMs for anomaly explanation~\citep{park2025vidguard}, or SAM~\citep{kirillov2023segment} and SAM2~\citep{ravi2025sam} for artifact grounding~\citep{kang2025legion,huang2025sida,xu2024fakeshield}. Across these method families, two limitations remain. \textbf{(1) Objective Coupling:} forgery detection, anomaly explanation, and artifact grounding occupy distinct output spaces, forcing them into one response couples their objectives and can dilute the evidence needed for segmentation. \textbf{(2) Evidence Misalignment:} detectors, MLLMs, and segmentation models specialize in authenticity discrimination, semantic reasoning, and mask decoding, so subtle manipulation cues are not shared across tasks. Existing interfaces further reduce grounding to a single latent state~\citep{yuan2025sa2va} or decoded coordinates~\citep{li2026omni}, without complementary semantic and geometric prompts. These gaps call for task-specific output protocols, shared forensic evidence, and a structured prompt interface for pixel-level video grounding.

To address these limitations, we propose \textbf{ManiVid} to unify forgery detection, artifact grounding, and anomaly explanation for manipulated videos. We construct \textbf{ManiVid-38K}, the first dataset to combine paired, open-vocabulary localized manipulations of general videos with authenticity labels, forgery masks, and detailed anomaly explanations. This high-quality dataset contains about \textbf{19K} real--fake video pairs, mostly at \textbf{1080P} resolution, generated by \textbf{15} models under \textbf{2} generation paradigms. It spans \textbf{6} common content categories and \textbf{6} manipulation tasks with comprehensive annotations. 
Representative manipulation cases and statistical analyses are shown in Figure~\ref{fig:teaser}
.
We sample \textbf{1K} pairs from ManiVid-38K to construct \textbf{ManiVidBench} for evaluation.
As illustrated in Figure~\ref{fig:teaser} (Bottom), we propose \textbf{ManiVidLens}, an end-to-end solution that combines an MLLM with SAM2~\citep{ravi2025sam}. The three tasks share parameters but follow task-specific protocols. Its \textit{Forensic Evidence Router} (FER) routes low-level cues from the frozen AIDE encoder~\citep{yan2025sanity} to both the language and mask streams. Its \textit{Prompt Distill Module} (PDM) maps grounding-token states to dense semantic, sparse geometric, and segmentation prompts, and cosine distillation transfers SAM2's box-encoding prior to the two box-token states. At inference, these prompts drive mask decoding and propagation without human input.

Our main contributions are summarized as follows:

\begin{itemize}[leftmargin=22pt, topsep=0pt, itemsep=1pt, partopsep=1pt, parsep=1pt]

\item We introduce \textbf{ManiVid} with its \textbf{ManiVid-38K} dataset and \textbf{ManiVidBench} benchmark. The dataset is the first to combine paired, open-vocabulary localized manipulations of general videos with authenticity labels, pixel-level masks, and anomaly explanations for every real--fake video pair.

\item We propose \textbf{ManiVidLens}, a unified end-to-end framework for explainable video forgery analysis. It couples an MLLM with SAM2 to perform forgery detection, artifact grounding, and anomaly explanation via task-specific protocols, shared forensic evidence, and structured semantic and geometric prompts.

\item Extensive experiments on \textbf{ManiVidBench} show that \textbf{ManiVidLens} achieves state-of-the-art artifact grounding and anomaly explanation, with forgery detection comparable to dedicated classifiers.

\end{itemize}

\section{Related Work}
\begin{figure*}[t]
    \centering
    \vspace{-1mm}
    \includegraphics[width=\textwidth]{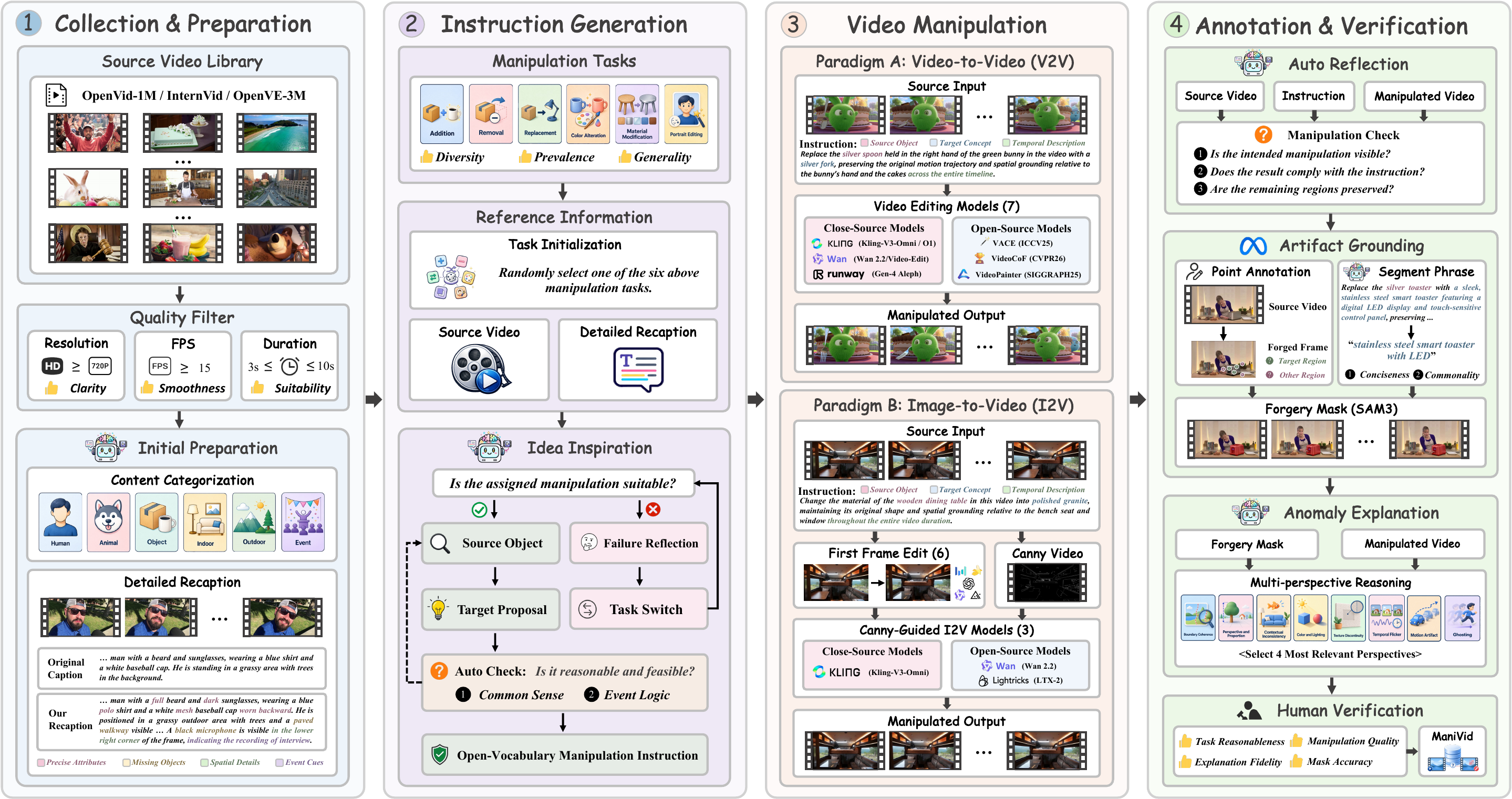}
    \vspace{-6mm}

    \caption{\textbf{Dataset Curation.} ManiVid collects high-quality real videos from three public datasets and uses the MLLM assistant to generate \textbf{\textit{open-vocabulary}}, \textbf{\textit{reasonable}}, and \textbf{\textit{feasible}} manipulation instructions. Videos are then manipulated by diverse models, followed by the auto reflection and human verification. ($\cdot$) denotes the number of generation models used.}
    \vspace{-4mm}
    
    \label{fig:dataset_main}
\end{figure*}

\noindent\textbf{Video Forgery Detection Datasets.}
Existing video forgery datasets cover fully synthetic videos and localized manipulations.
Fully synthetic benchmarks~\citep{bai2024ai,chen2026demamba,ni2026genvidbench,wang2024vidprom,li2025aegis,park2025vidguard} collect T2V or I2V outputs and vary in generator coverage, semantic alignment, and annotation granularity.
Localized manipulation datasets largely focus on facial forgeries~\citep{rossler2018faceforensics,dolhansky2020deepfake}, while VideoSham~\citep{mittal2023video} broadens coverage to general spatial and temporal edits.
Video editing datasets offer more diverse operations~\citep{bian2025videopainter,he2025openve} but primarily serve editing-model development rather than forensic analysis, with no guarantee of source video authenticity.
ManiVid-38K fills this gap with comprehensively annotated pairs of verified authentic videos and their diverse, high-quality open-vocabulary edits.

\noindent\textbf{Video Forgery Detection Methods.}
Frame-wise image detectors struggle with AI-generated videos~\citep{vahdati2024beyond}, motivating video detectors that capture both spatial artifacts and temporal inconsistencies.
Recent approaches model representation trajectories, normalized spatiotemporal gradients, and second-order temporal differences~\citep{interno2026ai,zhang2026physics,zheng2025d3}.
They primarily target fully generated videos and predict authenticity, leaving artifact localization and anomaly explanation outside their output scope.
MLLM-based methods extend image forgery detection with textual explanations~\citep{wen2026spot,zhu2026fakevlm,huang2025so,lin2025seeing}.
VidGuard-R1~\citep{park2025vidguard} brings detection and explanation to videos but does not localize artifacts.
Omni-Fake-R1~\citep{li2026omni} jointly predicts authenticity, localization, and explanations across modalities, with spatial outputs limited to coarse bounding boxes.
Instead, ManiVidLens unifies forgery detection, pixel-level artifact grounding, and anomaly explanation to enable explainable forensic analysis of video manipulation.

\section{ManiVid-38K Dataset}




\begin{table*}[!tbp]
\centering

\caption{\textbf{Comparison with Existing Datasets.} \textbf{V.P.} denotes the availability of semantic-aligned real-fake pairs; \textbf{O.V.I.} denotes the release of open-vocabulary instructions; \textbf{H.V.} denotes human verification. Datasets that cannot guarantee source videos are real are \textbf{\textcolor{Red}{excluded}}.} 
\vspace{-2mm}
\definecolor{mygray}{gray}{.92}
\renewcommand{\arraystretch}{1.0}
\belowrulesep=-0.25pt
\aboverulesep=-0.25pt
\resizebox{1.0\linewidth}{!}{
    \begin{tabular}{c|l|c*{9}{c}c}
    \toprule[1.5pt]
    & \multirow{2}{*}{\textbf{Dataset}} & \multirow{2}{*}{\textbf{Domain}} & \multicolumn{4}{c}{\textbf{Video Generation}} & \multicolumn{5}{c}{\textbf{Comprehensive Annotation}} & \multirow{2}{*}{\textbf{Scale}} \\
    \cmidrule(l{10pt}r{10pt}){4-7} \cmidrule(l{6pt}r{10pt}){8-12}
    & & & \textbf{Task} & \textbf{Model} & \textbf{Paradigm} & \textbf{V.P.} & \textbf{O.V.I} & \textbf{Label} & \textbf{Mask} & \textbf{Explanation} & \textbf{H.V.} & \\
    \hline
    \multirow{5}{*}{\rotatebox[origin=c]{90}{\textit{Synth.}}} & VidProM~\citep{wang2024vidprom} & General & - & 4 & T2V & \textcolor{Red}{\XSolidBrush} & \textcolor{Green}{\CheckmarkBold} & \textcolor{Green}{\CheckmarkBold} & \textcolor{Red}{\XSolidBrush} & \textcolor{Red}{\XSolidBrush} & \textcolor{Red}{\XSolidBrush} & 6.69M \\
    & GenVideo~\citep{chen2026demamba} & General & - & 20 & T2V, I2V & \textcolor{Red}{\XSolidBrush} & \textcolor{Red}{\XSolidBrush} & \textcolor{Green}{\CheckmarkBold} & \textcolor{Red}{\XSolidBrush} & \textcolor{Red}{\XSolidBrush} & \textcolor{Red}{\XSolidBrush} & 2.31M \\
    & AEGIS~\citep{li2025aegis} & General & - & 10 & T2V, I2V & \textcolor{Red}{\XSolidBrush} & \textcolor{Red}{\XSolidBrush} & \textcolor{Green}{\CheckmarkBold} & \textcolor{Red}{\XSolidBrush} & \textcolor{Green}{\CheckmarkBold} & \textcolor{Green}{\CheckmarkBold} & 10470 \\
    & GenVideoBench~\citep{ni2026genvidbench} & General & - & 11 & T2V, I2V & \textcolor{Green}{\CheckmarkBold} & \textcolor{Green}{\CheckmarkBold} & \textcolor{Green}{\CheckmarkBold} & \textcolor{Red}{\XSolidBrush} & \textcolor{Red}{\XSolidBrush} & \textcolor{Red}{\XSolidBrush} & 6.78M \\
    & VidGuard-R1~\citep{park2025vidguard} & General & - & 2 & I2V & \textcolor{Green}{\CheckmarkBold} & \textcolor{Red}{\XSolidBrush} & \textcolor{Green}{\CheckmarkBold} & \textcolor{Red}{\XSolidBrush} & \textcolor{Green}{\CheckmarkBold} & \textcolor{Red}{\XSolidBrush} & 140K \\
    \hline
    \multirow{5}{*}{\rotatebox[origin=c]{90}{\textit{Mani.}}} & FaceForensics++~\citep{rossler2019faceforensics++} & Facial & 2 & 4 & V2V & \textcolor{Green}{\CheckmarkBold} & \textcolor{Red}{\XSolidBrush} & \textcolor{Green}{\CheckmarkBold} & \textcolor{Green}{\CheckmarkBold} & \textcolor{Red}{\XSolidBrush} & \textcolor{Red}{\XSolidBrush} & 5K \\
    & DFDC~\citep{dolhansky2020deepfake} & Facial & 1 & 5 & V2V & \textcolor{Green}{\CheckmarkBold} & \textcolor{Red}{\XSolidBrush} & \textcolor{Green}{\CheckmarkBold} & \textcolor{Red}{\XSolidBrush} & \textcolor{Red}{\XSolidBrush} & \textcolor{Red}{\XSolidBrush} & 128K \\
    & VideoSham~\citep{mittal2023video} & General & 6 & 0 & Manual & \textcolor{Green}{\CheckmarkBold} & \textcolor{Green}{\CheckmarkBold} & \textcolor{Green}{\CheckmarkBold} & \textcolor{Red}{\XSolidBrush} & \textcolor{Red}{\XSolidBrush} & \textcolor{Green}{\CheckmarkBold} & 826 \\
    & VideoFACT~\citep{nguyen2024videofact} & General & 4 & 3 & V2V & \textcolor{Green}{\CheckmarkBold} & \textcolor{Red}{\XSolidBrush} & \textcolor{Green}{\CheckmarkBold} & \textcolor{Green}{\CheckmarkBold} & \textcolor{Red}{\XSolidBrush} & \textcolor{Red}{\XSolidBrush} & 12.38K \\
    & \textbf{ManiVid~(Ours)} & \textbf{General} & \textbf{6} & \textbf{15} & \textbf{V2V, I2V} & \textcolor{Green}{\CheckmarkBold} & \textcolor{Green}{\CheckmarkBold} & \textcolor{Green}{\CheckmarkBold} & \textcolor{Green}{\CheckmarkBold} & \textcolor{Green}{\CheckmarkBold} & \textcolor{Green}{\CheckmarkBold} & \textbf{$\sim$38K} \\
    \bottomrule[1.5pt]
    \end{tabular}
}

\vspace{-4mm}
\label{tab:dataset_comparison}
\end{table*}

\subsection{Motivation and Highlights}
\label{dataset:motivation}
Compared with fully synthetic videos, manipulated videos have received less attention because they are harder to construct, annotate, and verify, so high-quality datasets remain scarce. Existing manipulation datasets suffer from three systematic limitations.
\textbf{(1) Limited Diversity.} Previous works~\citep{rossler2019faceforensics++,dolhansky2020deepfake} focus mainly on face-centric manipulations such as face swapping, leaving diverse open-world content and manipulation scenarios underexplored.
\textbf{(2) Closed-Vocabulary Instructions.} Image manipulation datasets such as SIDA~\citep{huang2025sida} build instructions from predefined source and target vocabularies, which restricts diversity and often yields targets that violate common sense or event logic.
\textbf{(3) Insufficient Annotations.} Most datasets provide only binary authenticity labels and lack fine-grained supervision such as forgery masks and anomaly explanations, which substantially limits interpretability.
To address these issues, we construct \textbf{ManiVid-38K} with three key highlights.
\textbf{(1) Broad Coverage.} It covers six content categories (\emph{e.g.}, human, object) and six manipulation tasks (\emph{e.g.}, addition, removal), spanning diverse open-world scenarios and realistic manipulation operations.
\textbf{(2) Open-Vocabulary Instructions.} Instead of predefined vocabularies, an MLLM-assisted self-reflection pipeline generates open-vocabulary, context-aware instructions that are both diverse and logically consistent.
\textbf{(3) Comprehensive Annotations.} ManiVid-38K provides binary labels, pixel-level forgery masks, and anomaly explanations to facilitate research on explainable video manipulation analysis.
Section~\ref{app:motivation} in the supplementary material discusses this motivation in more detail.
\vspace{-1mm}
\subsection{Curation Pipeline}
\label{dataset:construction}
As shown in Fig.~\ref{fig:dataset_main}, ManiVid-38K is curated in four sequential steps: \textbf{(1) Collection and Preparation}; \textbf{(2) Instruction Generation}; \textbf{(3) Video Manipulation}; \textbf{(4) Annotation and Verification}. All steps use Qwen3-VL-235B-A22B-Instruct~\citep{bai2025qwen3} as the MLLM assistant. Section~\ref{app:curation_pipeline} in the supplementary material provides further dataset curation details.

\noindent \textbf{Collection and Preparation.}
We collect source videos from three public datasets: InternVid~\citep{wang2024internvid}, OpenVid-1M~\citep{nan2025openvid}, and OpenVE-3M~\citep{he2025openve}. Specifically, 
InternVid and OpenVid-1M predate recent high-fidelity video generators (\emph{e.g.}, Sora~\citep{openai2024videoworldsimulators}) and therefore provide relatively reliable authentic sources.
For OpenVE-3M, we manually select source videos free of noticeable AI-generated artifacts.
We then filter out videos with low resolution, low frame rates, or excessive durations to ensure source quality.
Finally, the assistant categorizes and recaptions each video, and we sample videos evenly across all six video content categories.

\noindent \textbf{Instruction Generation.}
For each source video, we first randomly assign one of the six manipulation tasks. Given the video and its detailed recaption, the assistant determines whether the assigned task is suitable for the current scene. 
If so, it selects a source object and proposes a semantically coherent target from its open-world knowledge while respecting common sense and event logic. Otherwise, it tries alternative tasks until it can propose a feasible source--target pair. For example, if a video contains no human subjects, the process will reassign it from \textit{portrait editing} to another task.

\noindent \textbf{Video Manipulation.}
To reflect the rapid evolution of video generation tools, we manipulate videos through \textit{two complementary paradigms}:
\textbf{(1) Direct V2V.} We directly edit source videos with seven recent editing models, including four closed-source models~\citep{team2025kling, kling_video_3_omni, runwayaleph, wavespeed_wan22_video_edit} and three open-source models~\citep{yang2026videocof, jiang2025vace, bian2025videopainter}.
\textbf{(2) Indirect I2V.} 
We formulate this paradigm as image editing followed by controllable I2V generation. Specifically, we first edit the initial frame according to the instruction using six image editing models, including three closed-source models~\citep{google_nano_banana_2, bytedance_seedream5, openai2025gptimage15} and three open-source models~\citep{blackforestlabs_flux2_dev, blackforestlabs_flux2_klein_9b, wu2025qwen}. We then extract Canny edges from the source video and condition generation on both the edited first frame and the edge video. Three I2V models support Canny guidance: the closed-source Kling-V3-Omni~\citep{kling_video_3_omni} and two open-source models~\citep{wan2025wan, hacohen2026ltx}.


\noindent \textbf{Annotation and Verification.}
We establish a rigorous annotation and verification pipeline to ensure data quality.
First, we prompt the assistant to reflect on manipulation success, instruction adherence, and the preservation of unedited regions, and we retain only qualified samples.
We then use the assistant to extract phrase-level concepts from the instructions as segmentation prompts. Since SAM3~\citep{carion2026sam} struggles with uncommon or ambiguous concepts, we crowdsource positive and negative points as spatial priors to improve segmentation accuracy.
Given the masks and manipulated videos, the assistant selects the four most relevant of eight predefined forensic perspectives (\emph{e.g.}, Boundary Coherence) to analyze the edited regions and generate anomaly explanations.
Finally, all authors carefully verify every real--fake video pair and its annotations against four criteria: \textit{1) task reasonableness}, \textit{2) manipulation quality}, \textit{3) explanation fidelity}, and \textit{4) mask accuracy}, with 24.8\% of samples deemed unqualified and discarded from our ManiVid-38K dataset.

\begin{figure*}[!tbp]
    \centering

    \includegraphics[width=\textwidth]{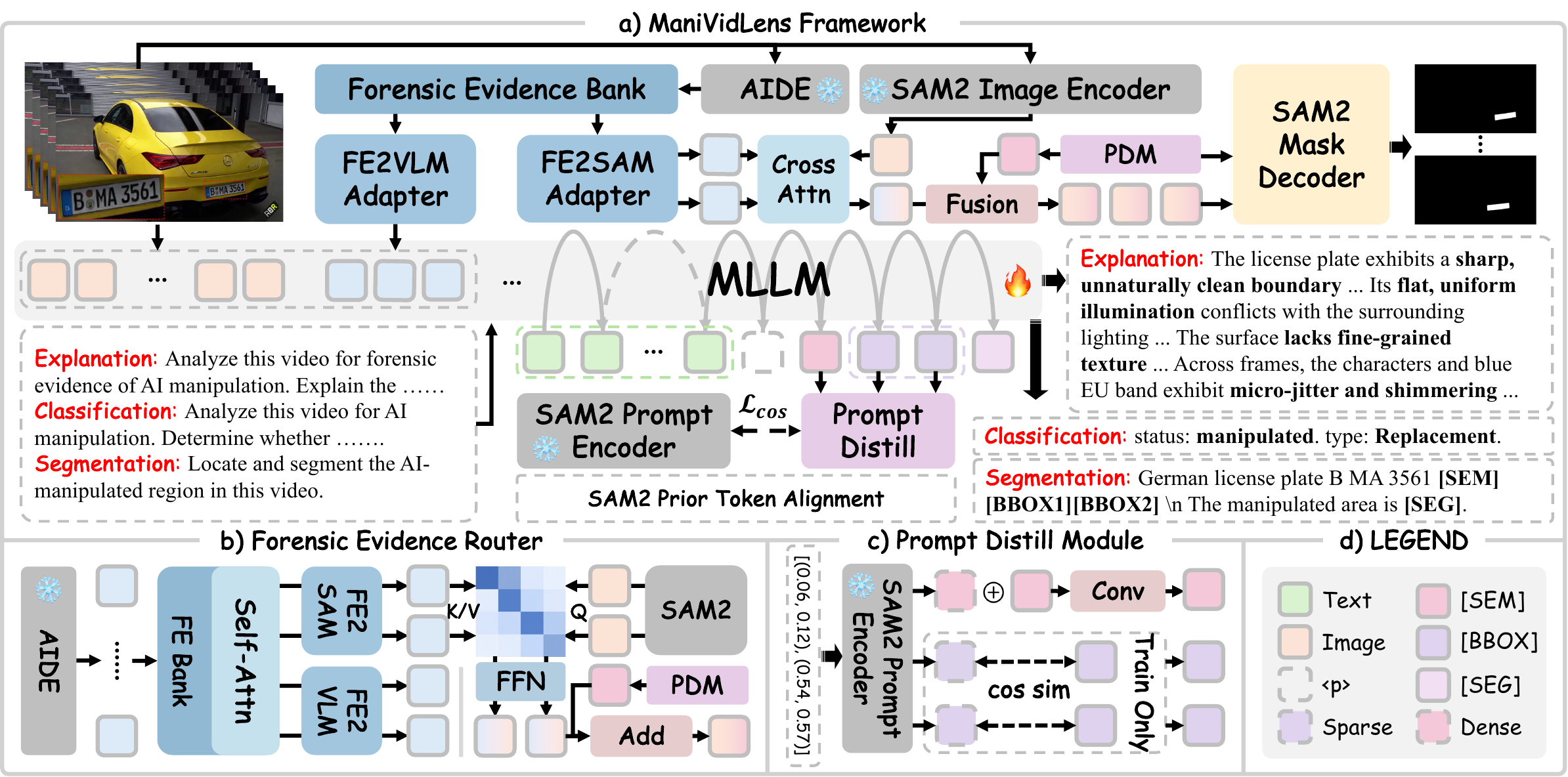}
    \vspace{-3.5mm}

    \caption{\textbf{Model Architecture.} \textbf{(a)} \textit{ManiVidLens} unifies forgery detection, artifact grounding, and anomaly explanation. \textbf{(b)} The \textit{Forensic Evidence Router} (FER) builds a compact forensic evidence bank and routes it to the MLLM and SAM2. \textbf{(c)} The \textit{Prompt Distill Module} (PDM) converts MLLM grounding-token states into SAM2 prompts. Dashed arrows mark \textbf{training-only} paths.}

    \vspace{-4mm}
    \label{fig:model}
\end{figure*}

\begin{figure*}[tbp]
    \centering

    \includegraphics[width=\textwidth]{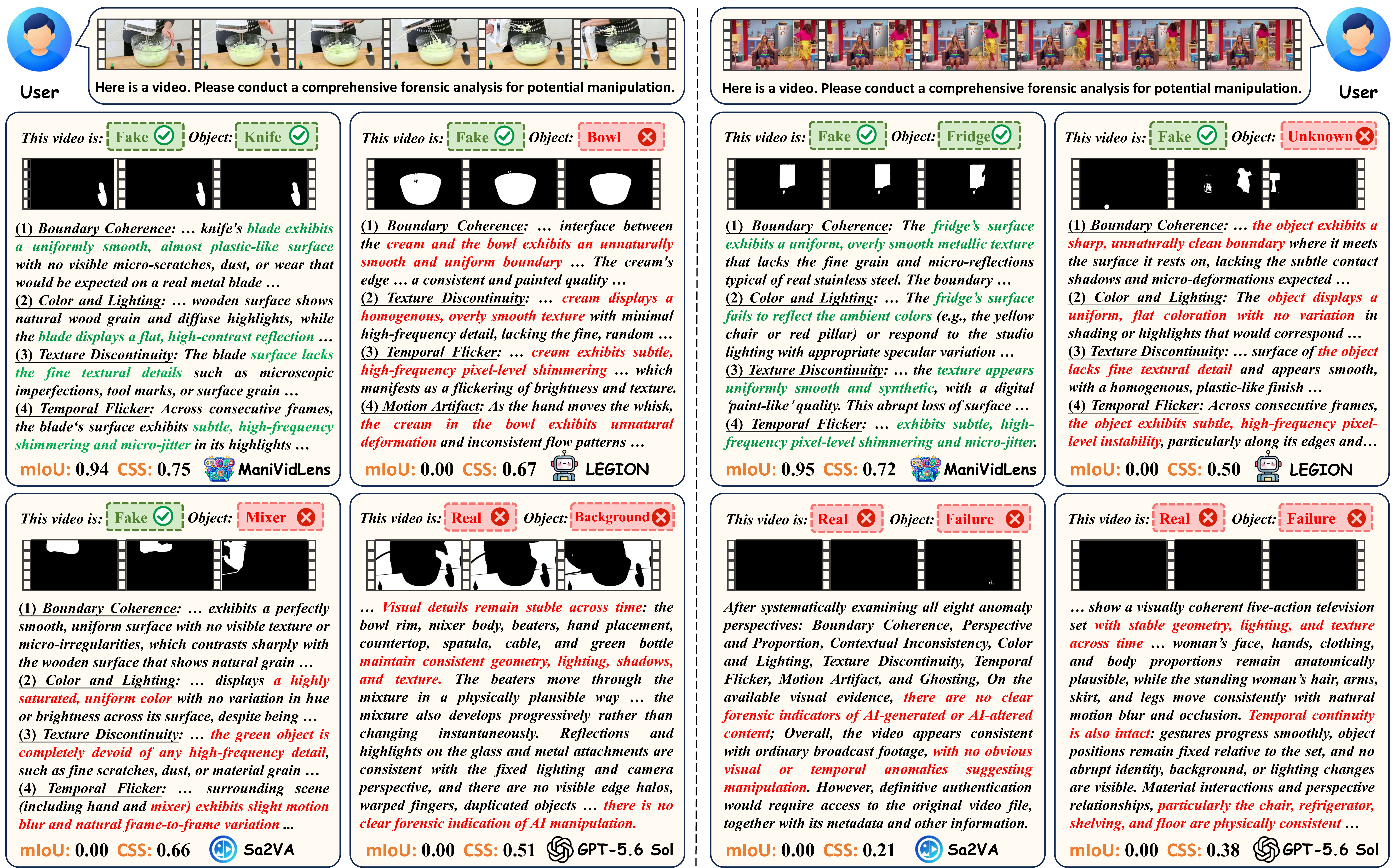}
    \vspace{-6mm}

    \caption{\textbf{Qualitative Comparison of ManiVidLens with Different Models.} We visualize two representative manipulation cases: (Left) adding a knife and (Right) changing the refrigerator material to steel. Across both cases, ManiVidLens demonstrates consistently better forensic analysis than LEGION (Image-based), Sa2VA (Video-based), and GPT-5.6 Sol (General MLLM).}

    \vspace{-4mm}
    \label{fig:main_case}
\end{figure*}

\vspace{-2mm}
\subsection{Video Forgery Dataset Comparison}
\label{dataset:analysis}
Tab.~\ref{tab:dataset_comparison} highlights the limited diversity, annotation granularity, and scale of existing video forgery datasets, particularly those focusing on manipulation. In contrast, ManiVid-38K is a diverse, open-vocabulary manipulation dataset for general videos. Unlike OpenVE-3M~\citep{he2025openve}, it ensures authentic source videos and provides comprehensive annotations, including authenticity labels, forgery masks, anomaly explanations, and intermediate outputs, enabling research on explainable video forgery analysis. Sections~\ref{app:statistical} and~\ref{app:case_visual} in the supplementary material present more analyses and visual examples.




\section{Method}

\vspace{-1mm}
\subsection{Overview}

ManiVidLens addresses ManiVid through the end-to-end pipeline in Figure~\ref{fig:model}(a). Given a video $\mathcal{V}=\{I_t\}_{t=1}^{T}$ and a task-specific instruction $q^r$, the model generates a response $\mathbf{y}^r$, where $r\in\{\mathrm{cls},\mathrm{seg},\mathrm{exp}\}$ denotes forgery detection, artifact grounding, or anomaly explanation.
All tasks share five temporally ordered conditioning frames $\mathcal{S}=\{I_{t_k}\}_{k=1}^{5}$, their evidence $\mathcal{A}=\{(\mathbf{R}_k,\mathbf{s}_k)\}_{k=1}^{5}$ from the frozen AIDE encoder~\citep{yan2025sanity}, and model parameters, but use task-specific instructions and response targets as their output spaces differ.

The detection response $\mathbf{y}^{\mathrm{cls}}$ states whether the video is manipulated, and the explanation response $\mathbf{y}^{\mathrm{exp}}$ describes the observed anomalies. For a manipulated video, the grounding response $\mathbf{y}^{\mathrm{seg}}$ gives a short target description followed by $[\mathrm{SEM}]$, $[\mathrm{BBOX1}]$, $[\mathrm{BBOX2}]$, and $[\mathrm{SEG}]$, whose states prompt SAM2 to decode masks on the conditioning frames and propagate them to all $T$ frames; for an original video, it states that no manipulated region exists and omits these tokens. Section~\ref{app:manividlens_protocol} in the supplementary material details the task-specific output protocols, while Section~\ref{app:manividlens_inference} describes the full inference pipeline.

\subsection{Forensic Evidence Router}

The Forensic Evidence Router (FER) turns frozen frame-level forensic cues into a shared evidence bank for the MLLM and SAM2 (Figure~\ref{fig:model}(b)). For each conditioning frame, the AIDE encoder returns four low-level forensic tokens $\mathbf{R}_k=\{\mathbf{r}_{k,j}\}_{j=1}^{4}$ and one semantic token $\mathbf{s}_k$. With $\bar{\mathbf{r}}_k$ as the mean forensic token, FER builds the evidence bank $\mathbf{E}_k\in\mathbb{R}^{6\times d_e}$ of each frame with one Transformer block:
\begin{equation}
    \label{eq:forensic-bank}
    \mathbf{E}_k=\operatorname{Attn}\bigl([\phi_s(\mathbf{s}_k),\Phi_r(\mathbf{R}_k),\phi_g(\mathbf{s}_k,\bar{\mathbf{r}}_k)]+\mathbf{T}_{\mathrm{bank}}\bigr),
\end{equation}
where $\phi_s$, $\Phi_r$, and $\phi_g$ project semantic, forensic, and global cues into a shared space, and the learned $\mathbf{T}_{\mathrm{bank}}$ distinguishes the six evidence slots of each bank.

For the MLLM, the \emph{Forensic Evidence-to-Vision-Language Model} (FE2VLM) adapter projects the global, semantic, and mean forensic tokens of each bank into the MLLM space and places them at the three $[\mathrm{AIDE}]$ positions after each frame. For SAM2, the \emph{Forensic Evidence-to-SAM} (FE2SAM) adapter projects the global and four forensic tokens into the SAM2 space and injects them into the frame's SAM2 image features via cross-attention. A shared authenticity gate downweights this evidence for videos predicted as original, shifting FE2VLM toward the learned $[\mathrm{AIDE}]$ embedding and FE2SAM toward the original SAM2 features. Section~\ref{app:manividlens_fer} in the supplementary material details them.

\subsection{Prompt Distill Module}

As shown in Figure~\ref{fig:model}(c), the Prompt Distill Module (PDM) maps the special-token states of the grounding response to SAM2 prompts: the $[\mathrm{SEM}]$ state is fused with SAM2's dense no-mask embedding for semantic context, the $[\mathrm{BBOX1}]$ and $[\mathrm{BBOX2}]$ states form sparse geometric prompts, and the $[\mathrm{SEG}]$ state supplies a segmentation prompt. These prompts guide mask decoding on the conditioning frames before SAM2 propagates the object state through the video. PDM reads teacher-forced token states during training and generated states at inference.
To give the box tokens a geometric prior, we take the minimum bounding rectangle of the valid boxes on the conditioning frames, and the frozen SAM2 prompt encoder maps its top-left and bottom-right corners to embeddings $\bar{\mathbf{b}}_{i,1}$ and $\bar{\mathbf{b}}_{i,2}$, which supervise the projected states $\mathbf{z}^{b_1}_i$ and $\mathbf{z}^{b_2}_i$ through a cosine distillation loss:
\begin{equation}
    \label{eq:prompt-distill-loss}
    \mathcal{L}_{\mathrm{cos}}
    = \frac{1}{2|\Omega|}
    \sum_{i\in\Omega}\sum_{j=1}^{2}
    \left(
    1 -
    \frac{\langle \mathbf{z}^{b_j}_i,\bar{\mathbf{b}}_{i,j}\rangle}
    {\|\mathbf{z}^{b_j}_i\|_2\|\bar{\mathbf{b}}_{i,j}\|_2}
    \right).
\end{equation}
where $\Omega$ indexes samples with a complete special-token span and at least one valid box. The enclosing boxes distill a video-level geometric prior, while SAM2 resolves per-frame locations through its image features and temporal memory.

\subsection{\texorpdfstring{Two-Stage Training Strategy}{Two-Stage Training Strategy}}
We train ManiVidLens in two stages $s\in\{1,2\}$ as follows:
\begin{equation}
    \label{eq:two_stage_training}
    \begin{gathered}
    \mathcal{L}^{(s)}=\lambda_{\mathrm{lm}}^{(s)}\mathcal{L}_{\mathrm{LM}}+\lambda_{\mathrm{cls}}^{(s)}\mathcal{L}_{\mathrm{cls}}+\delta_{s2}\mathcal{L}_{\mathrm{grd}},\\
    \mathcal{L}_{\mathrm{grd}}=\lambda_m\mathcal{L}_{\mathrm{mask}}+\lambda_d\mathcal{L}_{\mathrm{dice}}+\lambda_c\mathcal{L}_{\mathrm{cos}}.
    \end{gathered}
\end{equation}
$\mathcal{L}_{\mathrm{LM}}$ supervises the responses $\mathbf{y}^r$, $\mathcal{L}_{\mathrm{cls}}$ trains an auxiliary video-authenticity estimator, the mask cross-entropy $\mathcal{L}_{\mathrm{mask}}$ and Dice loss $\mathcal{L}_{\mathrm{dice}}$ supervise pixel masks, and $\mathcal{L}_{\mathrm{cos}}$ distills the box tokens.
Stage~1 learns the output protocols with all mask-grounding modules frozen. Stage~2 ($\delta_{s2}{=}1$) resumes from Stage~1, adds $\mathcal{L}_{\mathrm{grd}}$ on manipulated samples with mask annotations, and lowers the language and classification weights. Section~\ref{app:manividlens_training} in the supplementary material reports all loss weights, sampling ratios, and frozen modules.

\section{Experiments}
\begin{table*}[tbp]
    \centering
    \vspace{-2mm}
    \caption{\textbf{Artifact Grounding Performance.} Results are reported for each ManiVidBench manipulation category and their average. \textbf{Synth.} and \textbf{Mani.} denote fully synthetic and manipulated content, respectively. $\text{\faStarO}$ marks frame-wise approximations on 30 uniformly sampled frames; $\bm\square$ marks native box outputs evaluated with shared SAM2 propagation over every video frame.}
    \vspace{-2mm}
    \definecolor{mygray}{gray}{.92}
    \renewcommand{\arraystretch}{1.1}
    \belowrulesep=-0.25pt
    \aboverulesep=-0.25pt
    \resizebox{1.0\linewidth}{!}{
        \begin{tabular}{c|l|c|*{12}{c}|*{2}{c}}
        \toprule[1.5pt]
        & \multirow{2}{*}{\textbf{Method}} & \multirow{2}{*}{\textbf{Focus}} & \multicolumn{2}{c}{\textbf{Addition}} & \multicolumn{2}{c}{\textbf{Removal}} & \multicolumn{2}{c}{\textbf{Replacement}} & \multicolumn{2}{c}{\textbf{Color}} & \multicolumn{2}{c}{\textbf{Material}} & \multicolumn{2}{c|}{\textbf{Portrait}} & \multicolumn{2}{c}{\textbf{Avg.}} \\
        \cmidrule(l{3pt}r{3pt}){4-5} \cmidrule(l{3pt}r{3pt}){6-7} \cmidrule(l{3pt}r{3pt}){8-9} \cmidrule(l{3pt}r{3pt}){10-11} \cmidrule(l{3pt}r{3pt}){12-13} \cmidrule(l{3pt}r{3pt}){14-15} \cmidrule(l{3pt}r{3pt}){16-17}
        & & & mIoU & J\&F & mIoU & J\&F & mIoU & J\&F & mIoU & J\&F & mIoU & J\&F & mIoU & J\&F & mIoU & J\&F \\
        \hline
        \multirow{3}{*}{\rotatebox[origin=c]{90}{\textit{Image}}} & LEGION$^\text{\faStarO}$~\citep{kang2025legion} & Synth. & 0.343 & 0.352 & 0.091 & 0.102 & \underline{0.385} & \underline{0.392} & 0.323 & 0.346 & 0.360 & 0.370 & 0.192 & \underline{0.213} & 0.312 & 0.324 \\
        & FakeShield$^\text{\faStarO}$~\citep{xu2024fakeshield} & Mani. & 0.365 & 0.374 & 0.068 & 0.073 & 0.332 & 0.333 & \underline{0.429} & \underline{0.430} & 0.373 & 0.371 & 0.161 & 0.153 & 0.318 & 0.320 \\
        & SIDA$^\text{\faStarO}$~\citep{huang2025sida} & Both & 0.013 & 0.113 & 0.004 & 0.020 & 0.010 & 0.019 & 0.020 & 0.042 & 0.009 & 0.025 & 0.002 & 0.004 & 0.011 & 0.041 \\
        \hline
        \multirow{4}{*}{\rotatebox[origin=c]{90}{\textit{Video}}} & VideoLISA~\citep{bai2024one} & General & 0.191 & 0.202 & 0.064 & 0.077 & 0.166 & 0.183 & 0.111 & 0.128 & 0.161 & 0.174 & 0.029 & 0.050 & 0.138 & 0.153 \\
        & VISA~\citep{yan2024visa} & General & 0.157 & 0.173 & 0.054 & 0.060 & 0.076 & 0.086 & 0.040 & 0.052 & 0.080 & 0.090 & 0.018 & 0.022 & 0.079 & 0.089 \\
        & Sa2VA~\citep{yuan2025sa2va} & General & \underline{0.414} & \underline{0.432} & 0.132 & 0.135 & 0.343 & 0.356 & 0.390 & 0.406 & \underline{0.399} & \underline{0.406} & 0.182 & 0.186 & \underline{0.336} & \underline{0.348} \\
        & Omni-Fake-R1~\citep{li2026omni} & Both & 0.287 & 0.306 & 0.086 & 0.097 & 0.224 & 0.251 & 0.238 & 0.263 & 0.335 & 0.350 & 0.072 & 0.099 & 0.226 & 0.248 \\
        \hline
        \multirow{3}{*}{\rotatebox[origin=c]{90}{\textit{MLLM}}} & InternVL3.5-30B-A3B$^{\bm\square}$~\citep{wang2025internvl3} & General & 0.109 & 0.118 & 0.061 & 0.063 & 0.042 & 0.048 & 0.020 & 0.029 & 0.041 & 0.051 & 0.039 & 0.048 & 0.053 & 0.060 \\
        & Qwen3.5-35B-A3B$^{\bm\square}$~\citep{qwen3.5} & General & 0.381 & 0.395 & \textbf{0.221} & \textbf{0.222} & 0.301 & 0.319 & 0.294 & 0.310 & 0.355 & 0.370 & \underline{0.200} & 0.209 & 0.306 & 0.319 \\
        & GPT-5.6-Sol$^{\bm\square}$~\citep{openai2026gpt56} & General & 0.270 & 0.274 & \underline{0.192} & \underline{0.195} & 0.158 & 0.166 & 0.238 & 0.246 & 0.210 & 0.218 & 0.031 & 0.033 & 0.198 & 0.205 \\
        \midrule
        \multicolumn{2}{l|}{\textbf{ManiVidLens~(Ours)}} & \textbf{Mani.} & \textbf{0.444} & \textbf{0.467} & 0.135 & 0.137 & \textbf{0.421} & \textbf{0.439} & \textbf{0.487} & \textbf{0.507} & \textbf{0.526} & \textbf{0.532} & \textbf{0.290} & \textbf{0.306} & \textbf{0.407} & \textbf{0.422} \\
        \bottomrule[1.5pt]
        \end{tabular}
    }
    \vspace{-1mm}
    \label{tab:grounding}

\end{table*}

\begin{table*}[tbp]
    \centering
    \caption{\textbf{Anomaly Explanation Performance.} Results are reported for each ManiVidBench manipulation category and their average. \textbf{Synth.} and \textbf{Mani.} denote fully synthetic and manipulated content. \textbf{R-L} denotes \textit{ROUGE-L}, and \textbf{CSS} denotes \textit{Cosine Similarity Score}.}
    \vspace{-2mm}
    \definecolor{mygray}{gray}{.92}
    \renewcommand{\arraystretch}{1.1}
    \belowrulesep=-0.25pt
    \aboverulesep=-0.25pt
    \resizebox{1.0\linewidth}{!}{
        \begin{tabular}{c|l|c|*{12}{>{\centering\arraybackslash}p{0.85cm}}|*{2}{>{\centering\arraybackslash}p{0.85cm}}}
        \toprule[1.5pt]
        & \multirow{2}{*}{\textbf{Method}} & \multirow{2}{*}{\textbf{Focus}} & \multicolumn{2}{c}{\textbf{Addition}} & \multicolumn{2}{c}{\textbf{Removal}} & \multicolumn{2}{c}{\textbf{Replacement}} & \multicolumn{2}{c}{\textbf{Color}} & \multicolumn{2}{c}{\textbf{Material}} & \multicolumn{2}{c|}{\textbf{Portrait}} & \multicolumn{2}{c}{\textbf{Avg.}} \\
        \cmidrule(l{3pt}r{3pt}){4-5} \cmidrule(l{3pt}r{3pt}){6-7} \cmidrule(l{3pt}r{3pt}){8-9} \cmidrule(l{3pt}r{3pt}){10-11} \cmidrule(l{3pt}r{3pt}){12-13} \cmidrule(l{3pt}r{3pt}){14-15} \cmidrule(l{3pt}r{3pt}){16-17}
        & & & R-L & CSS & R-L & CSS & R-L & CSS & R-L & CSS & R-L & CSS & R-L & CSS & R-L & CSS \\
        \hline
        \multirow{3}{*}{\rotatebox[origin=c]{90}{\textit{Image}}} & LEGION~\citep{kang2025legion} & Synth. & \underline{0.255} & 0.693 & 0.239 & 0.681 & \underline{0.258} & \underline{0.727} & 0.251 & \underline{0.709} & 0.246 & \underline{0.721} & 0.247 & \underline{0.718} & 0.251 & \underline{0.710} \\
        & FakeShield~\citep{xu2024fakeshield} & Mani. & 0.182 & 0.559 & 0.128 & 0.483 & 0.165 & 0.541 & 0.174 & 0.579 & 0.162 & 0.558 & 0.116 & 0.455 & 0.162 & 0.541 \\
        & SIDA~\citep{huang2025sida} & Both & 0.249 & 0.674 & \underline{0.251} & \underline{0.694} & 0.255 & 0.681 & \underline{0.252} & 0.699 & \underline{0.252} & 0.700 & \underline{0.251} & \textbf{0.746} & \underline{0.252} & 0.691 \\
        \hline
        \multirow{3}{*}{\rotatebox[origin=c]{90}{\textit{Video}}} & VidGuard-R1~\citep{park2025vidguard} & Synth. & 0.200 & 0.541 & 0.194 & 0.569 & 0.202 & 0.587 & 0.206 & 0.621 & 0.200 & 0.608 & 0.200 & 0.597 & 0.201 & 0.586 \\
        & Sa2VA~\citep{yuan2025sa2va} & General & 0.206 & 0.619 & 0.161 & 0.538 & 0.171 & 0.556 & 0.176 & 0.565 & 0.184 & 0.589 & 0.081 & 0.382 & 0.174 & 0.561 \\
        & Omni-Fake-R1~\citep{li2026omni} & Both & 0.239 & \underline{0.703} & 0.224 & 0.671 & 0.233 & 0.688 & 0.226 & 0.689 & 0.226 & 0.693 & 0.212 & 0.668 & 0.230 & 0.688 \\
        \hline
        \multirow{3}{*}{\rotatebox[origin=c]{90}{\textit{MLLM}}} & InternVL3.5-30B-A3B~\citep{wang2025internvl3} & General & 0.090 & 0.431 & 0.088 & 0.432 & 0.090 & 0.435 & 0.087 & 0.425 & 0.088 & 0.448 & 0.085 & 0.412 & 0.089 & 0.432 \\
        & Qwen3.5-35B-A3B~\citep{qwen3.5} & General & 0.096 & 0.444 & 0.094 & 0.452 & 0.096 & 0.453 & 0.095 & 0.443 & 0.095 & 0.466 & 0.096 & 0.445 & 0.095 & 0.450 \\
        & GPT-5.6-Sol~\citep{openai2026gpt56} & General & 0.082 & 0.436 & 0.082 & 0.450 & 0.082 & 0.446 & 0.081 & 0.439 & 0.081 & 0.454 & 0.080 & 0.446 & 0.082 & 0.444 \\
        \midrule
        \multicolumn{2}{l|}{\textbf{ManiVidLens~(Ours)}} & \textbf{Mani.} & \textbf{0.603} & \textbf{0.813} & \textbf{0.576} & \textbf{0.773} & \textbf{0.613} & \textbf{0.831} & \textbf{0.582} & \textbf{0.772} & \textbf{0.587} & \textbf{0.795} & \textbf{0.545} & 0.699 & \textbf{0.593} & \textbf{0.796} \\
        \bottomrule[1.5pt]
        \end{tabular}
    }
    \vspace{-4mm}
    \label{tab:explanation}

\end{table*}

\begin{table*}[tbp]
    \centering
    \caption{\textbf{Forgery Detection Performance.} Results are reported for each ManiVidBench manipulation category and their average over all 2,000 videos. \textbf{Synth.} and \textbf{Mani.} denote fully synthetic and manipulated content. \textbf{Acc} and \textbf{F1} denote video-level accuracy and F1. Image expert models use majority voting over 30 uniformly sampled frames at a $0.5$ threshold.}
    \vspace{-2mm}
    \definecolor{mygray}{gray}{.92}
    \renewcommand{\arraystretch}{1.1}
    \belowrulesep=-0.25pt
    \aboverulesep=-0.25pt
    \resizebox{1.0\linewidth}{!}{
        \begin{tabular}{c|l|c|*{12}{>{\centering\arraybackslash}p{0.85cm}}|*{2}{>{\centering\arraybackslash}p{0.85cm}}}
        \toprule[1.5pt]
        & \multirow{2}{*}{\textbf{Method}} & \multirow{2}{*}{\textbf{Focus}} & \multicolumn{2}{c}{\textbf{Addition}} & \multicolumn{2}{c}{\textbf{Removal}} & \multicolumn{2}{c}{\textbf{Replacement}} & \multicolumn{2}{c}{\textbf{Color}} & \multicolumn{2}{c}{\textbf{Material}} & \multicolumn{2}{c|}{\textbf{Portrait}} & \multicolumn{2}{c}{\textbf{Avg.}} \\
        \cmidrule(l{3pt}r{3pt}){4-5} \cmidrule(l{3pt}r{3pt}){6-7} \cmidrule(l{3pt}r{3pt}){8-9} \cmidrule(l{3pt}r{3pt}){10-11} \cmidrule(l{3pt}r{3pt}){12-13} \cmidrule(l{3pt}r{3pt}){14-15} \cmidrule(l{3pt}r{3pt}){16-17}
        & & & Acc & F1 & Acc & F1 & Acc & F1 & Acc & F1 & Acc & F1 & Acc & F1 & Acc & F1 \\
        \hline
        \multirow{3}{*}{\rotatebox[origin=c]{90}{\textit{Image}}} & CNNSpot~\citep{wang2020cnn} & Synth. & 0.568 & 0.665 & 0.607 & 0.687 & 0.608 & 0.696 & 0.615 & 0.712 & 0.615 & 0.699 & 0.593 & 0.700 & 0.602 & 0.693 \\
        & UnivFD~\citep{ojha2023towards} & Synth. & 0.599 & 0.598 & 0.592 & 0.619 & 0.603 & 0.614 & 0.654 & 0.689 & 0.649 & 0.662 & 0.653 & 0.687 & 0.620 & 0.638 \\
        & AIDE~\citep{yan2025sanity} & Synth. & 0.866 & 0.865 & 0.882 & 0.881 & 0.818 & 0.819 & 0.849 & 0.850 & 0.851 & 0.856 & 0.746 & 0.741 & 0.842 & \underline{0.842} \\
        \hline
        \multirow{5}{*}{\rotatebox[origin=c]{90}{\textit{Video}}} & D3~\citep{zheng2025d3} & Synth. & 0.500 & 0.667 & 0.500 & 0.667 & 0.500 & 0.667 & 0.503 & 0.668 & 0.500 & 0.667 & 0.500 & 0.667 & 0.501 & 0.667 \\
        & VidGuard-R1~\citep{park2025vidguard} & Synth. & 0.521 & 0.627 & 0.508 & 0.610 & 0.525 & 0.631 & 0.573 & 0.663 & 0.519 & 0.613 & 0.517 & 0.617 & 0.530 & 0.631 \\
        & ReStraV~\citep{interno2026ai} & Synth. & 0.542 & 0.411 & 0.569 & 0.469 & 0.564 & 0.416 & 0.570 & 0.441 & 0.588 & 0.509 & 0.517 & 0.374 & 0.562 & 0.438 \\
        & NSG-VD~\citep{zhang2026physics} & Synth. & 0.521 & 0.646 & 0.496 & 0.629 & 0.503 & 0.638 & 0.492 & 0.626 & 0.515 & 0.648 & 0.449 & 0.606 & 0.502 & 0.635 \\
        & DeMamba~\citep{chen2026demamba} & Synth. & \textbf{0.974} & \textbf{0.973} & \underline{0.943} & \underline{0.940} & \textbf{0.890} & \textbf{0.885} & \textbf{0.924} & \textbf{0.922} & \underline{0.935} & \underline{0.932} & \underline{0.831} & \underline{0.818} & \textbf{0.922} & \textbf{0.919} \\
        \hline
        \multirow{3}{*}{\rotatebox[origin=c]{90}{\textit{MLLM}}} & InternVL3.5-30B-A3B~\citep{wang2025internvl3} & General & 0.503 & 0.349 & 0.508 & 0.358 & 0.508 & 0.355 & 0.503 & 0.339 & 0.508 & 0.352 & 0.508 & 0.356 & 0.506 & 0.351 \\
        & Qwen3.5-35B-A3B~\citep{qwen3.5} & General & 0.579 & 0.525 & 0.599 & 0.561 & 0.559 & 0.489 & 0.549 & 0.451 & 0.576 & 0.511 & 0.559 & 0.490 & 0.569 & 0.502 \\
        & GPT-5.6-Sol~\citep{openai2026gpt56} & General & 0.699 & 0.697 & 0.718 & 0.715 & 0.666 & 0.660 & 0.648 & 0.640 & 0.687 & 0.680 & 0.593 & 0.593 & 0.674 & 0.669 \\
        \midrule
        \multicolumn{2}{l|}{\textbf{ManiVidLens~(Ours)}} & \textbf{Mani.} & \underline{0.969} & \underline{0.968} & \textbf{0.947} & \textbf{0.945} & \underline{0.885} & \underline{0.883} & \underline{0.919} & \underline{0.919} & \textbf{0.939} & \textbf{0.941} & \textbf{0.839} & \textbf{0.846} & \underline{0.920} & \textbf{0.919} \\
        \bottomrule[1.5pt]
        \end{tabular}
    }
    \vspace{-4mm}
    \label{tab:detection}
    

\end{table*}


\begin{table*}[tbp]
\centering
\vspace{2mm}
\definecolor{mygray}{gray}{.92}
\renewcommand{\arraystretch}{1.1}
\belowrulesep=-0.25pt
\aboverulesep=-0.25pt
\begin{minipage}[t]{\columnwidth}
\centering

\providecommand{\ManiVidFramesCaption}{\textbf{Ablation on Sampled Frames.}
5 frames are used by default. Inference time sums all three tasks, averaged per video.}

\caption{\ManiVidFramesCaption}
\label{tab:ablation_frames}
\vspace{-2mm}

\resizebox{\linewidth}{!}{
    \begin{tabular}{l|*{6}{c}|c}
    \toprule[1.5pt]
    \multirow{2}{*}{\textbf{Setting}} & \multicolumn{2}{c}{\textbf{Grounding}} & \multicolumn{2}{c}{\textbf{Explanation}} & \multicolumn{2}{c|}{\textbf{Detection}} & \multirow{2}{*}{\shortstack{\textbf{Inference}\\\textbf{Time}~$\downarrow$}} \\
    \cmidrule(l{3pt}r{3pt}){2-3} \cmidrule(l{3pt}r{3pt}){4-5} \cmidrule(l{3pt}r{3pt}){6-7}
    & mIoU & J\&F & R-L & CSS & Acc & F1 \\
    \hline
    3 Frames & 0.354 & 0.368 & 0.575 & 0.763 & 0.906 & 0.905 & \textbf{13.83s} \\
    5 Frames & 0.407 & \underline{0.422} & \underline{0.593} & 0.796 & 0.920 & \underline{0.919} & \underline{15.98s} \\
    7 Frames & \underline{0.408} & \textbf{0.431} & \textbf{0.596} & \textbf{0.803} & \underline{0.926} & \textbf{0.926} & 17.72s \\
    9 Frames & \textbf{0.413} & 0.416 & 0.592 & \underline{0.797} & \textbf{0.927} & \textbf{0.926} & 19.20s \\
    \bottomrule[1.5pt]
    \end{tabular}
}
\vspace{-2mm}
\end{minipage}
\hfill
\begin{minipage}[t]{\columnwidth}
\centering
\setlength{\tabcolsep}{8.4pt} 

\providecommand{\ManiVidModulesCaption}{\textbf{Ablation on PDM and FER.}
Base omits both modules; + PDM and + FER add one module each; + Both uses both.}
\caption{\ManiVidModulesCaption}
\label{tab:ablation_module}
\vspace{-2mm}

\resizebox{\linewidth}{!}{
    \begin{tabular}{l|*{6}{c}}
    \toprule[1.5pt]
    \multirow{2}{*}{\textbf{Setting}} & \multicolumn{2}{c}{\textbf{Grounding}} & \multicolumn{2}{c}{\textbf{Explanation}} & \multicolumn{2}{c}{\textbf{Detection}} \\
    \cmidrule(l{3pt}r{3pt}){2-3} \cmidrule(l{3pt}r{3pt}){4-5} \cmidrule(l{3pt}r{3pt}){6-7}
    & mIoU & J\&F & R-L & CSS & Acc & F1 \\
    \hline
    Base & 0.336 & 0.348 & 0.552 & 0.758 & 0.882 & 0.893 \\
    + PDM & 0.368 & 0.380 & 0.565 & 0.771 & 0.903 & 0.905 \\
    + FER & \underline{0.378} & \underline{0.389} & \textbf{0.598} & \textbf{0.808} & \underline{0.918} & \underline{0.916} \\
    + Both~(Ours) & \textbf{0.407} & \textbf{0.422} & \underline{0.593} & \underline{0.796} & \textbf{0.920} & \textbf{0.919} \\
    \bottomrule[1.5pt]
    \end{tabular}
}
\vspace{-2mm}
\end{minipage}
\vspace{-3mm}

\end{table*}

\subsection{Experimental Setup}

\noindent\textbf{Implementation Details.}
ManiVidLens uses Qwen3-VL-4B-Instruct~\citep{bai2025qwen3} and SAM2 Hiera-L~\citep{ravi2025sam} as its MLLM and video segmentation backbones. We train it on the ManiVid-38K training split for one epoch per stage with task-specific prompts and response formats. All image and video experts are trained on the same split, whereas general MLLMs are not fine-tuned. All models are evaluated on ManiVidBench's 2K videos across six manipulation categories.

\noindent\textbf{Evaluation Metrics.}
We evaluate \textit{artifact grounding} with mean IoU (mIoU) and J\&F~\citep{yuan2025sa2va}, \textit{anomaly explanation} with ROUGE-L~\citep{lin2004rouge} and Cosine Similarity Score (CSS) using MiniLM embeddings~\citep{reimers2019sentence,xu2024fakeshield}, and \textit{forgery detection} with video-level accuracy and F1. For grounding, image experts predict masks on 30 uniformly sampled frames as an approximation of the complete video, whereas general MLLMs generate only boxes from five conditioning frames. These boxes prompt shared SAM2 mask propagation across every frame. Sections~\ref{app:frame_sampling_scope} and~\ref{app:evaluation_metrics} provides further implementation and evaluation details.
\textbf{\texttt{Bold}} and \underline{\texttt{underlined}} values mark the best and second-best results per column.

\subsection{Main Results}
\label{sec:main_results}

\noindent\textbf{Artifact Grounding.}
ManiVidLens reaches 0.407 mIoU and 0.422 J\&F (Table~\ref{tab:grounding}), surpassing Sa2VA, the strongest overall comparison, by \textbf{0.071 mIoU} and \textbf{0.074 J\&F}. The strongest image expert (FakeShield) and general MLLM (Qwen3.5) reach only 0.318 and 0.306 mIoU, respectively, showing that our advantage holds over both direct mask prediction and SAM2 propagation from predicted boxes. ManiVidLens leads both metrics in five of six categories, with its largest mIoU margins on Material (\textbf{+0.127}) and Portrait (\textbf{+0.090}). Color and Material edits preserve object identity while changing appearance; their 0.487 and 0.526 mIoU scores show that the advantage extends to edits that object recognition alone cannot expose. Removal is the sole exception (0.135 mIoU vs. Qwen3.5's 0.221) despite 0.947 detection accuracy: the reference mask covers the removed object's extent, but the input region contains reconstructed background without object semantics, exposing a gap between detection and precise spatial recovery of edited regions.

\noindent\textbf{Anomaly Explanation.}
ManiVidLens obtains a ROUGE-L of 0.593 and a CSS of 0.796 (Table~\ref{tab:explanation}), exceeding the strongest comparison for each metric by \textbf{0.341} and \textbf{0.086}, respectively. It leads all six categories in ROUGE-L and five in CSS; Portrait is the sole CSS exception (0.699 vs. SIDA's 0.746).
Because these metrics may be sensitive to lexical overlap and structured formats, Section~\ref{app:human_evaluation} in supplementary material complements them with a human evaluation.

\noindent\textbf{Forgery Detection.}
ManiVidLens achieves 0.920 accuracy and 0.919 F1 (Table~\ref{tab:detection}), approaching the dedicated video detector DeMamba with a gap of only 0.002 in accuracy and a tie on F1 while sharing one model across all three tasks. It leads both metrics on Removal, Material, and Portrait. Together with the parallel gains in grounding, this suggests that shared cues help both to localize edits and to judge authenticity.
Section~\ref{app:attribution} further reports results on the more challenging task of manipulation attribution.


\noindent\textbf{Qualitative Analysis.}
Figure~\ref{fig:main_case} shows that ManiVidLens aligns object identification, masks across successive video frames, and forensic explanations, attributing boundary and texture anomalies to the localized objects.
Image experts, represented by LEGION, confuse the knife with the bowl and produce varying refrigerator masks across frames, consistent with their lack of temporal modeling.
Video experts, represented by Sa2VA, select the mixer in the first case and miss the material edit in the second, illustrating the need for low-level forensic cues alongside temporal context.
General MLLMs, represented by GPT-5.6 Sol, classify both manipulations as real yet return full-frame boxes under grounding prompts, allowing SAM2 masks to overlap targets incidentally.
Section~\ref{app:more_performance_analysis} provides per-category analyses across all tasks and ManiVidLens results for each generation model.

\subsection{Ablation Study}
\label{sec:ablation_study}

\noindent\textbf{Sampled Frames.}
Moving from three to five frames raises mIoU \textbf{from 0.354 to 0.407}, J\&F from 0.368 to 0.422, and inference time from 13.83\,s to 15.98\,s (Table~\ref{tab:ablation_frames}).
The mIoU gain covers all six categories, led by Color (+0.092) and Material (+0.076).
Seven frames add 0.001 mIoU, 0.007 CSS, and 0.006 accuracy at 10.9\% higher cost, reaching the best J\&F and explanation scores (0.596 ROUGE-L and 0.803 CSS).
Their small aggregate mIoU gain combines improvements in Addition and Portrait with declines of 0.040 and 0.048 in Color and Material.
Nine frames achieve the best mIoU (0.413) and detection accuracy (0.927) in 19.20\,s, adding only 0.006 mIoU over five frames while J\&F falls from 0.431 at seven frames to 0.416.
These task-dependent peaks favor five frames as our default, capturing most of the mIoU gain at reduced inference cost across all three tasks.

\noindent\textbf{Architectural Components.}
PDM and FER each improve all three tasks over Base (Table~\ref{tab:ablation_module}), independently raising mIoU by 0.032 and 0.042 and J\&F by 0.032 and 0.041, respectively.
Together, they reach \textbf{0.407 mIoU and 0.422 J\&F}, gains of 0.071 and 0.074 over Base, with mIoU improvements in all six categories.
Adding PDM to FER further improves Color (+0.065 mIoU), Material (+0.077), and Portrait (+0.088).
On Portrait, PDM alone falls from 0.182 to 0.170, while FER alone reaches 0.202 and the combination reaches 0.290, demonstrating the complementary roles of learned prompts and forensic evidence in localization.
PDM also improves ROUGE-L and CSS by 0.013 each, and accuracy and F1 by 0.021 and 0.012.
FER alone gives the best explanation scores (0.598 ROUGE-L and 0.808 CSS), compared with 0.593 and 0.796 for the full model.
The full model leads detection at 0.920 accuracy and 0.919 F1, gains of \textbf{0.038} and \textbf{0.026} over Base.
Additional OOD and robustness analyses are reported in Sections~\ref{app:cross_dataset_detection} and \ref{app:robustness}.


\section{Conclusion}


We introduce \textbf{ManiVid} for unified and explainable forensic analysis of manipulated videos, supported by the \textbf{ManiVid-38K} dataset and \textbf{ManiVidBench}. \textbf{ManiVid-38K} is the first dataset combining paired, open-vocabulary localized video manipulations with authenticity labels, pixel-level masks, and anomaly explanations. 
\textbf{ManiVidLens} combines an MLLM and SAM2 through task-specific protocols, shared forensic evidence, and structured prompts to detect forgery, localize artifacts, and explain anomalies.
On \textbf{ManiVidBench}, it achieves state-of-the-art artifact grounding and anomaly explanation with detection comparable to dedicated classifiers, while precisely localizing object removal remains an important open challenge. The results highlight complementary gains from forensic evidence and structured prompts. Together, these resources support developing and evaluating more interpretable video manipulation forensics.

\clearpage
\section*{LLM Usage Statement}
We used AI tools (ChatGPT) at a \textbf{minimal} level and \emph{only} for linguistic polishing (grammar, spelling, punctuation, and minor word-choice/style edits). All technical contributions and writing decisions were authored and verified by the authors, and no confidential or proprietary data were provided to AI services.
We also used video generation models, including Kling, to construct ManiVid-38K dataset (see the supplementary material for the full list of models).

{\small
\bibliographystyle{ieeenat_fullname}
\bibliography{main}

@inproceedings{yan2025sanity,
  title={A sanity check for ai-generated image detection},
  author={Yan, Shilin and Li, Ouxiang and Cai, Jiayin and Hao, Yanbin and Jiang, Xiaolong and Hu, Yao and Xie, Weidi},
  booktitle={International Conference on Learning Representations},
  volume={2025},
  pages={70702--70720},
  year={2025}
}

@inproceedings{li2025aegis,
  title={AEGIS: Authenticity Evaluation Benchmark for AI-Generated Video Sequences},
  author={Li, Jieyu and Zhang, Xin and Zhou, Joey Tianyi},
  booktitle={Proceedings of the 33rd ACM International Conference on Multimedia},
  pages={13346--13353},
  year={2025}
}

@article{interno2026ai,
  title={{AI}-Generated Video Detection via Perceptual Straightening},
  author={Intern{\`o}, Christian and Geirhos, Robert and Olhofer, Markus and Liu, Sunny and Hammer, Barbara and Klindt, David},
  journal={Advances in Neural Information Processing Systems},
  volume={38},
  pages={20672--20705},
  year={2025}
}

@inproceedings{bai2024ai,
  title={Ai-generated video detection via spatial-temporal anomaly learning},
  author={Bai, Jianfa and Lin, Man and Cao, Gang and Lou, Zijie},
  booktitle={Chinese Conference on Pattern Recognition and Computer Vision (PRCV)},
  pages={460--470},
  year={2024},
  organization={Springer}
}

@inproceedings{vahdati2024beyond,
  title={Beyond deepfake images: Detecting ai-generated videos},
  author={Vahdati, Danial Samadi and Nguyen, Tai D and Azizpour, Aref and Stamm, Matthew C},
  booktitle={2024 IEEE/CVF Conference on Computer Vision and Pattern Recognition Workshops (CVPRW)},
  pages={4397--4408},
  year={2024},
  organization={IEEE}
}

@inproceedings{zhu2022celebv,
  title={CelebV-HQ: A large-scale video facial attributes dataset},
  author={Zhu, Hao and Wu, Wayne and Zhu, Wentao and Jiang, Liming and Tang, Siwei and Zhang, Li and Liu, Ziwei and Loy, Chen Change},
  booktitle={European conference on computer vision},
  pages={650--667},
  year={2022},
  organization={Springer}
}

@article{yuan2024chronomagic,
  title={Chronomagic-bench: A benchmark for metamorphic evaluation of text-to-time-lapse video generation},
  author={Yuan, Shenghai and Huang, Jinfa and Xu, Yongqi and Liu, Yaoyang and Zhang, Shaofeng and Shi, Yujun and Zhu, Ruijie and Cheng, Xinhua and Luo, Jiebo and Yuan, Li},
  journal={Advances in Neural Information Processing Systems},
  volume={37},
  pages={21236--21270},
  year={2024}
}

@inproceedings{wang2020cnn,
  title={CNN-generated images are surprisingly easy to spot… for now},
  author={Wang, Sheng-Yu and Wang, Oliver and Zhang, Richard and Owens, Andrew and Efros, Alexei A},
  booktitle={2020 IEEE/CVF Conference on Computer Vision and Pattern Recognition (CVPR)},
  pages={8692--8701},
  year={2020},
  organization={IEEE}
}

@inproceedings{zheng2025d3,
  title={D3: Training-free ai-generated video detection using second-order features},
  author={Zheng, Chende and Suo, Ruiqi and Lin, Chenhao and Zhao, Zhengyu and Yang, Le and Liu, Shuai and Yang, Minghui and Wang, Cong and Shen, Chao},
  booktitle={Proceedings of the IEEE/CVF International Conference on Computer Vision},
  pages={12852--12862},
  year={2025}
}

@article{chen2026demamba,
  title={Demamba: Ai-generated video detection on million-scale genvideo benchmark},
  author={Chen, Haoxing and Hong, Yan and Huang, Zizheng and Xu, Zhuoer and Gu, Zhangxuan and Li, Yaohui and Lan, Jun and Zhu, Huijia and Zhang, Jianfu and Wang, Weiqiang and others},
  journal={Science China Information Sciences},
  volume={69},
  number={6},
  pages={162103},
  year={2026},
  publisher={Springer}
}

@article{rossler2018faceforensics,
  title={Faceforensics: A large-scale video dataset for forgery detection in human faces},
  author={R{\"o}ssler, Andreas and Cozzolino, Davide and Verdoliva, Luisa and Riess, Christian and Thies, Justus and Nie{\ss}ner, Matthias},
  journal={arXiv preprint arXiv:1803.09179},
  year={2018}
}

@inproceedings{rossler2019faceforensics++,
  title={Faceforensics++: Learning to detect manipulated facial images},
  author={Rossler, Andreas and Cozzolino, Davide and Verdoliva, Luisa and Riess, Christian and Thies, Justus and Nie{\ss}ner, Matthias},
  booktitle={Proceedings of the IEEE/CVF international conference on computer vision},
  pages={1--11},
  year={2019}
}

@article{xu2024fakeshield,
  title={Fakeshield: Explainable image forgery detection and localization via multi-modal large language models},
  author={Xu, Zhipei and Zhang, Xuanyu and Li, Runyi and Tang, Zecheng and Huang, Qing and Zhang, Jian},
  journal={The Thirteenth International Conference on Learning Representations},
  year={2025}
}

@article{zhu2026fakevlm,
  title={FakeVLM-R1: Internalizing Physical Laws via CoT for Synthetic Image Detection},
  author={Zhu, Leqi and Ye, Junyan and Lin, Kaiqing and Yan, Zhiyuan and He, Conghui and Li, Weijia},
  journal={arXiv preprint arXiv:2605.30062},
  year={2026}
}

@misc{blackforestlabs_flux2_dev,
  author       = {{Black Forest Labs}},
  title        = {{FLUX.2 [dev] Model Card}},
  year         = {2025},
  howpublished = {\url{https://huggingface.co/black-forest-labs/FLUX.2-dev}},
  note         = {Accessed: 2026-08-03}
}

@misc{blackforestlabs_flux2_klein_9b,
  author       = {{Black Forest Labs}},
  title        = {{FLUX.2 [klein] 9B Model Card}},
  year         = {2026},
  howpublished = {\url{https://huggingface.co/black-forest-labs/FLUX.2-klein-9B}},
  note         = {Accessed: 2026-08-03}
}

@article{ni2026genvidbench,
  title={{GenVidBench}: A 6-Million Benchmark for {AI}-Generated Video Detection},
  author={Ni, Zhenliang and Yan, Qiangyu and Huang, Mouxiao and Yuan, Tianning and Tang, Yehui and Hu, Hailin and Chen, Xinghao and Wang, Yunhe},
  journal={Proceedings of the AAAI Conference on Artificial Intelligence},
  volume={40},
  number={18},
  pages={15582--15590},
  year={2026},
  doi={10.1609/aaai.v40i18.38587}
}

@misc{openai2026gpt56,
  author       = {{OpenAI}},
  title        = {{GPT-5.6}},
  year         = {2026},
  month        = jul,
  howpublished = {\url{https://deploymentsafety.openai.com/gpt-5-6}},
  note         = {Accessed: 2026-09-08}
}

@inproceedings{wang2024internvid,
  title={Internvid: A large-scale video-text dataset for multimodal understanding and generation},
  author={Wang, Yi and He, Yinan and Li, Yizhuo and Li, Kunchang and Yu, Jiashuo and Ma, Xin and Li, Xinhao and Chen, Guo and Chen, Xinyuan and Wang, Yaohui and others},
  booktitle={International Conference on Learning Representations},
  volume={2024},
  pages={42055--42079},
  year={2024}
}

@article{wang2025internvl3,
  title={Internvl3. 5: Advancing open-source multimodal models in versatility, reasoning, and efficiency},
  author={Wang, Weiyun and Gao, Zhangwei and Gu, Lixin and Pu, Hengjun and Cui, Long and Wei, Xingguang and Liu, Zhaoyang and Jing, Linglin and Ye, Shenglong and Shao, Jie and others},
  journal={arXiv preprint arXiv:2508.18265},
  year={2025}
}

@misc{openai2025gptimage15,
  author       = {{OpenAI}},
  title        = {Introducing GPT-Image-1.5},
  year         = {2025},
  month        = dec,
  howpublished = {\url{https://openai.com/index/new-chatgpt-images-is-here/}},
  note         = {Accessed: 2026-08-24}
}

@article{team2025kling,
  title={Kling-omni technical report},
  author={Team, Kling and Chen, Jialu and Ci, Yuanzheng and Du, Xiangyu and Feng, Zipeng and Gai, Kun and Guo, Sainan and Han, Feng and He, Jingbin and He, Kang and others},
  journal={arXiv preprint arXiv:2512.16776},
  year={2025}
}

@misc{kling_video_3_omni,
  author  = {{Kling AI}},
  title   = {{Kling VIDEO 3.0 Omni Model User Guide}},
  year    = {2026},
  month   = feb,
  howpublished = {\url{https://kling.ai/quickstart/klingai-video-3-omni-model-user-guide}},
  note = {Accessed: 2026-08-03}
}

@inproceedings{kang2025legion,
  title={Legion: Learning to ground and explain for synthetic image detection},
  author={Kang, Hengrui and Wen, Siwei and Wen, Zichen and Ye, Junyan and Li, Weijia and Feng, Peilin and Zhou, Baichuan and Wang, Bin and Lin, Dahua and Zhang, Linfeng and others},
  booktitle={2025 IEEE/CVF International Conference on Computer Vision (ICCV)},
  pages={18937--18947},
  year={2025},
  organization={IEEE}
}

@article{hu2021lora,
  title={Lora: Low-rank adaptation of large language models},
  author={Hu, Edward J and Shen, Yelong and Wallis, Phillip and Allen-Zhu, Zeyuan and Li, Yuanzhi and Wang, Shean and Wang, Lu and Chen, Weizhu},
  journal={arXiv preprint arXiv:2106.09685},
  year={2021}
}

@article{hacohen2026ltx,
  title={Ltx-2: Efficient joint audio-visual foundation model},
  author={HaCohen, Yoav and Brazowski, Benny and Chiprut, Nisan and Bitterman, Yaki and Kvochko, Andrew and Berkowitz, Avishai and Shalem, Daniel and Lifschitz, Daphna and Moshe, Dudu and Porat, Eitan and others},
  journal={arXiv preprint arXiv:2601.03233},
  year={2026}
}

@misc{google_nano_banana_2,
  author  = {{Google}},
  title   = {{Nano Banana 2: Gemini AI Image Generator and Photo Editor}},
  year    = {2026},
  howpublished = {\url{https://gemini.google/overview/image-generation/}},
  note = {Accessed: 2026-08-03}
}

@inproceedings{li2026omni,
  title={Omni-Fake: Benchmarking Unified Multimodal Social Media Deepfake Detection},
  author={Li, Tianxiao and Huang, Zhenglin and Wen, Haiquan and He, Yiwei and Li, Xinze and Zhu, Bingyu and Duan, Wuhui and Chen, Congang and Fu, Zeyu and Dong, Yi and others},
  booktitle={Proceedings of the IEEE/CVF Conference on Computer Vision and Pattern Recognition},
  pages={30299--30311},
  year={2026}
}

@article{zhang2026limitations,
  title={On the Limitations of Cross-Lingual Consistency in Multilingual Text-to-image Generation},
  author={Zhang, Sicheng and Yan, Zhonghao and Xie, Binzhu and Qiu, Shi and Naseer, Muzammal and Akhtar, Naveed and Shah, Mubarak},
  journal={arXiv preprint arXiv:2608.11002},
  year={2026}
}

@article{bai2024one,
  title={One token to seg them all: Language instructed reasoning segmentation in videos},
  author={Bai, Zechen and He, Tong and Mei, Haiyang and Wang, Pichao and Gao, Ziteng and Chen, Joya and Liu, Lei and Zhang, Zheng and Shou, Mike Z},
  journal={Advances in Neural Information Processing Systems},
  volume={37},
  pages={6833--6859},
  year={2024}
}

@misc{pku_yuan_lab_2024_opensoraplan,
  author    = {{PKU-Yuan Lab and Tuzhan AI et al.}},
  title     = {Open-Sora-Plan},
  month     = apr,
  year      = {2024},
  howpublished = {GitHub and Zenodo, \url{https://doi.org/10.5281/zenodo.10948109}}
}

@article{he2025openve,
  title={OpenVE-3M: A Large-Scale High-Quality Dataset for Instruction-Guided Video Editing},
  author={He, Haoyang and Wang, Jie and Zhang, Jiangning and Xue, Zhucun and Bu, Xingyuan and Yang, Qiangpeng and Wen, Shilei and Xie, Lei},
  journal={arXiv preprint arXiv:2512.07826},
  year={2025}
}

@inproceedings{nan2025openvid,
  title={Openvid-1m: A large-scale high-quality dataset for text-to-video generation},
  author={Nan, Kepan and Xie, Rui and Zhou, Penghao and Fan, Tiehan and Yang, Zhenheng and Chen, Zhijie and Li, Xiang and Yang, Jian and Tai, Ying},
  booktitle={International Conference on Learning Representations},
  volume={2025},
  pages={1045--1064},
  year={2025}
}

@inproceedings{chen2024panda70m,
  title     = {Panda-70M: Captioning 70M Videos with Multiple Cross-Modality Teachers},
  author    = {Chen, Tsai-Shien and Siarohin, Aliaksandr and Menapace, Willi and Deyneka, Ekaterina and Chao, Hsiang-wei and Jeon, Byung Eun and Fang, Yuwei and Lee, Hsin-Ying and Ren, Jian and Yang, Ming-Hsuan and Tulyakov, Sergey},
  booktitle = {Proceedings of the IEEE/CVF Conference on Computer Vision and Pattern Recognition},
  pages     = {13320--13331},
  year      = {2024}
}

@misc{sentence_transformers_minilm,
  author       = {Reimers, Nils and Gurevych, Iryna},
  title        = {paraphrase-MiniLM-L6-v2},
  year         = {2020},
  howpublished = {Hugging Face Model Hub},
  note         = {Sentence-Transformers model},
  url          = {https://huggingface.co/sentence-transformers/paraphrase-MiniLM-L6-v2}
}

@article{zhang2026physics,
  title={Physics-Driven Spatiotemporal Modeling for {AI}-Generated Video Detection},
  author={Zhang, Shuhai and Lian, ZiHao and Yang, Jiahao and Li, Daiyuan and Pang, Guoxuan and Liu, Feng and Han, Bo and Li, Shutao and Tan, Mingkui},
  journal={Advances in Neural Information Processing Systems},
  volume={38},
  pages={174683--174730},
  year={2025}
}

@inproceedings{carreira2017quo,
  title={Quo vadis, action recognition? a new model and the kinetics dataset},
  author={Carreira, Joao and Zisserman, Andrew},
  booktitle={2017 IEEE conference on computer vision and pattern recognition (CVPR)},
  pages={4724--4733},
  year={2017},
  organization={IEEE}
}

@article{wu2025qwen,
  title={Qwen-image technical report},
  author={Wu, Chenfei and Li, Jiahao and Zhou, Jingren and Lin, Junyang and Gao, Kaiyuan and Yan, Kun and Yin, Sheng-ming and Bai, Shuai and Xu, Xiao and Chen, Yilei and others},
  journal={arXiv preprint arXiv:2508.02324},
  year={2025}
}

@misc{qwen3.5,
    title  = {{Qwen3.5}: Towards Native Multimodal Agents},
    author = {{Qwen Team}},
    month  = {February},
    year   = {2026},
    url    = {https://qwen.ai/blog?id=qwen3.5}
}

@article{bai2025qwen3,
  title={Qwen3-vl technical report},
  author={Bai, Shuai and Cai, Yuxuan and Chen, Ruizhe and Chen, Keqin and Chen, Xionghui and Cheng, Zesen and Deng, Lianghao and Ding, Wei and Gao, Chang and Ge, Chunjiang and others},
  journal={arXiv preprint arXiv:2511.21631},
  year={2025}
}

@inproceedings{lin2004rouge,
  title={Rouge: A package for automatic evaluation of summaries},
  author={Lin, Chin-Yew},
  booktitle={Text summarization branches out},
  pages={74--81},
  year={2004}
}

@misc{runwayaleph,
  author       = {{Runway}},
  title        = {{Runway Aleph: A New Way to Edit, Transform and Generate Video}},
  year         = {2025},
  howpublished = {\url{https://runway.com/research/introducing-runway-aleph}},
  note         = {Accessed: 2026-08-03}
}

@article{yuan2025sa2va,
  title={Sa2va: Marrying sam2 with llava for dense grounded understanding of images and videos},
  author={Yuan, Haobo and Li, Xiangtai and Zhang, Tao and Sun, Yueyi and Huang, Zilong and Xu, Shilin and Ji, Shunping and Tong, Yunhai and Qi, Lu and Feng, Jiashi and others},
  journal={arXiv preprint arXiv:2501.04001},
  year={2025}
}

@inproceedings{ravi2025sam,
  title={Sam 2: Segment anything in images and videos},
  author={Ravi, Nikhila and Gabeur, Valentin and Hu, Yuan-Ting and Hu, Ronghang and Ryali, Chaitanya and Ma, Tengyu and Khedr, Haitham and R{\"a}dle, Roman and Rolland, Chloe and Gustafson, Laura and others},
  booktitle={International Conference on Learning Representations},
  volume={2025},
  pages={28085--28128},
  year={2025}
}

@inproceedings{carion2026sam,
  title={Sam 3: Segment anything with concepts},
  author={Carion, Nicolas and Gustafson, Laura and Hu, Yuan-Ting and Debnath, Shoubhik and Hu, Ronghang and Suris Coll-Vinent, Didac and Ryali, Chaitanya and Alwala, Kalyan Vasudev and Khedr, Haitham and Huang, Andrew and others},
  booktitle={International Conference on Learning Representations},
  volume={2026},
  pages={138846--138923},
  year={2026}
}

@misc{bytedance_seedream5,
  author  = {{ByteDance Seed}},
  title   = {{Seedream 5.0 Lite}},
  year    = {2026},
  howpublished = {\url{https://seed.bytedance.com/en/seedream5_0_lite}},
  note = {Accessed: 2026-08-03}
}

@article{lin2025seeing,
  title={Seeing before reasoning: A unified framework for generalizable and explainable fake image detection},
  author={Lin, Kaiqing and Yan, Zhiyuan and Chen, Ruoxin and Ye, Junyan and Zhang, Ke-Yue and Zhou, Yue and Jin, Peng and Li, Bin and Yao, Taiping and Ding, Shouhong},
  journal={arXiv preprint arXiv:2509.25502},
  year={2025}
}

@inproceedings{kirillov2023segment,
  title={Segment anything},
  author={Kirillov, Alexander and Mintun, Eric and Ravi, Nikhila and Mao, Hanzi and Rolland, Chloe and Gustafson, Laura and Xiao, Tete and Whitehead, Spencer and Berg, Alexander C and Lo, Wan-Yen and others},
  booktitle={2023 IEEE/CVF international conference on computer vision (ICCV)},
  pages={3992--4003},
  year={2023},
  organization={IEEE}
}

@inproceedings{reimers2019sentence,
    title = "Sentence-BERT: Sentence Embeddings using Siamese BERT-Networks",
    author = "Reimers, Nils and Gurevych, Iryna",
    booktitle = "Proceedings of the 2019 Conference on Empirical Methods in Natural Language Processing",
    month = "11",
    year = "2019",
    publisher = "Association for Computational Linguistics",
    url = "http://arxiv.org/abs/1908.10084",
}

@inproceedings{huang2025sida,
  title={Sida: Social media image deepfake detection, localization and explanation with large multimodal model},
  author={Huang, Zhenglin and Hu, Jinwei and Li, Xiangtai and He, Yiwei and Zhao, Xingyu and Peng, Bei and Wu, Baoyuan and Huang, Xiaowei and Cheng, Guangliang},
  booktitle={Proceedings of the Computer Vision and Pattern Recognition Conference},
  pages={28831--28841},
  year={2025}
}

@article{huang2025so,
  title={So-fake: Benchmarking and explaining social media image forgery detection},
  author={Huang, Zhenglin and Li, Tianxiao and Li, Xiangtai and Wen, Haiquan and He, Yiwei and Zhang, Jiangning and Fei, Hao and Yang, Xi and Huang, Xiaowei and Peng, Bei and others},
  journal={arXiv preprint arXiv:2505.18660},
  year={2025}
}

@article{wen2026spot,
  title={Spot the Fake: Large Multimodal Model-Based Synthetic Image Detection with Artifact Explanation},
  author={Wen, Siwei and Ye, Junyan and Feng, Peilin and Kang, Hengrui and Wen, Zichen and Chen, Yize and Wu, Jiang and Wu, Wenjun and He, Conghui and Li, Weijia},
  journal={Advances in Neural Information Processing Systems},
  volume={38},
  pages={58972--59005},
  year={2025}
}

@article{dolhansky2020deepfake,
  title={The deepfake detection challenge (dfdc) dataset},
  author={Dolhansky, Brian and Bitton, Joanna and Pflaum, Ben and Lu, Jikuo and Howes, Russ and Wang, Menglin and Ferrer, Cristian Canton},
  journal={arXiv preprint arXiv:2006.07397},
  year={2020}
}

@inproceedings{ojha2023towards,
  title={Towards universal fake image detectors that generalize across generative models},
  author={Ojha, Utkarsh and Li, Yuheng and Lee, Yong Jae},
  booktitle={2023 IEEE/CVF Conference on Computer Vision and Pattern Recognition (CVPR)},
  pages={24480--24489},
  year={2023},
  organization={IEEE}
}

@inproceedings{zhang2025trade,
  title={Trade-offs in image generation: How do different dimensions interact?},
  author={Zhang, Sicheng and Xie, Binzhu and Yan, Zhonghao and Zhang, Yuli and Zhou, Donghao and Chen, Xiaofei and Qiu, Shi and Liu, Jiaqi and Xie, Guoyang and Lu, Zhichao},
  booktitle={2025 IEEE/CVF International Conference on Computer Vision (ICCV)},
  pages={01--12},
  year={2025},
  organization={IEEE}
}

@article{xue2025ultravideo,
  title={Ultravideo: High-quality uhd video dataset with comprehensive captions},
  author={Xue, Zhucun and Zhang, Jiangning and Hu, Teng and He, Haoyang and Chen, Yinan and Cai, Yuxuan and Wang, Yabiao and Wang, Chengjie and Liu, Yong and Li, Xiangtai and others},
  journal={arXiv preprint arXiv:2506.13691},
  year={2025}
}

@inproceedings{jiang2025vace,
  title={Vace: All-in-one video creation and editing},
  author={Jiang, Zeyinzi and Han, Zhen and Mao, Chaojie and Zhang, Jingfeng and Pan, Yulin and Liu, Yu},
  booktitle={2025 IEEE/CVF International Conference on Computer Vision (ICCV)},
  pages={17191--17202},
  year={2025},
  organization={IEEE}
}

@misc{openai2024videoworldsimulators,
  author       = {OpenAI},
  title        = {Video Generation Models as World Simulators},
  year         = {2024},
  howpublished = {\url{https://openai.com/research/video-generation-models-as-world-simulators}},
  note         = {Accessed: 2026-07-31}
}

@inproceedings{mittal2023video,
  title={Video manipulations beyond faces: A dataset with human-machine analysis},
  author={Mittal, Trisha and Sinha, Ritwik and Swaminathan, Viswanathan and Collomosse, John and Manocha, Dinesh},
  booktitle={Proceedings of the IEEE/CVF winter conference on applications of computer vision},
  pages={643--652},
  year={2023}
}

@inproceedings{yang2026videocof,
  title={VideoCoF: Unified video editing with temporal reasoner},
  author={Yang, Xiangpeng and Xie, Ji and Yang, Yiyuan and Ma, Yue and Huang, Yan and Xu, Min and Wu, Qiang},
  booktitle={Proceedings of the IEEE/CVF Conference on Computer Vision and Pattern Recognition},
  pages={37940--37949},
  year={2026}
}

@inproceedings{nguyen2024videofact,
  title={Videofact: detecting video forgeries using attention, scene context, and forensic traces},
  author={Nguyen, Tai D and Fang, Shengbang and Stamm, Matthew C},
  booktitle={2024 IEEE/CVF Winter Conference on Applications of Computer Vision (WACV)},
  pages={8548--8558},
  year={2024},
  organization={IEEE}
}

@inproceedings{bian2025videopainter,
  title={Videopainter: Any-length video inpainting and editing with plug-and-play context control},
  author={Bian, Yuxuan and Zhang, Zhaoyang and Ju, Xuan and Cao, Mingdeng and Xie, Liangbin and Shan, Ying and Xu, Qiang},
  booktitle={Proceedings of the Special Interest Group on Computer Graphics and Interactive Techniques Conference Conference Papers},
  pages={1--12},
  year={2025}
}

@article{park2025vidguard,
  title={{VidGuard-R1}: {AI}-Generated Video Detection and Explanation via Reasoning {MLLMs} and {RL}},
  author={Park, Kyoungjun and Yang, Yifan and Yi, Juheon and Zheng, Shicheng and Shen, Yifei and Han, Dongqi and Shan, Caihua and Muaz, Muhammad and Qiu, Lili},
  journal={International Conference on Learning Representations},
  year={2026}
}

@article{wang2024vidprom,
  title={Vidprom: A million-scale real prompt-gallery dataset for text-to-video diffusion models},
  author={Wang, Wenhao and Yang, Yi},
  journal={Advances in Neural Information Processing Systems},
  volume={37},
  pages={65618--65642},
  year={2024}
}

@inproceedings{yan2024visa,
  title={Visa: Reasoning video object segmentation via large language models},
  author={Yan, Cilin and Wang, Haochen and Yan, Shilin and Jiang, Xiaolong and Hu, Yao and Kang, Guoliang and Xie, Weidi and Gavves, Efstratios},
  booktitle={European Conference on Computer Vision},
  pages={98--115},
  year={2024},
  organization={Springer}
}

@article{yang2024vript,
  title={Vript: A video is worth thousands of words},
  author={Yang, Dongjie and Huang, Suyuan and Lu, Chengqiang and Han, Xiaodong and Zhang, Haoxin and Gao, Yan and Hu, Yao and Zhao, Hai},
  journal={Advances in Neural Information Processing Systems},
  volume={37},
  pages={57240--57261},
  year={2024}
}

@misc{wavespeed_wan22_video_edit,
  author       = {{WaveSpeedAI}},
  title        = {{Wan 2.2 Video Edit}},
  year         = {2025},
  howpublished = {\url{https://wavespeed.ai/models/wavespeed-ai/wan-2.2/video-edit}},
  note         = {Accessed: 2026-08-03}
}

@article{wan2025wan,
  title={Wan: Open and advanced large-scale video generative models},
  author={Wan, Team and Wang, Ang and Ai, Baole and Wen, Bin and Mao, Chaojie and Xie, Chen-Wei and Chen, Di and Yu, Feiwu and Zhao, Haiming and Yang, Jianxiao and others},
  journal={arXiv preprint arXiv:2503.20314},
  year={2025}
}
}

\clearpage

\clearpage
\onecolumn
\setcounter{page}{1}
\renewcommand{\thepage}{\arabic{page}}
\begin{center}
    \Large
    \textbf{\thetitle}\\[0.5em]
    Supplementary Material
\end{center}
\appendix
\captionsetup[table]{position=top,skip=2pt}
\setcounter{figure}{0}
\setcounter{table}{0}
\setcounter{equation}{0}
\renewcommand{\theHfigure}{supp.\arabic{figure}}
\renewcommand{\theHtable}{supp.\arabic{table}}
\renewcommand{\theHequation}{supp.\arabic{equation}}
\renewcommand{\thefigure}{S\arabic{figure}}
\renewcommand{\thetable}{S\arabic{table}}
\renewcommand{\theequation}{S\arabic{equation}}

\raggedbottom


\FloatBarrier

\FloatBarrier
\section{Motivation}
\label{app:motivation}
\FloatBarrier
\subsection{Scope Comparison: Full Synthesis vs. Manipulation}
\noindent \textbf{Full Synthesis.}
Fully synthetic videos are generated without reference to a source video, typically using existing T2V or I2V models. Under this scope, real and fake videos may differ substantially in both semantic and visual content, while basic video attributes such as resolution, frame rate, and duration are often unaligned. Consequently, they cannot form semantically and visually aligned real-fake video pairs. A common practice in existing dataset construction is to directly collect real videos from datasets such as Vript~\citep{yang2024vript} or Kinetics-400~\citep{carreira2017quo}, while generating fake videos separately from additional text prompts or reference images. For example, VidProM~\citep{wang2024vidprom} employs a large collection of user-provided T2V prompts. Such discrepancies in content and video attributes can introduce significant shortcuts for forgery detection. 
Unfortunately, existing video forgery datasets, such as the widely used GenVideo~\citep{chen2026demamba} and GenVideoBench~\citep{ni2026genvidbench}, predominantly focus on full synthesis and thus inevitably suffer from the aforementioned limitations.
\textbf{\textit{\textcolor{Red}{VidGuard-R1~\citep{park2025vidguard} claims that exploiting these shortcuts can yield over 96\% accuracy on GenVideo and GenVideoBench.}}}
Although VidGuard-R1 recognizes this issue and attempts to better align fake videos with their source videos by preserving the first frame, its subsequent I2V generation lacks fine-grained geometric control, such as Canny guidance, and still relies primarily on text prompts. As a result, the generated videos may still deviate from their source videos in temporal dynamics, such as the speed at which events evolve over time, making precise source-target alignment difficult to achieve.

\noindent \textbf{Manipulation.}
Manipulated videos are generated with explicit and complete source videos as references, typically through V2V/I2V models or manual editing. Unlike fully synthetic videos, manipulation modifies only localized regions, naturally preserving the overall semantic and visual content of the source videos while maintaining consistent basic video attributes. This alignment enables detectors to focus purely on spatial-temporal forgery artifacts rather than exploiting content or attribute shortcuts. In our data construction pipeline, the indirect I2V paradigm further introduces video-level Canny guidance during I2V generation after editing the first frame, precisely preserving the temporal evolution of unmanipulated regions in the source video. This produces realistic, high-quality manipulated videos and poses a substantially more challenging task for conventional forgery detection methods. 
Tab~\ref{tab:scope_comparison} illustrates the differences between these two scopes.

\vspace{2mm}
\begin{table*}[htb]
\centering
\vspace{-4mm}
\caption{\textbf{Scope Comparison between Full Synthesis and Manipulation.}}

\renewcommand{\arraystretch}{1.1}
\resizebox{\textwidth}{!}{
\begin{tabular}{lcc}
\toprule[1.5pt]
\textbf{Dimension} 
& \textbf{Full Synthesis} 
& \textbf{Manipulation} \\
\hline

\textbf{Source Video Reference}
& None or partial
& Complete \\

\textbf{Generation Paradigm}
& Direct T2V / I2V
& V2V / Conditional I2V  \\

\textbf{Modification Scope}
& Entire video
& Localized regions \\

\textbf{Real--Fake Video Pair}
& Unavailable
& Naturally paired \\

\textbf{Semantic Alignment}
& Low
& High \\

\textbf{Visual Alignment}
& Low
& High \\

\textbf{Temporal Alignment}
& Low
& High \\

\textbf{Shortcut Risk}
& High
& Low \\

\textbf{Detection Evidence}
& Global generation cues / Potential shortcuts
& Local spatial-temporal forgery artifacts \\

\textbf{Forensic Challenge}
& Distinguishing generated from real videos
& Identifying subtle localized inconsistencies \\

\bottomrule[1.5pt]
\end{tabular}}
\vspace{-2mm}
\label{tab:scope_comparison}
\end{table*}

\FloatBarrier
\subsection{Manipulation Dataset Limitation}
\label{app:motivation:limitation}
As noted in~\ref{dataset:motivation}, existing manipulation datasets suffer from three key limitations. \textbf{(1) Limited Diversity.} Existing datasets~\citep{rossler2019faceforensics++, dolhansky2020deepfake} are often restricted to specific domains and manipulation tasks, particularly face-related scenarios, which directly limits their applicability and potential in general-world scenarios. \textbf{(2) Closed-Vocabulary Instructions.} This limitation is common in existing image manipulation datasets. For instance, SIDA~\citep{huang2025sida} predefines a fixed set of target concepts for specific source objects. For \textit{Bird}, the replacement set is predefined as \{\textit{Owl, Duck, Chicken, Goose, Parrot, Eagle, Sparrow, Penguin}\}. Such a closed-vocabulary design introduces two major issues. \textbf{\textcolor{Red}{\textit{1) Homogeneity.}}} a fixed target set leads to repetitive manipulation patterns, substantially restricting the flexibility and diversity of manipulation operations. \textbf{\textit{\textcolor{Red}{2) Logical Inconsistency.}}} Although SIDA partially considers semantic similarity when constructing target sets to reduce potential conflicts, category-level similarity does not guarantee contextual or logical consistency. 
For example, consider a bird resting on a blooming grassland in spring. If the bird is selected as the source object, its candidate replacements may all belong to the same semantic category, \textit{Aves}. However, replacing it with a penguin would clearly violate common sense, as penguins are naturally adapted to cold environments rather than such a warm spring setting.
In contrast, we employs an MLLM assistant to jointly reason over the video content and its detailed caption to generate reasonable open-vocabulary instruction proposals. This design preserves manipulation diversity while ensuring that the proposed manipulations remain contextually and logically plausible.
\textbf{(3) Insufficient Annotations.} As shown in Tab.~\ref{tab:dataset_comparison}, existing video manipulation datasets typically provide only binary authenticity labels. VideoSham~\citep{mittal2023video} additionally releases manipulation instructions, but still lacks corresponding forgery masks and anomaly explanations, which are essential for explainable forensic analysis.

\FloatBarrier
\section{ManiVid-38K Dataset}
\label{app:manivid_dataset}

\FloatBarrier
\subsection{Curation Pipeline}
\label{app:curation_pipeline}

\FloatBarrier
\subsubsection{Source Video Collection}
We select three public datasets, OpenVid-1M~\citep{nan2025openvid}, InternVid~\citep{wang2024internvid}, and OpenVE-3M~\citep{he2025openve}, as sources of real videos. The processing procedure and authenticity assurance for each dataset are detailed below.

\noindent \textbf{OpenVid-1M.}
OpenVid-1M contains approximately 1M video-text pairs, including a high-resolution subset, OpenVidHD, with about 0.4M 1080P videos. Given its high video quality, we select OpenVidHD as the primary source of real videos for ManiVid-38K. OpenVidHD consists of videos from four sources: \textbf{(1) ChronoMagic}~\citep{yuan2024chronomagic}, \textbf{(2) CelebvHQ}~\citep{zhu2022celebv}, \textbf{(3) Open-Sora-Plan}~\citep{pku_yuan_lab_2024_opensoraplan}, and \textbf{(4) Panda}~\citep{chen2024panda70m}. CelebvHQ predates the release of Sora~\citep{openai2024videoworldsimulators} in February 2024, while Open-Sora-Plan and Panda explicitly identifies its videos as non-AI-generated. In contrast, we exclude ChronoMagic~(was released in April 2024) because its videos were collected from the Internet after the release of Sora, potentially introducing the risk of AI-generated content. Another advantage of OpenVidHD is its quality-related metadata, such as \textit{Aesthetics Score}. We randomly sample a subset and prioritize videos with higher \textit{Aesthetics Scores} to maximize the quality of the selected videos.

\noindent \textbf{InternVid.}
InternVid contains 7.1M 720P videos accompanied by brief text captions. It was first released in 2023, with videos sourced from YouTube, providing reasonable assurance of their authenticity. However, given its lower resolution compared with OpenVidHD and the additional effort required to download videos from YouTube, we sample only a small subset as source videos.

\noindent \textbf{OpenVE-3M.}
OpenVE-3M is a recent video editing dataset containing 3M source-target video pairs across diverse manipulation tasks. However, our comprehensive examination reveals three systematic issues: \textbf{(1) Source Authenticity.} OpenVE-3M collects 1M videos from Open-Sora-Plan~\citep{pku_yuan_lab_2024_opensoraplan}, OpenViDHD~\citep{nan2025openvid}, and UltraVideo~\citep{xue2025ultravideo} as its base video library. Part of UltraVideo is collected from YouTube in 2025 and may therefore be exposed to potential contamination by AI-generated content. \textbf{(2) Reverse Editing.} OpenVE-3M constructs reverse editing samples by treating generated edited videos as sources and the original videos as targets, accompanied by corresponding reverse editing instructions. Consequently, videos appearing on the source side cannot be reliably assumed to be authentic. \textbf{(3) Manipulation Plausibility.} We further observe some implausible manipulation instructions. For example, changing a young boy's hair to the gray-white hair of an elderly person can produce an obviously unrealistic result that is easily identifiable by human inspection.
Overall, OpenVE-3M is primarily developed from a video generation perspective, emphasizing generation quality rather than source authenticity. From a forgery detection perspective, however, its use requires additional caution. We therefore sample only a very small subset of source videos and manually inspect every selected sample to ensure that no noticeable manipulation or AI-generated artifacts are present.

\noindent \textbf{Deduplication \& Quality Filter.} 
We construct a source video library from the three datasets above and further review and filter the collected videos to ensure overall data quality.
The filenames provided by the source datasets allow us to identify clips originating from the same source video.
We therefore perform source-level deduplication based on these identifiers and retain at most one clip from each original source video.
This prevents highly similar or temporally adjacent clips from the same source video from appearing multiple times in ManiVid-38K.
We then apply quality filtering to the remaining videos and retain only those that simultaneously satisfy three criteria:
\textbf{\textit{\textcolor{Red}{1) resolution of at least 720p}}},
\textbf{\textit{\textcolor{Red}{2) frame rate of at least 15 FPS}}}, and
\textbf{\textit{\textcolor{Red}{3) duration no longer than 10 seconds}}}.
Finally, 41,108 source videos are retained in our source video library, as shown in Tab.~\ref{app:tab:dataset_collection}.

\begin{table}[!t]
    \centering
    \caption{\textbf{Data Statistics During Collection.} The number of source videos retained at different stages of the collection process.} 
    \vspace{1mm}
    \definecolor{mygray}{gray}{.92}
    \renewcommand{\arraystretch}{1.2}
    \belowrulesep=-0.25pt
    \aboverulesep=-0.25pt
    \resizebox{0.65\linewidth}{!}{
        \begin{tabular}{c|ccc|c}
        \toprule[1.5pt]
        \textbf{Source Dataset} & \textbf{OpenVid-1M} & \textbf{InternVid} & \textbf{OpenVE-3M} & \textbf{Total} \\
        \hline
        Sampled Num & 222391 & 5882 & 15000 & 243273 \\
        + Deduplication & 139697 & 3974 & 9292 & 152963 \\
        + Quality Filter & 35643 & 2959 & 2506 & 41108 \\
        \bottomrule[1.5pt]
        \end{tabular}
    }
    \vspace{-3mm}
    \label{app:tab:dataset_collection}
    \end{table}

\FloatBarrier
\subsubsection{Content Categorization}
To ensure the content diversity of ManiVid-38K dataset, we categorize the samples in the source video library according to their video content. We define six valid content categories and one invalid category (\textbf{\textit{Uncertain}}), as detailed in Tab.~\ref{app:tab:dataset_content_definition}. We then employ the MLLM assistant to categorize each source video based on the video input (\textbf{Required}) and its caption (\textbf{Optional}).  
Tab.~\ref{app:tab:dataset_content_prompt} presents the categorization prompt. Finally, we adopt balanced sampling to maintain a uniform number of samples across categories.
Fig.~\ref{fig:teaser} shows the final content distribution of our ManiVid-38K dataset.

\begin{table}[!htbp]
\centering
\footnotesize
\vspace{-2mm}
\caption{\textbf{Content Category Definitions and Examples.}}
\vspace{1mm}

\definecolor{mygray}{gray}{.92}
\renewcommand{\arraystretch}{1.25}
\belowrulesep=-0.25pt
\aboverulesep=-0.25pt
\resizebox{1.0\linewidth}{!}{
    \begin{tabular}{
    >{\centering\arraybackslash}m{2.5cm}|
    m{11cm}|
    >{\centering\arraybackslash}m{3cm}
    }
    \toprule[1.5pt]
    \textbf{Content Category} & \multicolumn{1}{c|}{\textbf{Definition}} & \textbf{Examples} \\
    \hline
    \textit{Human} &
    The video is centered on one or more humans. The person(s) usually occupy a large portion of the frame or are the clear narrative focus. The background is typically simple and does not form a complete environment. 
    \newline Notes: 
    \newline (1) If humans appear but are small and the environment is the main focus, consider \textit{Indoor} or \textit{Outdoor} instead.
    \newline (2) If the video mainly conveys a time-continuous activity or process where ``what is happening" is more important than ``who the person is", consider \textit{Event}.
    &
    \textit{Reporter, Police Officer, Businessman, Farmer, ...} \\
    \hline
    \textit{Animal} &
    The video is centered on one or more animals, which occupy a large portion of the frame or are the narrative focus.
    \newline Notes: 
    \newline (1) If animals are small or only decorative and the environment dominates, use \textit{Indoor} or \textit{Outdoor}.
    \newline (2) If the video mainly presents a continuous activity performed by the animals, consider \textit{Event}.
    &
    \textit{Cat, Dog, Bird, Horse, Cow, Sheep, ...} \\
    \hline
    \textit{Object} &
    The video focuses on one or a few inanimate objects. The background is simple and does not form a complete environment.
    \newline Notes: 
    \newline (1) Hands or partial human bodies may appear, but the object itself is the core subject.
    \newline (2) If the video emphasizes a multi-step, time-continuous process, consider \textit{Event}.
    &
    \textit{Knife, Fork, Food, Tool, Cup, Toy, Laptop, ...} \\
    \hline
    \textit{Indoor} &
    The video primarily presents a complete and visually rich indoor environment (e.g., rooms, offices, shops). No single subject dominates, or subjects are small and secondary. 
    \newline Notes: 
    \newline (1) Use this when the spatial layout, interior structure, and overall atmosphere are the main information.
    &
    \textit{Room, Office, Shop, Gym, Hospital, ...} \\
    \hline
    \textit{Outdoor} &
    The video primarily presents a complete and visually rich outdoor environment or scenery (e.g., natural landscapes, city streets, building exteriors, aerial views). No single subject dominates, or subjects are small and secondary.
    \newline Notes: 
    \newline (1) Even if people or animals appear, choose \textit{Outdoor} as long as they are minor and the scenery is the main focus.
    &
    \textit{Natural Landscape, City Street, Aerial View, ...} \\
    \hline
    \textit{Event} &
    The video is centered on a time-continuous activity or event. The key information lies in what is happening over time rather than the appearance of a specific person, animal, object, or environment.
    \newline Notes: 
    \newline (1) It includes any multi-step actions with clear temporal progression.
    \newline (2) The main subject may be human, animal, or object, but the activity itself is the narrative core.
    &
    \textit{Cooking Process, Ceremony, Sports Game, ...} \\
    \hline
    \textit{Uncertain} &
    The video does not fit any of the above categories, or the quality/content is insufficient (blurred, too short, occluded, missing key information) to make a reliable judgment.
    &
    - \\
    \bottomrule[1.5pt]
    \end{tabular}
}
\vspace{-4mm}
\label{app:tab:dataset_content_definition}
\end{table}
\begin{table*}[htb]\centering
    \caption{\textbf{Content Categorization Prompt.}}
    \vspace{1mm}
    \begin{minipage}{\textwidth}\vspace{0mm}    \centering
    \begin{tcolorbox} 
    \centering
    \fontsize{8pt}{10pt}\selectfont
    \begin{tabular}{p{\dimexpr\linewidth-4\tabcolsep\relax} c}
    \VarSty{ {\bf \normalsize Content Categorization Prompt} } &\\

    \vspace{0.1\baselineskip}

    {\small\bfseries Task Description:}

    \vspace{0.1\baselineskip}
    
    You are given a \textbf{video} and its \textbf{original caption}: \texttt{\textless ORIGINAL\_CAPTION\textgreater}.
    
    
    Your task is to classify the video into \textbf{exactly one} of the following categories based on the definitions below. If none of the categories apply, or the video quality/content is insufficient for a confident decision, choose category \textbf{\textit{G}}.
    
    \vspace{0.5\baselineskip}
    
    {\small\bfseries Category Definitions:}

    \vspace{0.3\baselineskip}
    
    \textbf{\textit{A. Human}}

    
    \textbf{Definition:}
    The video is centered on one or more humans. The person(s) usually occupy a large portion of the frame or are the clear narrative focus. The background is typically simple and does not form a complete environment.


    \textbf{Notes:}
        \begin{enumerate}[leftmargin=2em, labelsep=0.5em, itemsep=-2pt, topsep=-2pt]
            \item If humans appear but are small and the environment is the main focus, consider \textbf{\textit{Indoor}} or \textbf{\textit{Outdoor}} instead.
            \item If the video mainly conveys a time-continuous activity or process where ``what is happening" is more important than ``who the person is", consider \textbf{\textit{Event}}.
        \end{enumerate}

    \vspace{0.5\baselineskip}

    \textbf{\textit{B. Animal}}

    
    \textbf{Definition:} 
    The video is centered on one or more animals, which occupy a large portion of the frame or are the narrative focus.
    
    
    \textbf{Notes:}
        \begin{enumerate}[leftmargin=2em, labelsep=0.5em, itemsep=-2pt, topsep=-2pt]
            \item If animals are small or only decorative and the environment dominates, use \textbf{\textit{Indoor}} or \textbf{\textit{Outdoor}}.
            \item If the video mainly presents a continuous activity performed by the animals, consider \textbf{\textit{Event}}.
        \end{enumerate}

    \vspace{0.5\baselineskip}
    
    \textbf{\textit{C. Object}}
    
    
    \textbf{Definition:}
    The video focuses on one or a few inanimate objects (\emph{e.g.}, products, tools, food). The background is simple and does not form a complete environment.
    
    
    \textbf{Notes:}
        \begin{enumerate}[leftmargin=2em, labelsep=0.5em, itemsep=-2pt, topsep=-2pt]
            \item Hands or partial human bodies may appear, but the object itself is the core subject.
            \item If the video emphasizes a multi-step, time-continuous process (\emph{e.g.}, full cooking or repair workflow), consider \textbf{\textit{Event}}.
        \end{enumerate}

    \vspace{0.5\baselineskip}
    
    \textbf{\textit{D. Indoor}}
    
    
    \textbf{Definition:}
    The video primarily presents a complete and visually rich indoor environment (\emph{e.g.}, rooms, offices, shops). No single subject dominates, or subjects are small and secondary.
    
    
    \textbf{Notes:}
    Use this when the spatial layout, interior structure, and overall atmosphere are the main information.

    \vspace{0.5\baselineskip}
    
    \textbf{\textit{E. Outdoor}}
    
    
    \textbf{Definition:}
    The video primarily presents a complete and visually rich outdoor environment or scenery (\emph{e.g.}, natural landscapes, city streets, building exteriors, aerial views). No single subject dominates, or subjects are small and secondary.
    
    
    \textbf{Notes:}
    Even if people or animals appear, choose \textbf{\textit{Outdoor}} as long as they are minor and the scenery is the main focus.

    \vspace{0.5\baselineskip}
    
    \textbf{\textit{F. Event}}
    
    
    \textbf{Definition:}
    The video is centered on a time-continuous activity or event. The key information lies in what is happening over time rather than the appearance of a specific person, animal, object, or environment.
    
    
    \textbf{Notes:}
        \begin{enumerate}[leftmargin=2em, labelsep=0.5em, itemsep=-2pt, topsep=-2pt]
            \item Typical examples include sports games, cooking processes, performances, ceremonies, construction or repair workflows, and any multi-step actions with clear temporal progression.
            \item The main subject may be human, animal, or object, but the activity itself is the narrative core.
        \end{enumerate}

    \vspace{0.5\baselineskip}
    
    \textbf{\textit{G. Uncertain}}
    
    
    \textbf{Definition:}
    The video does not fit any of the above categories, or the quality/content is insufficient (blurred, too short, occluded, missing key information) to make a reliable judgment.

    \vspace{0.5\baselineskip}
    
    {\small\bfseries Requirements:}
        \begin{enumerate}[leftmargin=2em, labelsep=0.5em, itemsep=-2pt, topsep=2pt]
            \item Output a single JSON object and nothing else, with no explanation, no chain of thought, and no markdown fences: \{``category": \texttt{\textless CategoryLetter\textgreater}\}
            \item The \texttt{\textless CategoryLetter\textgreater} must be exactly one of: \textbf{\textit{A}} / \textbf{\textit{B}} / \textbf{\textit{C}} / \textbf{\textit{D}} / \textbf{\textit{E}} / \textbf{\textit{F}} / \textbf{\textit{G}}.
        \end{enumerate}

    \end{tabular}
    \end{tcolorbox}
    \vspace{-3mm}
    \label{app:tab:dataset_content_prompt}
    \end{minipage}
\end{table*}

\FloatBarrier
\subsubsection{Detailed Recaption}
Since OpenVE-3M does not provide captions for its videos, while the other two data sources, particularly InternVid, contain only extremely brief descriptions (with an average length of 17.6), the available captions are semantically sparse and often omit many potential objects, as illustrated in Fig.~\ref{fig:dataset_main}. Therefore, we employ an MLLM assistant to uniformly recaption all samples, enriching the textual information with \textbf{\textit{precise attributes}}, \textbf{\textit{added objects}}, \textbf{\textit{spatial details}}, and \textbf{\textit{event cues}}, which provides a foundation for subsequent manipulation source object selection, instruction generation, and feasibility evaluation. Tab.~\ref{app:tab:dataset_recaption_prompt} provides the prompt used for recaptioning, and Fig.~\ref{fig:teaser} shows the caption length distribution of our ManiVid-38K dataset.

\begin{table*}[htb]\centering
    \caption{\textbf{Video Recaption Prompt.}}
    \vspace{1mm}
    \begin{minipage}{\textwidth}\vspace{0mm}    \centering
    \begin{tcolorbox} 
    \centering
    \fontsize{8pt}{10pt}\selectfont
    \begin{tabular}{p{\dimexpr\linewidth-4\tabcolsep\relax} c}
    \VarSty{ {\bf \normalsize Video Recaption Prompt} } &\\
    
    \vspace{0.1\baselineskip}

    {\small\bfseries Task Description:}

    \vspace{0.1\baselineskip}
    
    You are given a \textbf{video} and its \textbf{original caption}: \texttt{\textless ORIGINAL\_CAPTION\textgreater}.
    
    
    Your task is to generate a \textbf{detailed caption} by refining and expanding the original caption with the visual content of the video. You are allowed to \textbf{correct obvious mistakes} in the original caption if they conflict with the video.
    
    \vspace{0.5\baselineskip}
    
    {\small\bfseries Requirements:}
        \begin{enumerate}[leftmargin=2em, labelsep=0.5em, itemsep=-2pt, topsep=2pt]
             \item Output a single JSON object and nothing else, with no explanation, no chain of thought, and no markdown fences: \{``refined\_caption": \texttt{\textless Detailed Caption\textgreater}\}
            \item The \texttt{\textless Detailed Caption\textgreater} should refine and enrich the original caption, may correct obvious errors, and should describe the main subjects, environment, any notable temporal changes and detailed object attributes.
            \item The \texttt{\textless Detailed Caption\textgreater} must be objective and factual. Avoid subjective opinions, emotional language, or rhetorical devices such as metaphors or similes.
        \end{enumerate}
    
    \end{tabular}
    \end{tcolorbox}
    \vspace{-3mm}
    \label{app:tab:dataset_recaption_prompt}
    \end{minipage}
\end{table*}

\FloatBarrier
\subsubsection{Instruction Generation}

\noindent \textbf{Task Definition.}
We select six common manipulation tasks for data construction, including \textbf{\textit{Addition}}, \textbf{\textit{Removal}}, \textbf{\textit{Replacement}}, \textbf{\textit{Color Alteration}}, \textbf{\textit{Material Modification}}, and \textbf{\textit{Portrait Editing}}. The definition and examples of each manipulation task are provided in Tab.~\ref{app:tab:manipulation_task_definition}.

\begin{table}[!htbp]
\centering
\footnotesize
\vspace{-4mm}
\caption{\textbf{Manipulation Task Definitions and Examples.}}
\vspace{1mm}
\definecolor{mygray}{gray}{.92}
\renewcommand{\arraystretch}{1.3}
\belowrulesep=-0.25pt
\aboverulesep=-0.25pt
\resizebox{1.0\linewidth}{!}{
    \begin{tabular}{
    >{\centering\arraybackslash}m{3.2cm}|
    m{7.5cm}|
    m{6.0cm}
    }
    \toprule[1.5pt]
    \textbf{Manipulation Task} &
    \multicolumn{1}{c|}{\textbf{Definition}} &
    \multicolumn{1}{c}{\textbf{Example}} \\
    \hline
    
    \textit{Addition} &
    Integrate a new, semantically relevant object into the scene. &
    \textit{Add a red backpack onto the empty bench in \texttt{<video>}.} \\
    \hline
    
    \textit{Removal} &
    Erase a specific object with clear boundaries and inpaint the background naturally. &
    \textit{Remove the blue suitcase from \texttt{<video>}, seamlessly inpainting the revealed pavement.} \\
    \hline
    
    \textit{Replacement} &
    Swap an object with a fine-grained variant or a logically related entity. &
    \textit{Replace the white sedan in \texttt{<video>} with a red SUV, preserving the original motion trajectory.} \\
    \hline
    
    \textit{Color Alteration} &
    Change the color of a specific object while preserving its original texture and shape. &
    \textit{Change the color of the woman's jacket in \texttt{<video>} to blue, maintaining its original fabric texture.} \\
    \hline
    
    \textit{Material Modification} &
    Alter the physical material properties. &
    \textit{Change the material of the rotating vase in \texttt{<video>} into polished metal, maintaining its original motion.} \\
    \hline
    
    \textit{Portrait Editing} &
    Modify facial features, expressions, or skin tones. (\textbf{Note}: Only applicable to close-up or medium shots of people). &
    \textit{Modify the facial features of the woman at the center of \texttt{<video>} to exhibit a smiling expression, keeping others consistent.} \\
    
    \bottomrule[1.5pt]
    \end{tabular}
}
\vspace{-3mm}
\label{app:tab:manipulation_task_definition}
\end{table}

\noindent \textbf{Task Initialization.}
We randomly assign an initial manipulation task to each source video, as we find that allowing the MLLM assistant to directly select a suitable task based on the video content leads to a severely imbalanced task distribution. Specifically, the MLLM tends to favor simpler tasks such as \textbf{\textit{Addition}} and \textbf{\textit{Removal}}, while being less inclined to attempt more challenging tasks such as \textbf{\textit{Material Modification}} and \textbf{\textit{Portrait Editing}}.

\noindent \textbf{Idea Inspiration.}
We instruct the MLLM assistant to brainstorm manipulation ideas based on the source video, the initially assigned task, and the detailed recaption. Since the initial task is assigned completely at random, two cases may arise: \textbf{(1) The assigned task is suitable for the source video.} In this case, the MLLM identifies a potential source object suitable for the manipulation task and leverages its prior knowledge to propose a reasonable target. \textbf{(2) The assigned task is unsuitable for the source video.} For example, a landscape video may be assigned the \textbf{\textit{Portrait Editing}} task. In this case, the MLLM summarizes the reason for the failure and switches to a manipulation task that it considers suitable until case \textbf{(1)} is reached. After generating the manipulation instruction, we further require the assistant to perform reflection to ensure that the instruction conforms to physical principles and event logic, and to provide a justification for the reasonableness of the instruction in its output, thereby avoiding unreasonable instructions such as those illustrated in~\ref{app:motivation:limitation}.
Tab.~\ref{app:tab:dataset_idea_prompt} provides the prompt used for manipulation instruction generation, and Fig.~\ref{fig:teaser} shows the final distribution of manipulation tasks in ManiVid-38K dataset.

\begingroup
\vspace{2mm}
\captionsetup{hypcap=false}
\captionof{table}{\textbf{Idea Inspiration Prompt.}}
\vspace{-2mm}
\begin{tcolorbox}[
    breakable,
    fontupper=\fontsize{8pt}{10pt}\selectfont,
    before upper={\setlength{\parindent}{0pt}\setlength{\parskip}{0pt}}
]
    \VarSty{ {\bf \normalsize Idea Inspiration Prompt} }\par

    \vspace{0.1\baselineskip}

    {\small\bfseries Task Description:}
    
    \vspace{0.1\baselineskip}
    
    You are an expert in video content editing and forensic analysis. Your task is to design a \textbf{high-quality, photorealistic video manipulation scheme} based on a source video, its detailed caption and an initial manipulation task. This scheme will be used to test the robustness of manipulation detection approaches.
    
    \vspace{0.3\baselineskip}
    
    {\small\bfseries Input Data:}
    
    
    \begin{enumerate}[leftmargin=2em, labelsep=0.5em, topsep=-2pt, itemsep=-2pt]
        \item \textbf{Source Video}: \texttt{<video>}
        \item \textbf{Detailed Caption}: \texttt{<detailed caption>}
        \item \textbf{Initial Manipulation Task}: \texttt{<initial manipulation task>}
    \end{enumerate}
    
    \vspace{0.5\baselineskip}
    
    {\small\bfseries Manipulation Task Definitions:}
    
    
    \begin{itemize}[leftmargin=2em, labelsep=0.5em, topsep=-2pt, itemsep=-2pt]
        \item \textbf{\textit{Addition}}: Integrate a new, semantically relevant object into the scene.
        \item \textbf{\textit{Removal}}: Erase a specific object with clear boundaries and inpaint the background naturally.
        \item \textbf{\textit{Replacement}}: Swap an object with a fine-grained variant or a logically related entity.
        \item \textbf{\textit{Color Alteration}}: Change the color of a specific object while preserving its original texture and shape.
        \item \textbf{\textit{Material Modification}}: Alter the physical material properties.
        \item \textbf{\textit{Portrait Editing}}: Modify facial features, expressions, or skin tones. (\textbf{Note}: Only applicable to close-up or medium shots of people).
    \end{itemize}
    
    \vspace{0.5\baselineskip}
    
    {\small\bfseries Idea Inspiration Principles:}
    
    \vspace{0.1\baselineskip}
    
    Your proposal must explicitly adhere to these four criteria:
    
    \begin{itemize}[leftmargin=2em, labelsep=0.5em, topsep=-2pt, itemsep=-2pt]
        \item \textbf{Logical Consistency}: The modification must follow the laws of physics, common sense, and maintain consistent lighting and occlusion.
        \item \textbf{Diversity}: Avoid repetitive or generic modifications. Use specific, varied nouns.
        \item \textbf{Contour Clarity (Segmentability)}: Target objects must have well-defined edges. Avoid ambiguous or amorphous entities like smoke, fire, or flowing water.
        \item \textbf{Spatio-Temporal Uniqueness}: Define the target entity with absolute precision across both the spatial layout and the temporal timeline to guarantee zero tracking ambiguity across consecutive frames. The target must be uniquely identifiable not just in a single isolated frame, but continuously throughout the entire video sequence.
    \end{itemize}
    
    \vspace{0.5\baselineskip}
    
    {\small\bfseries Task Feasibility Check:}
    
    \vspace{0.1\baselineskip}
    
    Evaluate whether the \textbf{Initial Manipulation Task} is feasible for the video content. If the assigned task is incompatible (\emph{e.g.}, \textbf{\textit{Portrait Editing}} for a landscape video), you \textbf{MUST} first summarize the reason for its infeasibility, then switch to another task and repeat this process until a reasonable and feasible task is selected.
    
    \vspace{0.5\baselineskip}
    
    {\small\bfseries Requirements:}
    
    \vspace{0.1\baselineskip}
    
    The output \textbf{MUST} consist of exactly 4 sections separated by the \texttt{<sep>} token. Do \textbf{not} include any introductory remarks or additional explanations. Do \textbf{not} include the name of each sections.
    
    \begin{enumerate}[leftmargin=2em, labelsep=0.5em, topsep=-2pt, itemsep=-2pt]
        \item \textbf{Change Task}: Output only one word, \texttt{YES} if you changed the assigned task type; otherwise, output \texttt{NO}.
        
        \item \textbf{Final Task}: Return exactly one of the 6 predefined task names that strictly corresponds to your actual, finalized manipulation instruction.
        
        \item \textbf{Design Justification}: Think step-by-step to draft the manipulation scheme. You \textbf{MUST} explicitly evaluate your proposed idea against inspiration principals before finalizing the object and instruction, and output a brief professional explanation.
        
        \item \textbf{Manipulation Instruction}:     
        Formulate the instruction with high information density. While the target entity must be precisely defined, you \textbf{MUST NOT} include irrelevant background context, atmospheric narrative, or edge concepts. 
        Optimize your phrasing for a downstream video generation model. Use the templates below just as your structural guide:
        
        \begin{itemize}[leftmargin=2em, labelsep=0.5em, topsep=-2pt, itemsep=-2pt]
            \item \textbf{\textit{Addition}}: Add [New Object] into \texttt{<video>}. Need to specify a natural location without forcing complex spatial relationships.
            \item \textbf{\textit{Removal}}: Remove [Original Object] from \texttt{<video>}, seamlessly inpainting the revealed [Background].
            \item \textbf{\textit{Replacement}}: Replace [Original Object] in \texttt{<video>} with [Target Object], preserving the original [Motion Trajectory].
            \item \textbf{\textit{Color Alteration}}: Change the color of [Original Object] in \texttt{<video>} to [New Color], maintaining its original [Specific Texture].
            \item \textbf{\textit{Material Modification}}: Change the material of [Original Object] in \texttt{<video>} into [New Material], maintaining its original [Motion].
            \item \textbf{\textit{Portrait Editing}}: Modify the facial features of [Human] in \texttt{<video>} to exhibit [New Expression/Identity], keeping others consistent.
        \end{itemize}
    \end{enumerate}
    
\end{tcolorbox}
\nopagebreak[4]
\label{app:tab:dataset_idea_prompt}
\endgroup

\FloatBarrier
\subsubsection{Manipulation Setup}

\noindent \textbf{Manipulation Paradigms.}
With the rapid maturation of generation toolchains, we adopt two complementary manipulation paradigms to better cover diverse real-world editing operations. In addition to the conventional \textbf{Direct V2V} paradigm, we introduce an \textbf{Indirect I2V} paradigm that combines first-frame editing with controllable I2V generation. Specifically, we assign 53.8\% of the source videos to the V2V paradigm and 46.2\% to the I2V paradigm.

\noindent \textbf{Model Allocation.}
Because generative models exhibit nonuniform trade-offs across quality, alignment, robustness, and linguistic conditions~\citep{zhang2025trade,zhang2026limitations}, manipulation models are randomly assigned within both paradigms to avoid model-selection bias. For \textbf{Direct V2V}, each source video is randomly assigned to one of the seven video editing models. For \textbf{Indirect I2V}, model allocation is performed independently at both stages: an image editing model is randomly selected from six models for first-frame editing, followed by a randomly selected Canny-guided I2V model from three models for final video generation. 
Tab.~\ref{app:tab:model_allocation} and Fig.~\ref{fig:teaser} show the final model allocation in our ManiVid-38K dataset.

\begingroup
\setlength{\tabcolsep}{4pt}
\renewcommand{\arraystretch}{1.3}
\belowrulesep=-0.25pt
\aboverulesep=-0.25pt
\setlength{\LTcapwidth}{\linewidth}

\newlength{\ManiVidAllocationWidth}
\setlength{\ManiVidAllocationWidth}{
    \dimexpr\linewidth-8\tabcolsep-3\arrayrulewidth\relax
}
\begin{longtable}{
    >{\fontsize{7pt}{8.5pt}\selectfont\centering\arraybackslash}m{0.174\ManiVidAllocationWidth}|
    >{\fontsize{7pt}{8.5pt}\selectfont\centering\arraybackslash}m{0.248\ManiVidAllocationWidth}|
    >{\fontsize{7pt}{8.5pt}\selectfont\centering\arraybackslash}m{0.280\ManiVidAllocationWidth}|
    >{\fontsize{7pt}{8.5pt}\selectfont\centering\arraybackslash}m{0.298\ManiVidAllocationWidth}
}

\caption{\textbf{Detailed Distribution of Manipulation Paradigms and Models in ManiVid-38K.}}
\vspace{1mm}
\label{app:tab:model_allocation}\\

\toprule[1.5pt]

\textbf{Paradigm} &
\textbf{V2V Model} &
\textbf{(Stage 1) Image Editing Model} &
\textbf{(Stage 2) Canny-guided I2V Model}
\\

\hline

\endfirsthead

\multicolumn{4}{c}{\footnotesize Table \thetable{} (continued)}
\\

\toprule[1.5pt]

\textbf{Paradigm} &
\textbf{V2V Model} &
\textbf{(Stage 1) Image Editing Model} &
\textbf{(Stage 2) Canny-guided I2V Model}
\\

\hline

\endhead

\hline
\multicolumn{4}{r}{\footnotesize\textit{Continued on next page}}
\\

\endfoot

\bottomrule[1.5pt]


\endlastfoot


\shortstack[c]{
    Direct V2V \\
    (53.80\%)
}
&
\shortstack[c]{
    \rule{0pt}{2.2ex}
    Kling-O1 \\
    (5.80\%) \\
    Kling-V3-Omni \\
    (2.65\%) \\
    Runway Aleph \\
    (8.48\%) \\
    Wan 2.2-Video-Edit \\
    (10.43\%) \\
    VideoCoF \\
    (21.36\%) \\
    VACE \\
    (4.39\%) \\
    VideoPainter \\
    (0.69\%)
    \rule[-0.8ex]{0pt}{1.5ex}
}
&
--
&
--
\\

\hline


\multirow[c]{6}{*}{
\shortstack[c]{
    Indirect I2V\\
    (46.20\%)
}
}
&
\multirow[c]{6}{*}{
\shortstack[c]{
    --
}
}
&
\shortstack[c]{
    Nano Banana 2\\
    (13.92\%)
}
&
\shortstack[c]{
    Kling-V3-Omni\\
    (1.77\%)\\
    Wan 2.2\\
    (5.54\%)\\
    LTX-2\\
    (6.60\%)
}
\\

\cline{3-4}

&
&
\shortstack[c]{
    Seedream 5.0\\
    (14.16\%)
}
&
\shortstack[c]{
    Kling-V3-Omni\\
    (1.81\%)\\
    Wan 2.2\\
    (7.72\%)\\
    LTX-2\\
    (4.63\%)
}
\\

\cline{3-4}

&
&
\shortstack[c]{
    GPT-Image-1.5\\
    (0.39\%)
}
&
\shortstack[c]{
    Kling-V3-Omni\\
    (0.00\%)\\
    Wan 2.2\\
    (0.39\%)\\
    LTX-2\\
    (0.00\%)
}
\\

\cline{3-4}

&
&
\shortstack[c]{
    FLUX.2-dev-32B\\
    (5.75\%)
}
&
\shortstack[c]{
    Kling-V3-Omni\\
    (1.14\%)\\
    Wan 2.2\\
    (1.27\%)\\
    LTX-2\\
    (3.34\%)
}
\\

\cline{3-4}

&
&
\shortstack[c]{
    FLUX.2-klein-9B\\
    (3.34\%)
}
&
\shortstack[c]{
    Kling-V3-Omni\\
    (0.00\%)\\
    Wan 2.2\\
    (1.41\%)\\
    LTX-2\\
    (1.93\%)
}
\\

\cline{3-4}

&
&
\shortstack[c]{
    Qwen-Image\\
    (8.64\%)
}
&
\shortstack[c]{
    Kling-V3-Omni\\
    (0.43\%)\\
    Wan 2.2\\
    (3.87\%)\\
    LTX-2\\
    (4.34\%)
}
\\

\end{longtable}

\endgroup

\FloatBarrier
\subsubsection{Generation Model Details}
\label{app:dataset_model_details}

\noindent\textbf{Kling-O1}~\citep{team2025kling}\quad
Kling-O1 belongs to the Kling-Omni framework, which unifies video generation and editing through multimodal instruction following.
It conditions video creation on text, reference images, and video context, enabling edits that combine semantic instructions with information from an existing scene.

\noindent\textbf{Kling-V3-Omni}~\citep{kling_video_3_omni}\quad
Kling-V3-Omni is a unified multimodal video model that supports reference-conditioned generation, video editing, and audiovisual output.
Its reference and element controls help maintain subject consistency across generated content, and we use it in both the direct V2V and indirect I2V manipulation paradigms.

\noindent\textbf{Runway Aleph}~\citep{runwayaleph}\quad
Runway Aleph is an in-context video model that edits an existing clip according to user instructions.
Its supported operations include adding, removing, and transforming objects, changing viewpoints, and modifying scene style or lighting.

\noindent\textbf{Wan 2.2-Video-Edit}~\citep{wavespeed_wan22_video_edit}\quad
Wan 2.2-Video-Edit is a Wan-based video-editing model provided through WaveSpeedAI that accepts a source video and a textual edit instruction.
It supports changes to visual content such as clothing and characters, allowing the source clip to be directly transformed under the V2V paradigm.

\noindent\textbf{VideoCoF}~\citep{yang2026videocof}\quad
VideoCoF is a unified video-editing framework that uses a chain-of-frames reasoning process to determine where an instruction should be applied.
Its diffusion model predicts edit-region reasoning latents before generating the target video, providing spatial guidance without requiring user-supplied masks.

\noindent\textbf{VACE}~\citep{jiang2025vace}\quad
VACE unifies reference-to-video generation, video-to-video editing, and masked video editing within a diffusion-based framework.
It combines reference content, source frames, and masks in a shared video condition unit, supplied to the model through a context adapter.

\noindent\textbf{VideoPainter}~\citep{bian2025videopainter}\quad
VideoPainter is a video-inpainting and editing framework with separate streams for background context and content generation.
A lightweight context encoder supplies masked-video information to a pretrained video diffusion transformer, while target-region identity resampling supports editing videos beyond a single generation window.

\noindent\textbf{Nano Banana 2}~\citep{google_nano_banana_2}\quad
Nano Banana 2, also known as Gemini 3.1 Flash Image, is Google's image-generation and editing model designed for rapid instruction-guided visual creation.
It supports reference-based edits and subject consistency across transformations, making it suitable for producing edited first frames from source images and manipulation instructions.

\noindent\textbf{Seedream 5.0}~\citep{bytedance_seedream5}\quad
Seedream 5.0 is ByteDance's image-generation and editing family, with the cited Lite release combining multimodal understanding, reasoning, and image synthesis.
It supports instruction-guided local edits and subject replacement while preserving surrounding content, and serves as a first-frame editor in our indirect I2V pipeline.

\noindent\textbf{GPT-Image-1.5}~\citep{openai2025gptimage15}\quad
GPT-Image-1.5 is a proprietary image-generation and editing model that follows natural-language instructions over text and image inputs.
It supports localized and broader visual transformations while retaining details such as subject identity, lighting, and composition, enabling controlled modification of the source video's first frame.

\noindent\textbf{FLUX.2-dev-32B}~\citep{blackforestlabs_flux2_dev}\quad
FLUX.2-dev-32B is an open-weight rectified-flow transformer with 32 billion parameters for image generation and editing.
It combines single- and multiple-reference image conditioning with guidance distillation for efficient inference.

\noindent\textbf{FLUX.2-klein-9B}~\citep{blackforestlabs_flux2_klein_9b}\quad
FLUX.2-klein-9B is a compact rectified-flow transformer that combines text-to-image generation with reference-based image editing.
Its nine-billion-parameter flow model is distilled to four inference steps, providing an efficient alternative for instruction-guided first-frame editing.

\noindent\textbf{Qwen-Image}~\citep{wu2025qwen}\quad
Qwen-Image is an image-generation and editing model that combines multimodal semantic understanding with a diffusion transformer.
Its editing design encodes the source image into semantic and reconstructive representations, helping preserve both the intended content and visual fidelity during instruction-guided changes.

\noindent\textbf{Wan 2.2}~\citep{wan2025wan}\quad
Wan 2.2 is an open video diffusion model in the Wan family that supports image-conditioned video synthesis.
In our indirect I2V pipeline, the edited first frame defines the target appearance, while source-video Canny maps guide the scene geometry.

\noindent\textbf{LTX-2}~\citep{hacohen2026ltx}\quad
LTX-2 is an open audiovisual generation model with asymmetric video and audio transformer streams connected through bidirectional cross-attention.
Its two streams generate temporally aligned audiovisual content; our indirect I2V pipeline conditions video generation on an edited first frame and Canny guidance.

\FloatBarrier
\subsubsection{Paradigm A: Direct V2V Cases}
Fig.~\ref{fig:dataset_per_v2v_model} shows examples of direct manipulation performed by different V2V models. Specifically, we select four recent closed-source models, including Kling-O1~\citep{team2025kling}, Kling-V3-Omni~\citep{kling_video_3_omni}, Runway Aleph~\citep{runwayaleph}, Wan 2.2-Video-Edit~\citep{wavespeed_wan22_video_edit}, together with three open-source models, including VideoCoF~\citep{yang2026videocof}, VACE~\citep{jiang2025vace} and VideoPainter~\citep{bian2025videopainter}.

\begin{figure*}[p]
    \centering
    \includegraphics[width=0.98\textwidth]{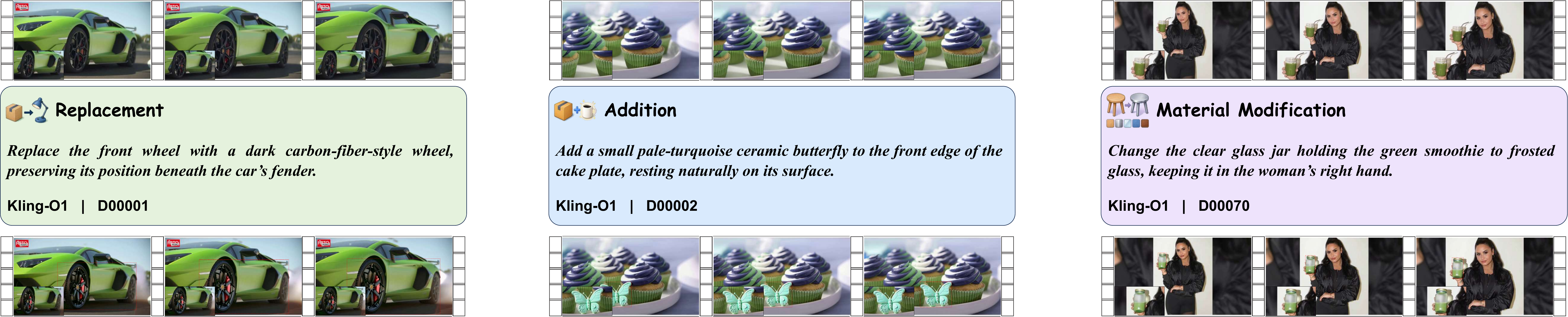}
    \par\vspace{-1mm}
    \hdashrule{0.98\textwidth}{0.6pt}{3pt 2pt}
    \par\vspace{2mm}
    \includegraphics[width=0.98\textwidth]{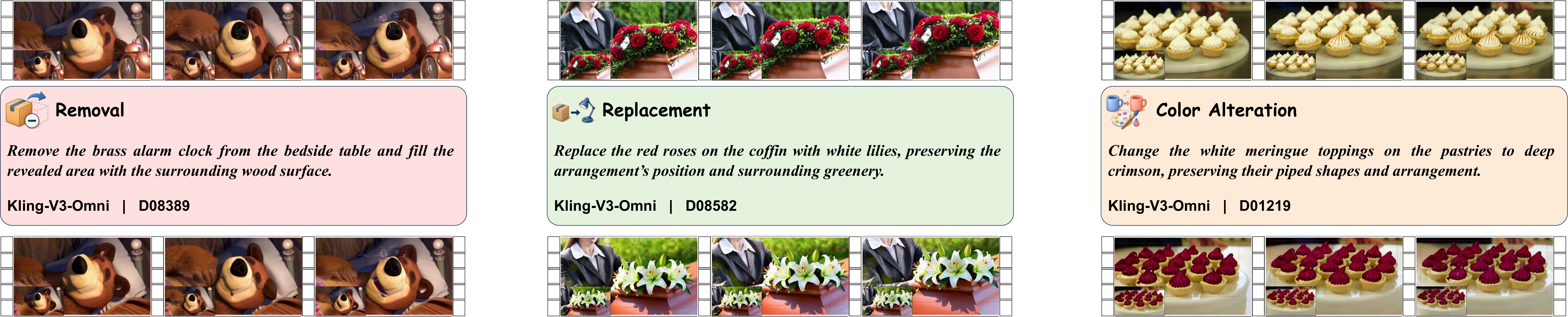}
    \par\vspace{-1mm}
    \hdashrule{0.98\textwidth}{0.6pt}{3pt 2pt}
    \par\vspace{2mm}
    \includegraphics[width=0.98\textwidth]{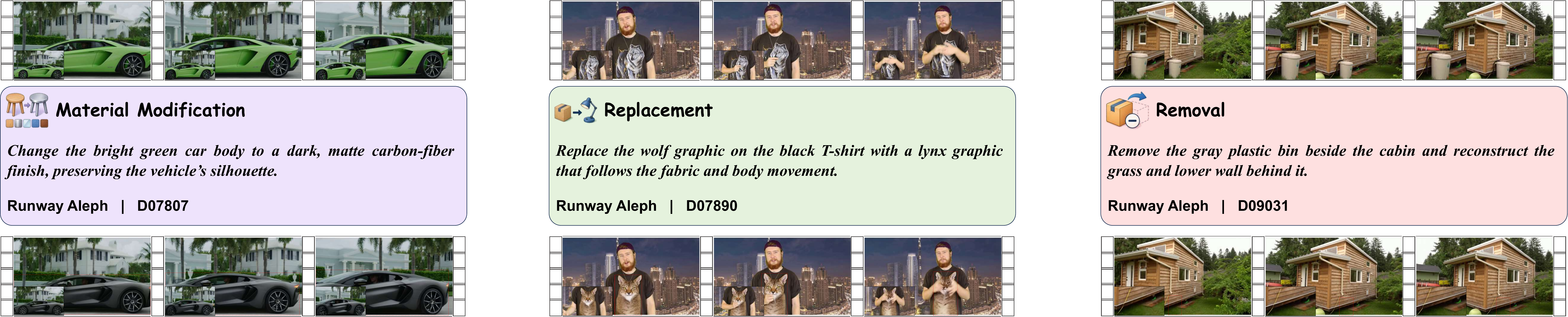}
    \par\vspace{-1mm}
    \hdashrule{0.98\textwidth}{0.6pt}{3pt 2pt}
    \par\vspace{2mm}
    \includegraphics[width=0.98\textwidth]{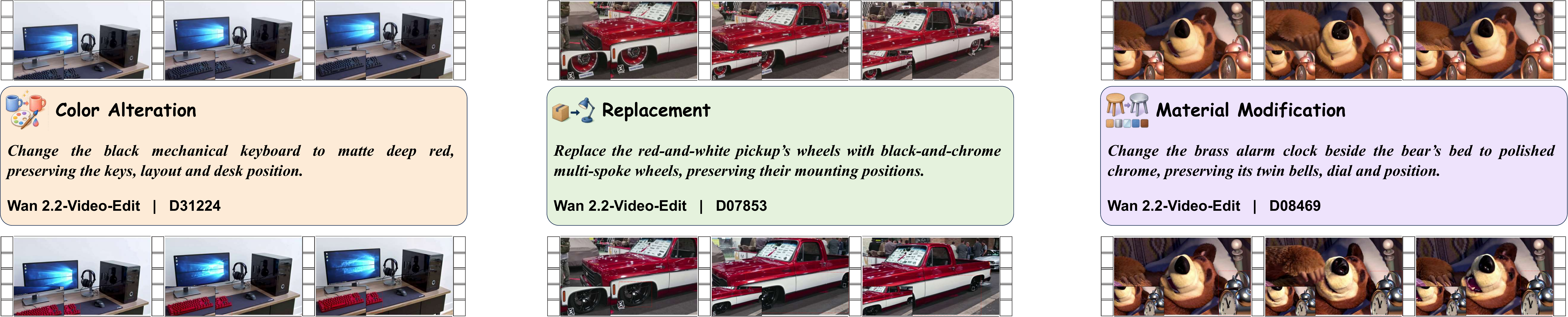}
    \par\vspace{-1mm}
    \caption{\textbf{Manipulation Examples of Seven Direct V2V Models (1/2).} Closed-source models. Zoom in for a better view.}
    \label{fig:dataset_per_v2v_model}
\end{figure*}

\begin{figure*}[p]
    \centering
    \ContinuedFloat
    \includegraphics[width=0.98\textwidth]{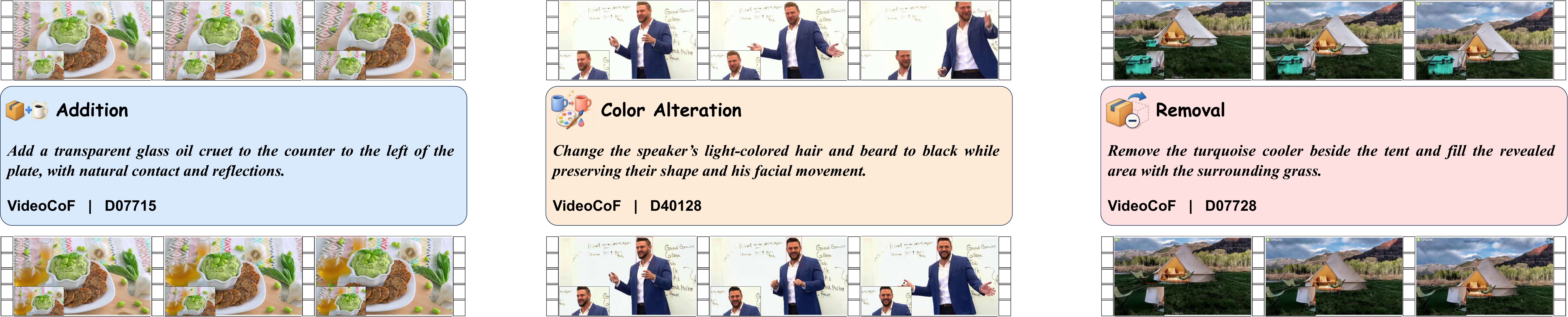}
    \par\vspace{-1mm}
    \hdashrule{0.98\textwidth}{0.6pt}{3pt 2pt}
    \par\vspace{2mm}
    \includegraphics[width=0.98\textwidth]{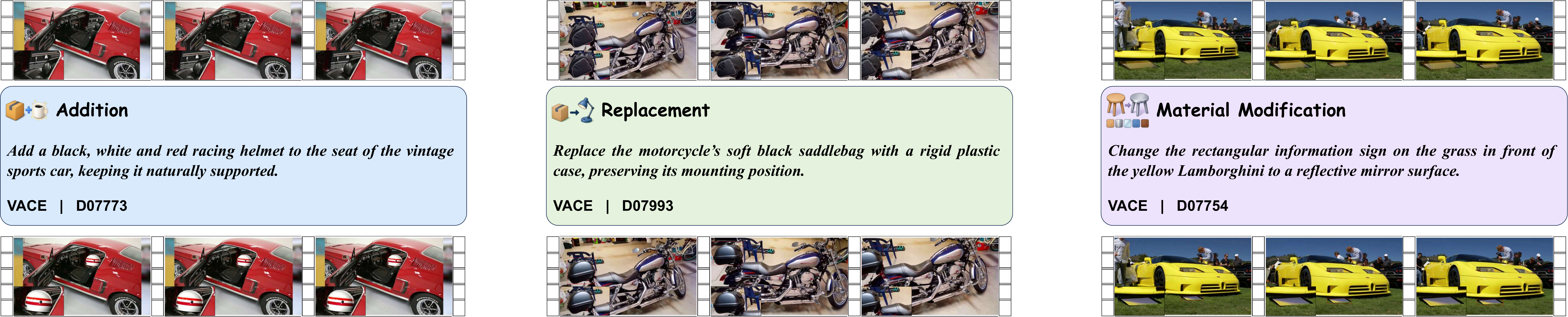}
    \par\vspace{-1mm}
    \hdashrule{0.98\textwidth}{0.6pt}{3pt 2pt}
    \par\vspace{2mm}
    \includegraphics[width=0.98\textwidth]{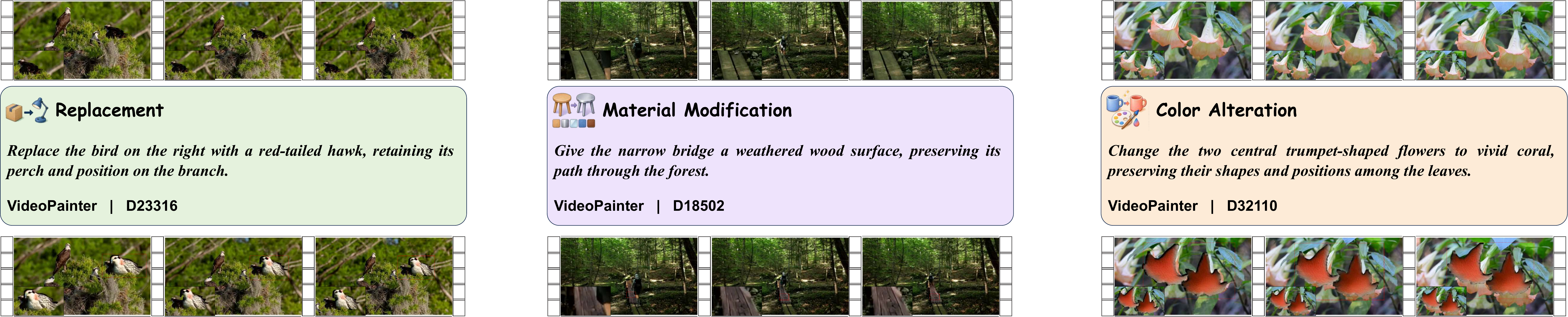}
    \par\vspace{-1mm}
    \caption[]{\textbf{Manipulation Examples of Seven Direct V2V Models (2/2).} Open-source models. Zoom in for a better view.}
\end{figure*}

\FloatBarrier
\subsubsection{Paradigm B: Indirect I2V Cases}
\begin{figure*}[h]
    \centering
    \vspace{-1mm}
    \includegraphics[width=0.98\textwidth]{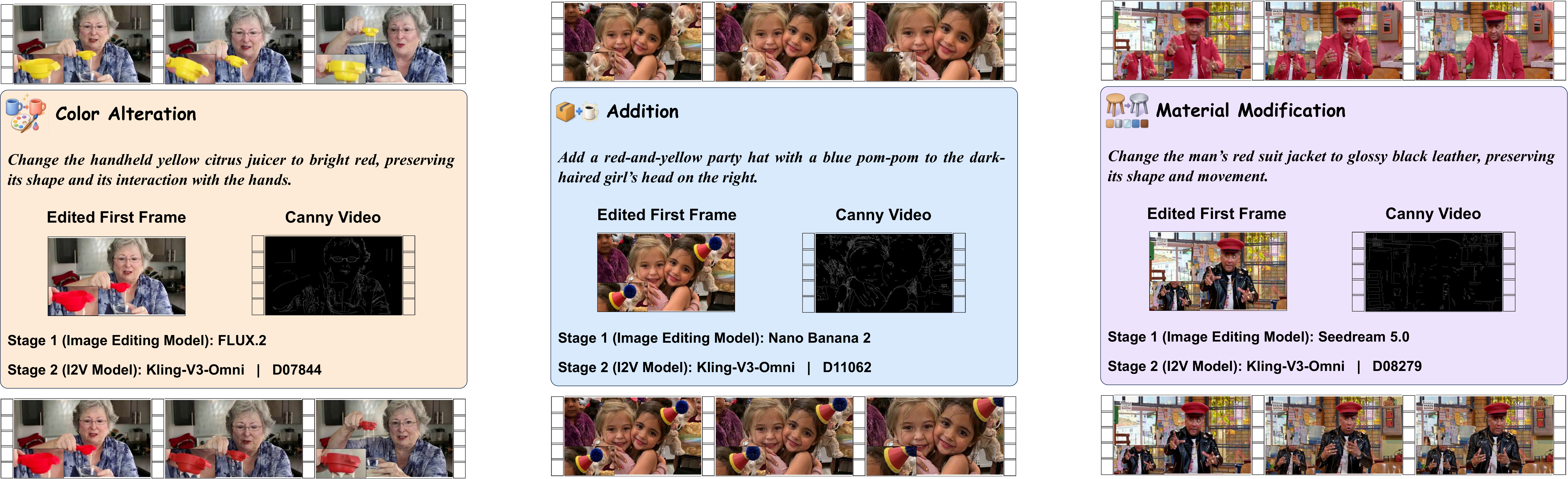}


    
    \parbox[c][1mm][c]{\textwidth}{%
    \centering
    \raisebox{1mm}{%
        \hdashrule{0.99\textwidth}{0.6pt}{3pt 2pt}%
    }%
    }   

    \includegraphics[width=0.98\textwidth]{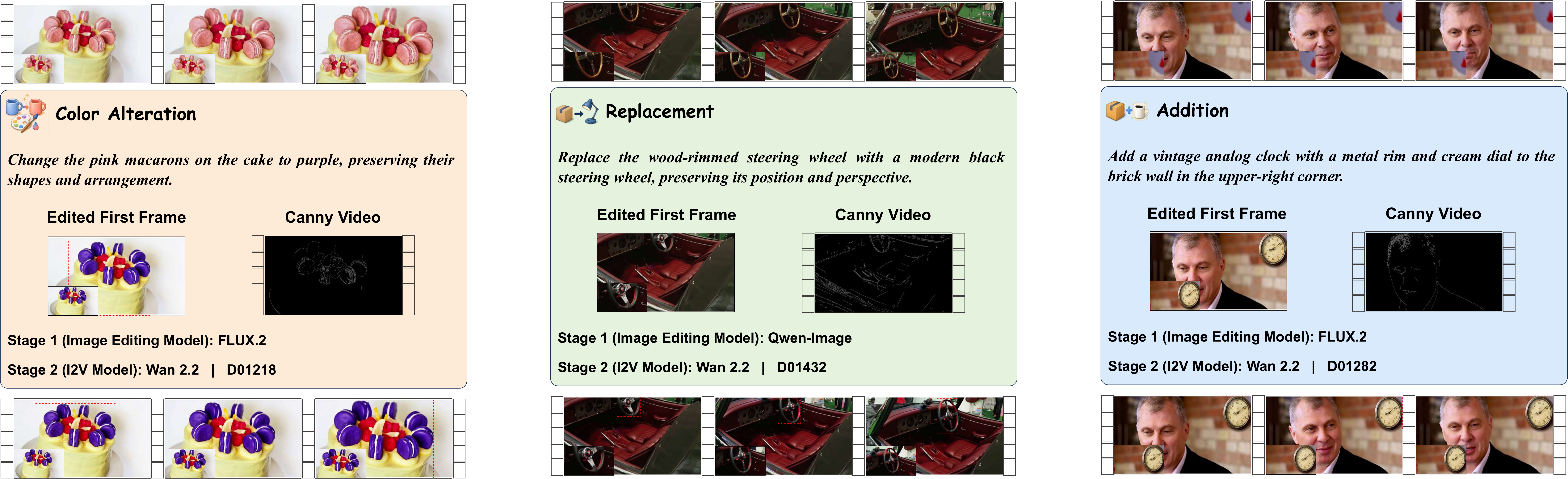}

    \parbox[c][1mm][c]{\textwidth}{%
    \centering
    \raisebox{1mm}{%
        \hdashrule{0.99\textwidth}{0.6pt}{3pt 2pt}%
    }%
    } 

    \includegraphics[width=0.98\textwidth]{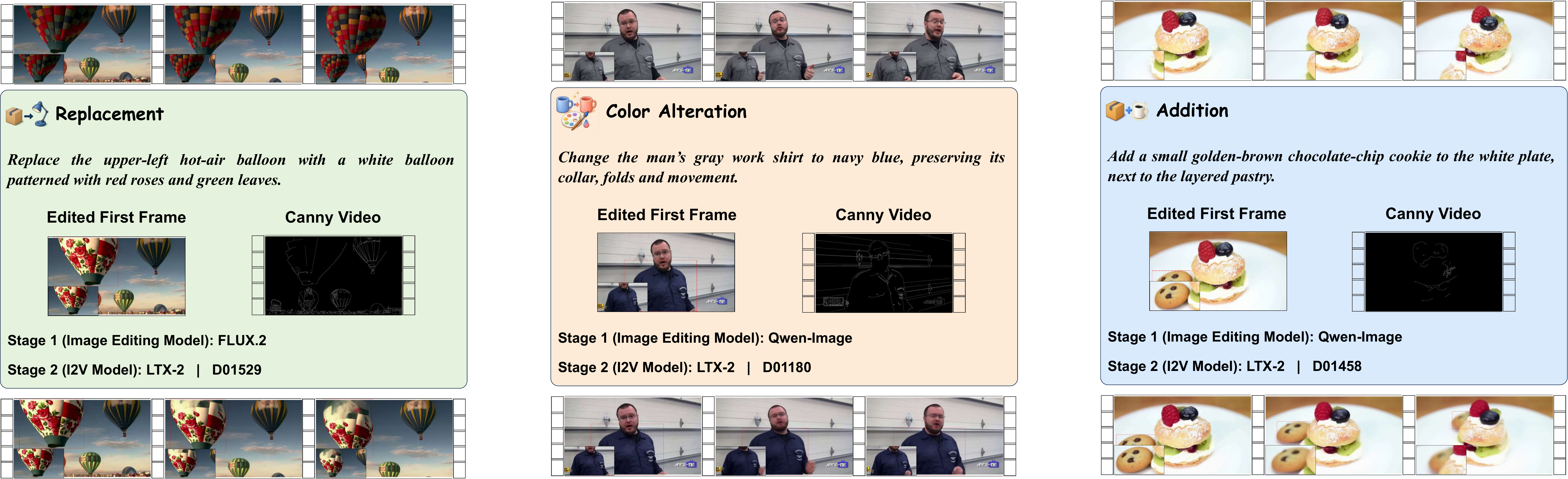}
    \vspace{-1mm}
    \caption{\textbf{Manipulation Examples of Three Indirect I2V Models.} The first row is Kling-V3-Omni~(closed source), while the last two rows are open source models. Zoom in for a better view.}
    \vspace{-3mm}
    
    \label{fig:dataset_per_i2v_model}
\end{figure*}

Fig.~\ref{fig:dataset_per_i2v_model} shows examples of indirect manipulation performed under the I2V paradigm. Specifically, for Stage 1 image editing, we select three recent closed-source models, including Nano Banana 2~\citep{google_nano_banana_2}, Seedream 5.0~\citep{bytedance_seedream5}, and GPT-Image-1.5~\citep{openai2025gptimage15}, together with three open-source models, including FLUX.2-dev~\citep{blackforestlabs_flux2_dev}, FLUX.2-klein-9B~\citep{blackforestlabs_flux2_klein_9b}, and Qwen-Image~\citep{wu2025qwen}. For Stage 2 controllable I2V generation, we select three Canny-guided I2V models, including the closed-source Kling-V3-Omni~\citep{kling_video_3_omni} and two open-source models, Wan 2.2~\citep{wan2025wan} and LTX-2~\citep{hacohen2026ltx}.

\FloatBarrier
\subsubsection{Automatic Reflection}
To ensure high-quality manipulation results while reducing unnecessary time and resource costs in subsequent annotation and quality control, we employ the MLLM assistant to automatically assess whether each manipulation is successful. Specifically, given the manipulation instruction, the assistant compares the source and manipulated videos, focusing on three key aspects: \textbf{\textit{\textcolor{Red}{1) Manipulation Visibility}}}, \textbf{\textit{\textcolor{Red}{2) Instruction Following}}}, and \textbf{\textit{\textcolor{Red}{3) Content Fidelity}}}. Only samples satisfying all three criteria proceed to subsequent annotation and quality control. During this process, we retain about 79.8\% of the samples in total. The prompt used for automatic reflection is provided in Tab.~\ref{app:tab:automatic_reflection_prompt}. 

\begin{table*}[htb]\centering
\vspace{-2mm}
\caption{\textbf{Automatic Reflection Prompt.}}
    \vspace{1mm}
    \begin{minipage}{\textwidth}\vspace{0mm}
    \centering
    \begin{tcolorbox} 
    \centering
    \fontsize{8pt}{10pt}\selectfont
    \begin{tabular}{p{\dimexpr\linewidth-4\tabcolsep\relax} c}

    \VarSty{{\bf \normalsize Automatic Reflection Prompt}} &\\
    
    \vspace{0.1\baselineskip}

    {\small\bfseries Task Description:}

    \vspace{0.1\baselineskip}
    
    You are given a \textbf{source video}, a \textbf{manipulation instruction}: 
    \texttt{\textless MANIPULATION\_INSTRUCTION\textgreater}, 
    and the corresponding \textbf{manipulated video}.
    
    Your task is to compare the source video and the manipulated video according to the manipulation instruction, and determine whether the manipulation is successfully performed.

    \vspace{0.5\baselineskip}
    
    {\small\bfseries Manipulation Check:}

    \begin{enumerate}[leftmargin=2em, labelsep=0.5em, itemsep=-2pt, topsep=2pt]
        \item \textbf{Manipulation Visibility:} Is the intended manipulation visible in the manipulated video?
        \item \textbf{Instruction Following:} Does the manipulated result comply with the manipulation instruction?
        \item \textbf{Content Fidelity:} Are the remaining unmanipulated regions preserved compared with the source video?
    \end{enumerate}

    \vspace{0.3\baselineskip}

    {\small\bfseries Requirements:}

    \begin{enumerate}[leftmargin=2em, labelsep=0.5em, itemsep=-2pt, topsep=2pt]
        \item Carefully compare the \textbf{source video} and the \textbf{manipulated video} with reference to the \textbf{manipulation instruction}.
        \item For each of the three checks above, output only \texttt{YES} or \texttt{NO}.
        \item The manipulation is considered successful only when all three checks are \texttt{YES}.
        \item Output a single JSON object and nothing else, with no explanation, no chain of thought, and no markdown fences:
        
        \{``manipulation\_visible": ``\texttt{\textless YES/NO\textgreater}", 
        ``instruction\_following": ``\texttt{\textless YES/NO\textgreater}", 
        ``content\_fidelity": ``\texttt{\textless YES/NO\textgreater}", 
        ``success": ``\texttt{\textless YES/NO\textgreater}"\}
    \end{enumerate}
    
    \end{tabular}
    \end{tcolorbox}
    \vspace{-2mm}

    \label{app:tab:automatic_reflection_prompt}

    \end{minipage}
\end{table*}


\FloatBarrier
\subsubsection{Artifact Grounding}
We first employ an MLLM assistant to extract a concise and common \textit{segmentation phrase} from the manipulation instruction, since SAM3~\citep{carion2026sam} is less compatible with rare concepts and lengthy textual prompts. We then feed the segmentation phrase together with the target video into SAM3 to obtain the forgery masks. Specifically, for the \textbf{\textit{Removal}} task, the source video is used as the target video because the manipulated video no longer contains the target semantic content; for all other manipulation tasks, the manipulated video is used instead.

\noindent \textbf{Dilemma: Poor Performance.}
However, we observe that \textbf{\textit{\textcolor{Red}{SAM3 performs poorly when the target reference is ambiguous or multiple similar objects are present}}}, which substantially degrades the quality of artifact annotations. Therefore, we introduce additional spatial priors to enhance the segmentation performance of SAM3.

\noindent \textbf{Solution: Point Annotation.}
Through preliminary experiments, we find that complex bbox or polygon annotations are unnecessary, as only a few simple positive and negative point prompts can substantially improve the segmentation performance of SAM3 in challenging scenarios. Specifically, \textbf{\textit{\textcolor{Green}{positive points}}} indicate manipulated regions, while \textbf{\textit{\textcolor{Red}{negative points}}} indicate potentially confusing preserved regions. For each sample, the total number of positive and negative points is limited to at most six. The entire process is conducted through crowdsourcing, during which annotators are also instructed to directly discard samples with poor manipulation quality or those inconsistent with the manipulation instruction. During this process, we retain about 77.9\% of the samples in total.

\FloatBarrier
\subsubsection{Anomaly Explanation}
Inspired by VidGuard-R1~\citep{park2025vidguard}, we define a comprehensive taxonomy of anomaly explanations from eight forensic perspectives: \textbf{(1) Boundary Coherence}, \textbf{(2) Perspective and Proportion}, \textbf{(3) Contextual Inconsistency}, \textbf{(4) Color and Lighting}, \textbf{(5) Texture Discontinuity}, \textbf{(6) Temporal Flicker}, \textbf{(7) Motion Artifact}, and \textbf{(8) Ghosting}, as detailed in Tab.~\ref{app:tab:anomaly_explanation_definition}. Given the manipulated video and its forgery mask, we prompt the MLLM assistant to select the four most applicable perspectives and generate corresponding forensic explanations. For removal cases, where the manipulated region corresponds to the reconstructed background rather than the removed object itself, the assistant is specifically instructed to analyze the visual traces of background inpainting without inferring the removed content. The explanation generation prompt is provided in Tab.~\ref{app:tab:anomaly_explanation_prompt}.

\begin{table}[!htbp]
\centering
\footnotesize
\caption{\textbf{Definitions of Anomaly Explanation Perspectives.}}
\definecolor{mygray}{gray}{.92}
\renewcommand{\arraystretch}{1.3}
\belowrulesep=-0.25pt
\aboverulesep=-0.25pt
\resizebox{1.0\linewidth}{!}{
    \begin{tabular}{
    >{\centering\arraybackslash}m{4.0cm}|
    m{12.5cm}
    }
    \toprule[1.5pt]
    \textbf{Anomaly Perspective} & \multicolumn{1}{c}{\textbf{Definition}} \\
    \hline

    \textit{Boundary Coherence} &
    Imperfect transitions between the manipulated region and the original footage, including overly smoothed or unnaturally sharp boundary edges, as well as texture discrepancies in surface noise, resolution, or material quality relative to the surrounding unedited areas. \\
    \hline

    \textit{Perspective and Proportion} &
    Geometric spatial errors where the manipulated content violates the camera perspective, exhibits focal-length-related distortions, or shows unnatural scaling and structural proportions relative to its immediate environment. \\
    \hline

    \textit{Contextual Inconsistency} &
    Semantic or logical anomalies where the manipulated content defies the established scene context, including physically impossible object interactions, gravity-defying kinematics, or elements that fundamentally contradict surrounding real-world physics. \\
    \hline

    \textit{Color and Lighting} &
    Mismatches in illumination, shadow trajectories, or color-space distribution between the manipulated content and the global physical environment. \\
    \hline

    \textit{Texture Discontinuity} &
    Sudden degradation in surface quality, excessive smoothing, or unnatural loss of high-frequency details within the manipulated region. \\
    \hline

    \textit{Temporal Flicker} &
    Unstable pixel generation or erratic high-frequency variations of static visual features across consecutive frames, typically manifested as localized shimmering or micro-jitter in textures that should remain structurally rigid. \\
    \hline

    \textit{Motion Artifact} &
    Kinematic inconsistencies, physical deformations during movement, or unnatural motion trajectories of the manipulated content over time. \\
    \hline

    \textit{Ghosting} &
    Unnatural translucency, residual visual echoes, or trailing artifacts associated with the manipulated content during frame transitions. \\

    \bottomrule[1.5pt]
    \end{tabular}
}
\vspace{-1mm}
\label{app:tab:anomaly_explanation_definition}
\end{table}

\begingroup
\vspace{15mm}
\captionsetup{hypcap=false}
\captionof{table}{\textbf{Anomaly Explanation Prompt.}}
\vspace{-2mm}
\label{app:tab:anomaly_explanation_prompt}
\begin{tcolorbox}[
    breakable,
    fontupper=\fontsize{8pt}{10pt}\selectfont,
    before upper={\setlength{\parindent}{0pt}\setlength{\parskip}{0pt}}
]
\VarSty{{\bf \normalsize Anomaly Explanation Prompt}}\par

\vspace{0.1\baselineskip}

{\small\bfseries Task Description:}

\vspace{0.1\baselineskip}

You are a leading digital forensics expert and video analyst, skilled at identifying deepfakes and AI-generated fake videos by analyzing subtle visual clues. Given a manipulated video and its target manipulation region, your task is to identify anomalous visual features and generate a rigorous forensic explanation that connects the observed abnormalities to AI manipulation. Specifically, analyze the targeted region from the following eight forensic perspectives and select the \textbf{four most applicable} ones for explanation.

\vspace{0.5\baselineskip}

{\small\bfseries Input Data:}

\begin{enumerate}[leftmargin=2em, labelsep=0.5em, topsep=0pt, itemsep=1pt]
    \item \textbf{Annotated Video}: A manipulated video in which the target manipulation region~(bounding boxes) is visually highlighted across frames.
    \item \textbf{Manipulation Context}: Textual information describing the manipulation type and target content, used jointly with the visual indication to identify the manipulated region.
\end{enumerate}

\vspace{0.5\baselineskip}

{\small\bfseries Anomaly Perspective Definitions:}

\vspace{0.1\baselineskip}

\begin{itemize}[leftmargin=2em, labelsep=0.5em, topsep=-2pt, itemsep=1pt]

    \item \textbf{\textit{Boundary Coherence}}:
    Imperfect transitions between the manipulated region and the original footage, exhibiting overly smoothed or unnaturally sharp boundary edges. This also includes texture discrepancies where the surface noise, resolution, or material quality fundamentally clashes with the surrounding unedited areas.

    \item \textbf{\textit{Perspective and Proportion}}:
    Geometric spatial errors where the manipulated content violates the camera's perspective grid, exhibits focal-length distortions, or shows unnatural scaling and structural proportions relative to its immediate environment.

    \item \textbf{\textit{Contextual Inconsistency}}:
    Semantic or logical anomalies where the manipulated content defies the established scene context. This includes physically impossible object interactions, gravity-defying kinematics, or elements that fundamentally contradict the surrounding real-world physics.

    \item \textbf{\textit{Color and Lighting}}:
    Mathematical mismatches in illumination, shadow trajectories, or color-space distribution compared with the global physical environment.

    \item \textbf{\textit{Texture Discontinuity}}:
    Sudden degradation in surface quality, over-smoothing, or unnatural loss of high-frequency details within the manipulated region.

    \item \textbf{\textit{Temporal Flicker}}:
    Unstable pixel generation or erratic high-frequency variations of static features across consecutive frames. Such mathematical instability typically manifests as localized shimmering or micro-jitter in textures that should remain structurally rigid.

    \item \textbf{\textit{Motion Artifact}}:
    Kinematic inconsistencies, physical deformations during movement, or unnatural trajectories of the manipulated content over time.

    \item \textbf{\textit{Ghosting}}:
    Unnatural translucency, residual visual echoes, or trailing artifacts during frame transitions.

\end{itemize}

\vspace{0.5\baselineskip}

{\small\bfseries Requirements:}

\begin{enumerate}[leftmargin=2em, labelsep=0.5em, topsep=2pt, itemsep=1pt]

    \item \textbf{Target Region Description}: Provide a concise and unambiguous referring expression of at most 20 words for the manipulated object or background. State exactly \textbf{what} it is, its key visual attributes such as color or shape, and \textbf{where} it is located. Do \textbf{not} mention the visual annotation or bounding box. The description should be suitable as a direct referring phrase for a vision segmentation model.

    \item \textbf{Forensic Reasoning}: For each selected perspective, follow a strict evidentiary progression:
    \begin{enumerate}[leftmargin=2em, label=(\alph*), labelsep=0.5em, topsep=-2pt, itemsep=-2pt]
        \item summarize the objective visual anomaly;
        \item explain how the anomaly violates physical laws or optical camera capture within the corresponding perspective;
        \item justify why this divergence serves as a signature of AI generation.
    \end{enumerate}

    \item \textbf{Removal Cases}: For \textbf{\textit{Removal}} manipulations, the target region corresponds to the reconstructed background after object removal rather than the removed object itself. Focus only on the visual traces of background inpainting and reconstruction. Do \textbf{not} attempt to restore, infer, or guess the removed content. The target-region description should instead refer to the reconstructed background patch, including its appearance, spatial location, and adjacent content.

    \item \textbf{Objectivity and Focus}: Do \textbf{not} hallucinate visual features, over-interpret normal artifacts, or introduce external examples. Strictly analyze the content within the targeted manipulation region.

    \item \textbf{Output Format}: Return the response strictly as a valid JSON object without Markdown formatting, code blocks, or conversational filler. The JSON must contain one \texttt{mani\_description} field and exactly four fields corresponding to the selected anomaly perspectives.

\end{enumerate}

\end{tcolorbox}
\nopagebreak[4]
\endgroup

\FloatBarrier
\subsubsection{Human Verification}
To ensure the final dataset quality, we introduce human verification, jointly conducted by all authors, as the final quality control stage, comprehensively evaluating each sample from four aspects: 
\begin{itemize}[leftmargin=10pt, topsep=0pt, itemsep=3pt, partopsep=1pt, parsep=1pt]

\item \textbf{\textit{Task Reasonableness.}}
Although the MLLM assistant has already assessed the reasonableness of task assignment and manipulation instructions during instruction generation, we conduct a final human inspection to identify potential oversights and ensure that the assigned task and instruction for each source video are feasible and consistent with common sense and event logic.

\item \textbf{\textit{Manipulation Quality.}}
Although manipulation quality has already been assessed by the MLLM assistant during the Auto-Reflection stage, we further perform human inspection to ensure that the manipulated videos contain no obvious visual artifacts or instruction misalignment, thereby maintaining the quality and challenge of the ManiVid-38K dataset.

\item \textbf{\textit{Explanation Fidelity.}} 
We manually inspect the anomaly explanations generated by the MLLM assistant for evident hallucinations, including incorrect or nonexistent manipulated objects, overly homogeneous descriptions, and anomaly descriptions inconsistent with the actual video content.

\item \textbf{\textit{Mask Accuracy.}} 
We visually inspect the generated segmentation masks against the target objects specified in the manipulation instructions, ensuring that the manipulated objects are correctly and completely segmented while excluding samples with evident under-segmentation, over-segmentation, or incorrect target segmentation.
\end{itemize}

Samples that fail any criterion are discarded, with about 75.2\% of the samples retained to constitute our ManiVid-38K dataset.

\FloatBarrier
\subsection{Statistical Analysis}
\label{app:statistical}
\subsubsection{Basic Characteristics Analysis}

\begin{figure*}[htb]
    \centering
    \vspace{-2mm}
    \includegraphics[width=\textwidth]{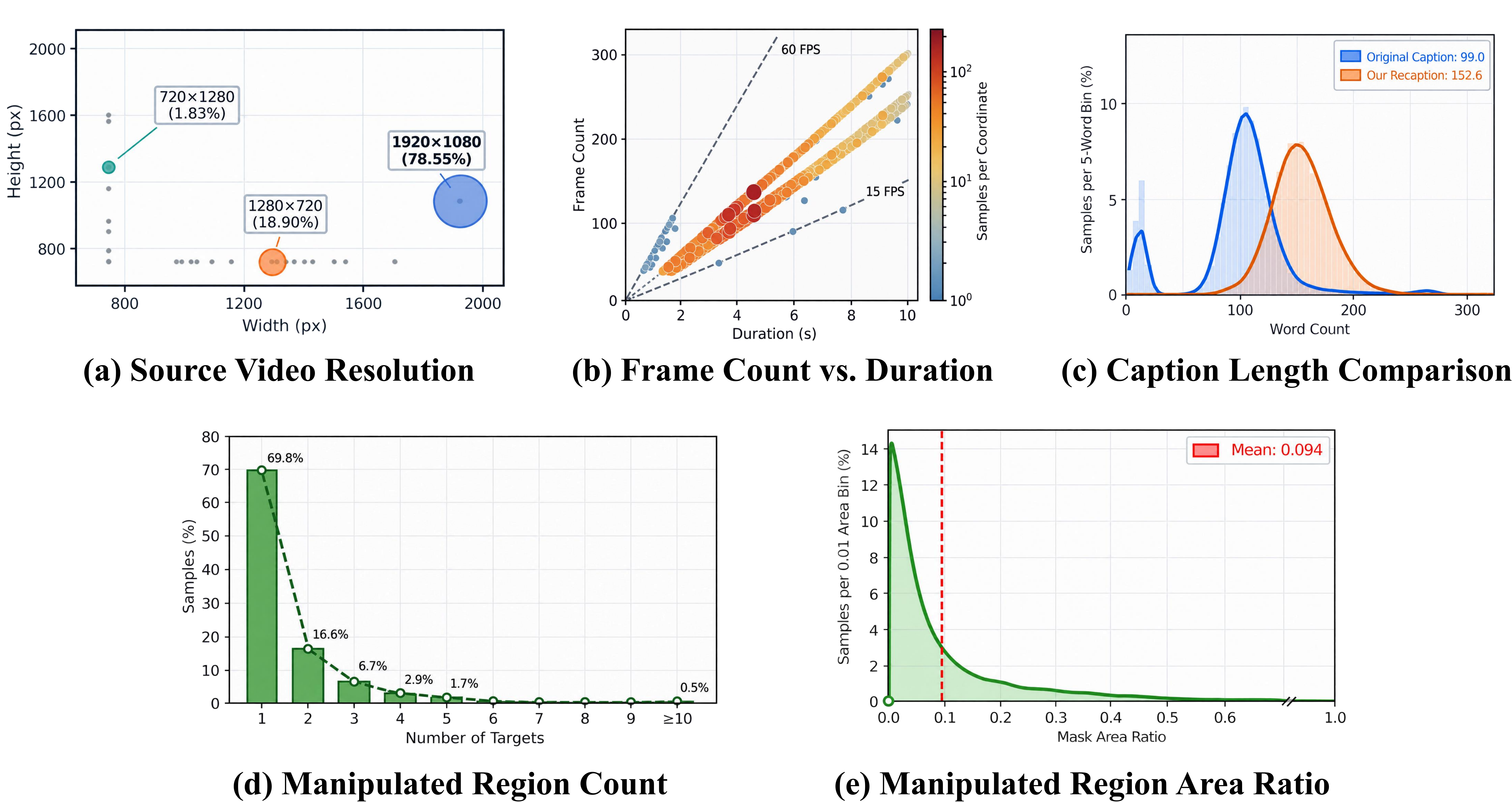}
    \vspace{-4mm}
    \caption{\textbf{Additional Statistical Analyses of ManiVid-38K.}}
    \vspace{3mm}
    \parbox[t]{0.96\textwidth}{\small\textbf{Reading the statistics.} The dataset combines high-resolution source videos, expanded recaptioned descriptions, sparse localized edits, and multiple manipulated regions. These distributions explain why the benchmark requires both semantic reasoning and precise pixel-level grounding.}
    
    \label{fig:manivid_more_statistics}
\end{figure*}

Figure~\ref{fig:manivid_more_statistics} summarizes the video and annotation characteristics of ManiVid-38K. Source videos predominantly have a resolution of 1920$\times$1080 (78.55\%), followed by 1280$\times$720 (18.90\%), with frame counts varying across durations and common frame rates. Recaptioning increases the average description length from 99.0 to 152.6 words, providing expanded textual descriptions for manipulation instruction generation. Spatially, 69.8\% of samples contain one manipulated region, while 16.6\% and 6.7\% contain two and three, respectively. The manipulated area occupies only 9.4\% of the frame on average, with the distribution concentrated toward smaller regions. These statistics characterize localized edits with varying spatial extent and region count, motivating precise artifact grounding within otherwise largely preserved scenes.

\FloatBarrier
\subsubsection{\texorpdfstring{Unedited-Region Fidelity Analysis}{Unedited-Region Fidelity Analysis}}
\label{app:unedited_fidelity}
To quantify how well the editing process preserves content outside the annotated manipulation masks, we evaluate pixel and semantic fidelity between each source video and its manipulated counterpart. For every corresponding frame, we exclude the annotated edited region and compute the mean squared error (MSE) over the remaining pixels and the cosine similarity between CLIP image embeddings of the masked frames. Lower MSE and higher CLIP similarity indicate better preservation.

\begin{table*}[t]
    \centering
    \small
    \setlength{\tabcolsep}{7pt}
    \caption{\textbf{Pixel and Semantic Fidelity of Unedited Regions Across Manipulation Tasks.}}
    \label{tab:unedited_fidelity_category}
    \begin{tabular}{lcc}
        \toprule
        Manipulation Task & MSE $\downarrow$ & CLIP similarity $\uparrow$ \\
        \midrule
        Addition & 0.019 & 0.940 \\
        Removal & 0.015 & \underline{0.945} \\
        Replacement & \underline{0.013} & \underline{0.945} \\
        Color Alteration & \underline{0.013} & \underline{0.945} \\
        Material Modification & 0.014 & \underline{0.945} \\
        Portrait Editing & \textbf{0.008} & \textbf{0.950} \\
        \midrule
        Average & 0.014 & 0.944 \\
        \bottomrule
    \end{tabular}
\end{table*}

Across all six manipulation types, MSE ranges from 0.008 to 0.019 and CLIP similarity from 0.940 to 0.950, indicating that unedited regions remain close to their source videos under both pixel-level and semantic measures.

\begin{table}[t]
    \centering
    \small
    \setlength{\tabcolsep}{5pt}
    \caption{\textbf{Unedited-region Fidelity by Manipulation Paradigms.}}
    \label{tab:unedited_fidelity_paradigm}
    \begin{tabular}{lcc}
        \toprule
        Manipulation Paradigm & MSE $\downarrow$ & CLIP similarity $\uparrow$ \\
        \midrule
        Direct V2V & \underline{0.014} & \underline{0.944} \\
        Indirect I2V & \textbf{0.013} & \textbf{0.945} \\
        \bottomrule
    \end{tabular}
\end{table}

The two editing paradigms have similar unedited-region fidelity scores, providing no evidence of a systematic difference in background preservation that would explain their performance gap.



\FloatBarrier
\subsection{Case Visualization}
\label{app:case_visual}
\begin{figure*}[t]
    \centering
    \includegraphics[width=0.94\textwidth]{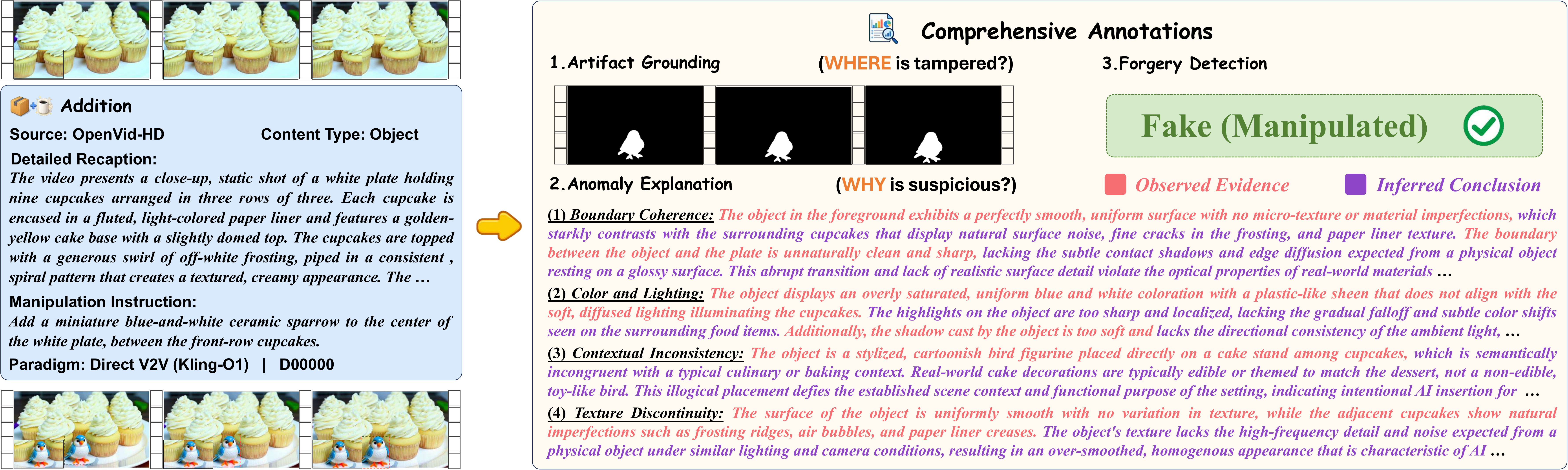}

    \vspace{2mm}


    \includegraphics[width=0.94\textwidth]{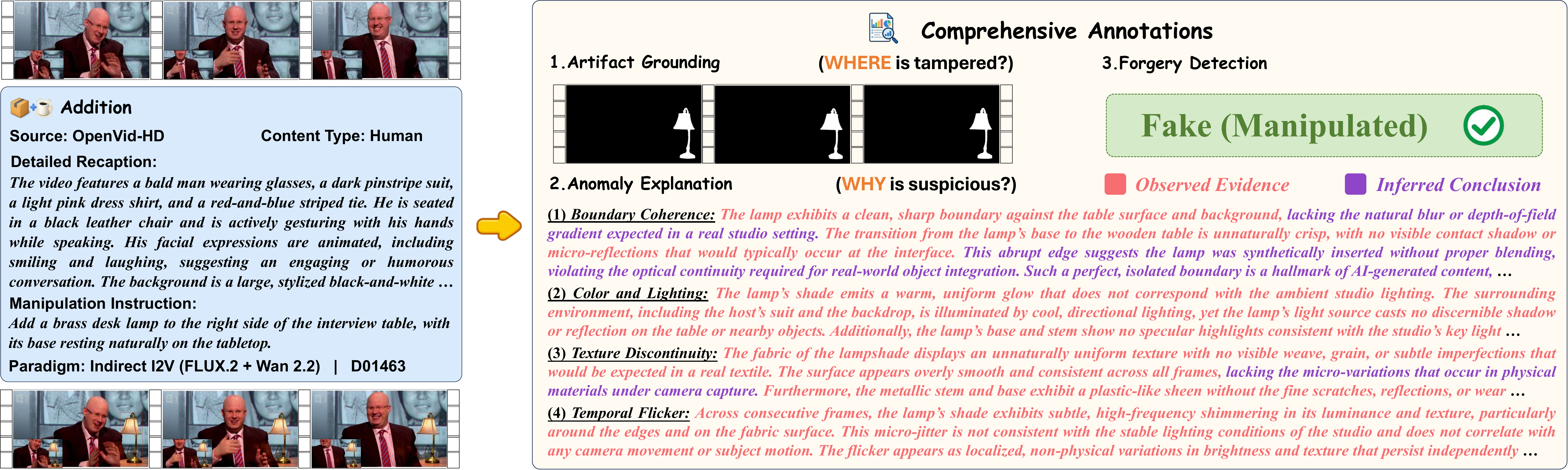}
    
    \parbox[c][1mm][c]{\textwidth}{%
    \centering
    \raisebox{1mm}{%
        \hdashrule{0.94\textwidth}{0.6pt}{3pt 2pt}%
    }%
    }
    \vspace{-3mm}

    \includegraphics[width=0.94\textwidth]{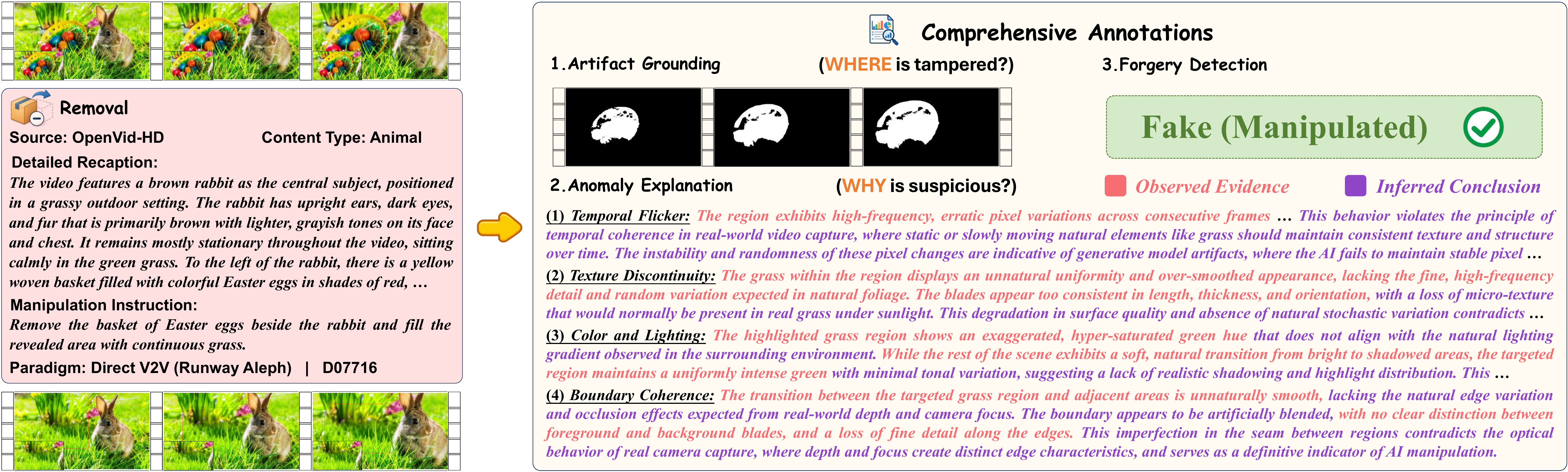}

    \vspace{2mm}


    \includegraphics[width=0.94\textwidth]{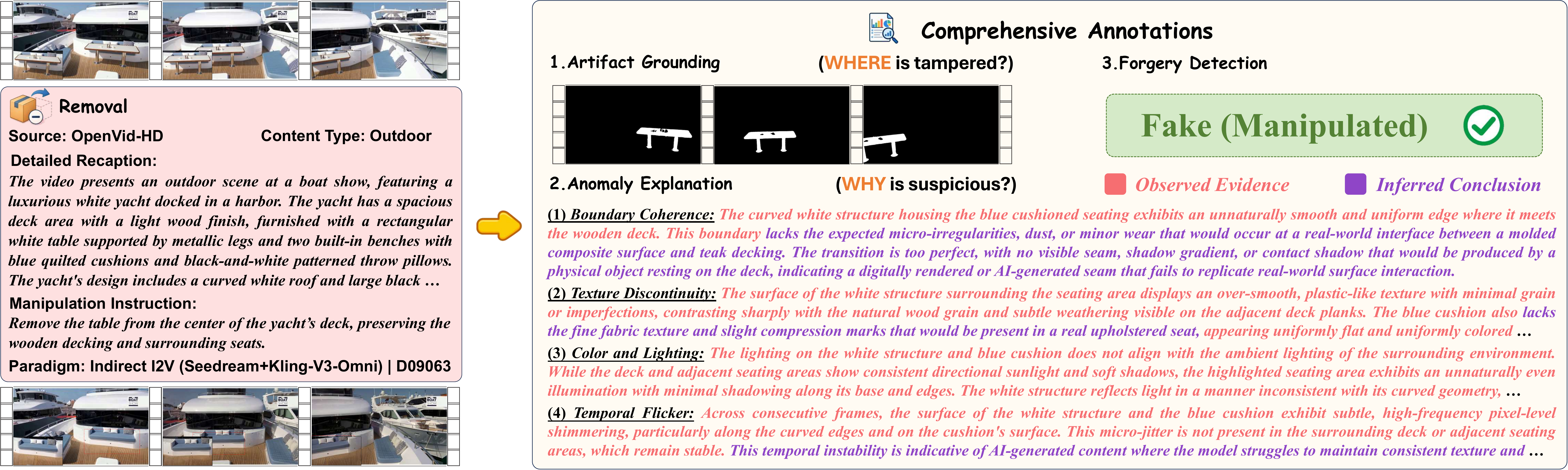}
 
    \caption{\textbf{Examples of \textit{Addition} and \textit{Removal} in ManiVid-38K.} Zoom in for a better view.}
    \vspace{3mm}
    \vspace{-2mm}
    
    \label{fig:dataset_per_task1}
\end{figure*}

\begin{figure*}[t]
    \centering
    \includegraphics[width=0.94\textwidth]{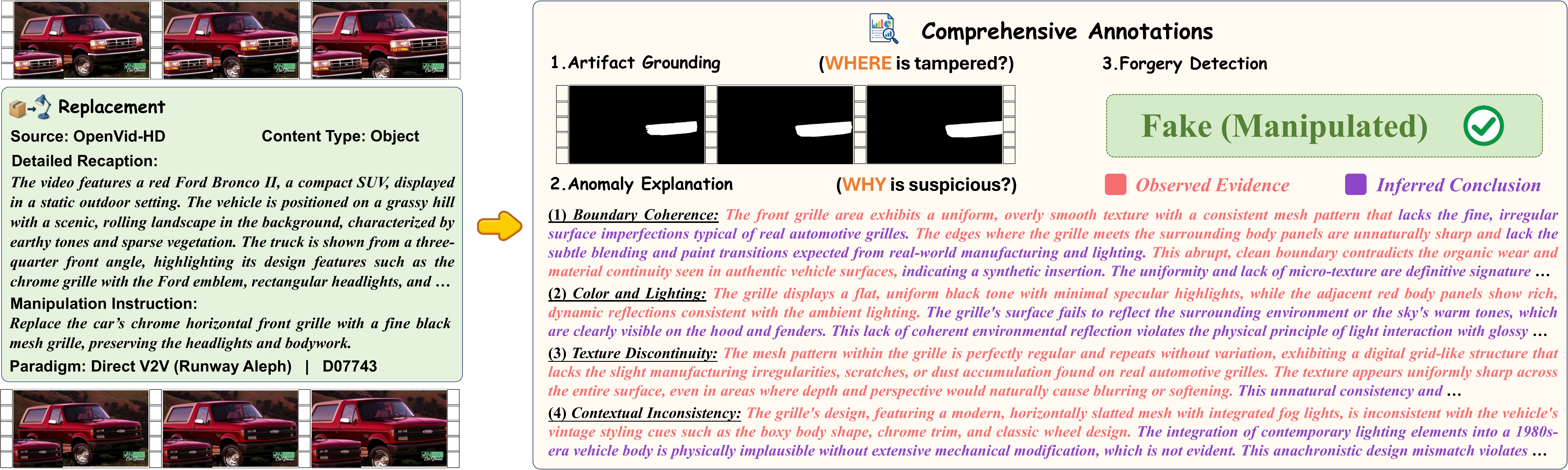}

    \vspace{2mm}


    \includegraphics[width=0.94\textwidth]{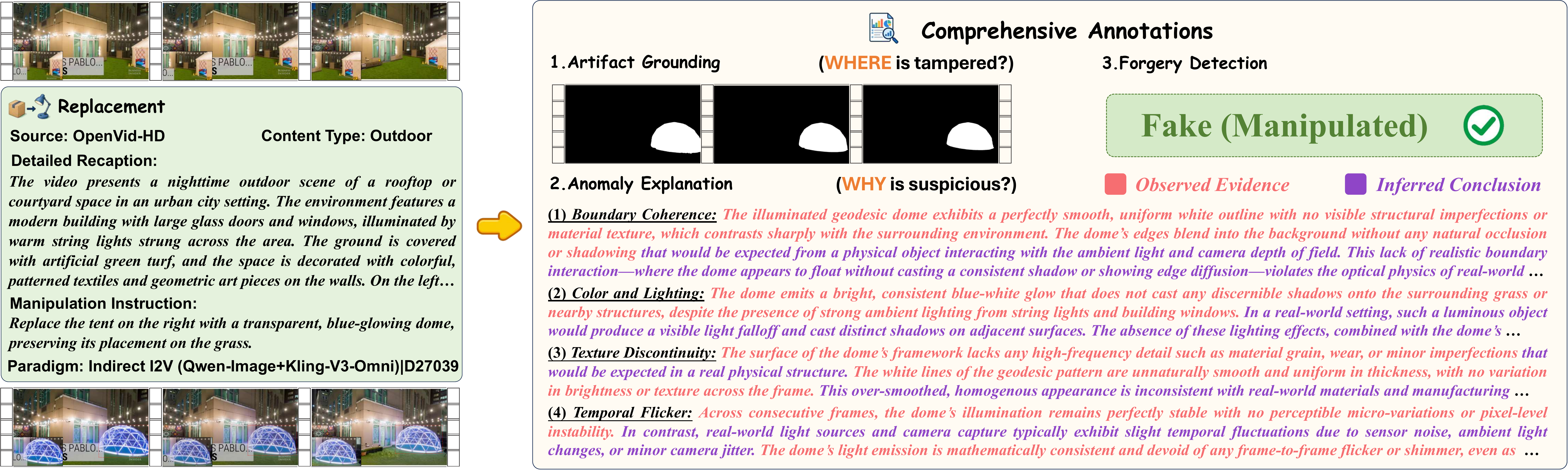}
    
    \parbox[c][1mm][c]{\textwidth}{%
    \centering
    \raisebox{1mm}{%
        \hdashrule{0.94\textwidth}{0.6pt}{3pt 2pt}%
    }%
    }
    \vspace{-3mm}

    \includegraphics[width=0.94\textwidth]{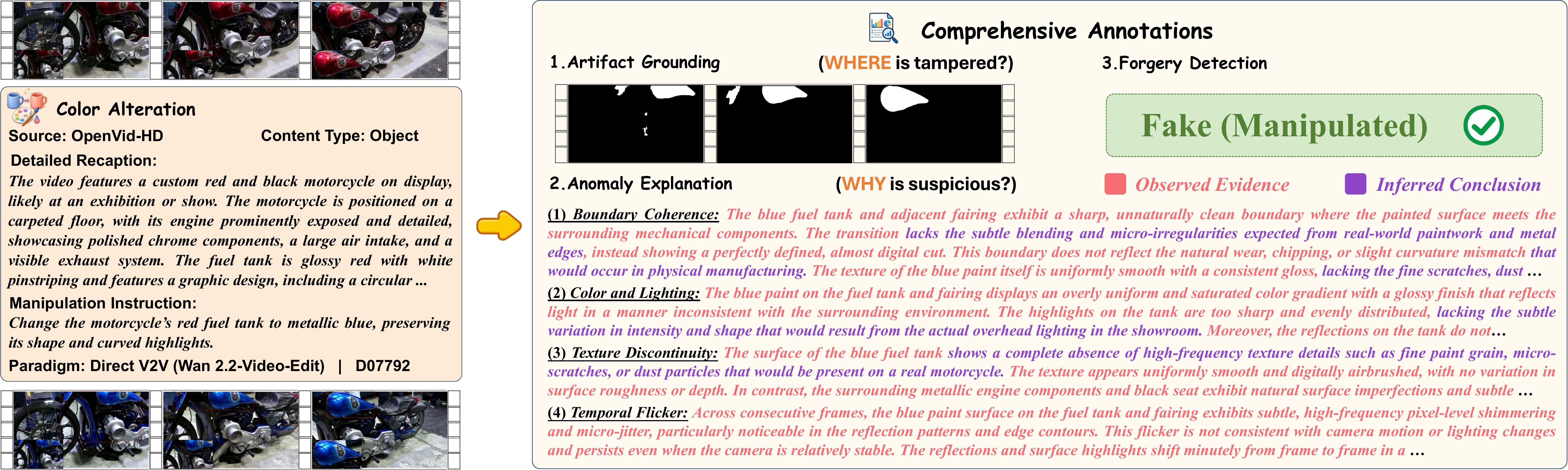}

    \vspace{2mm}


    \includegraphics[width=0.94\textwidth]{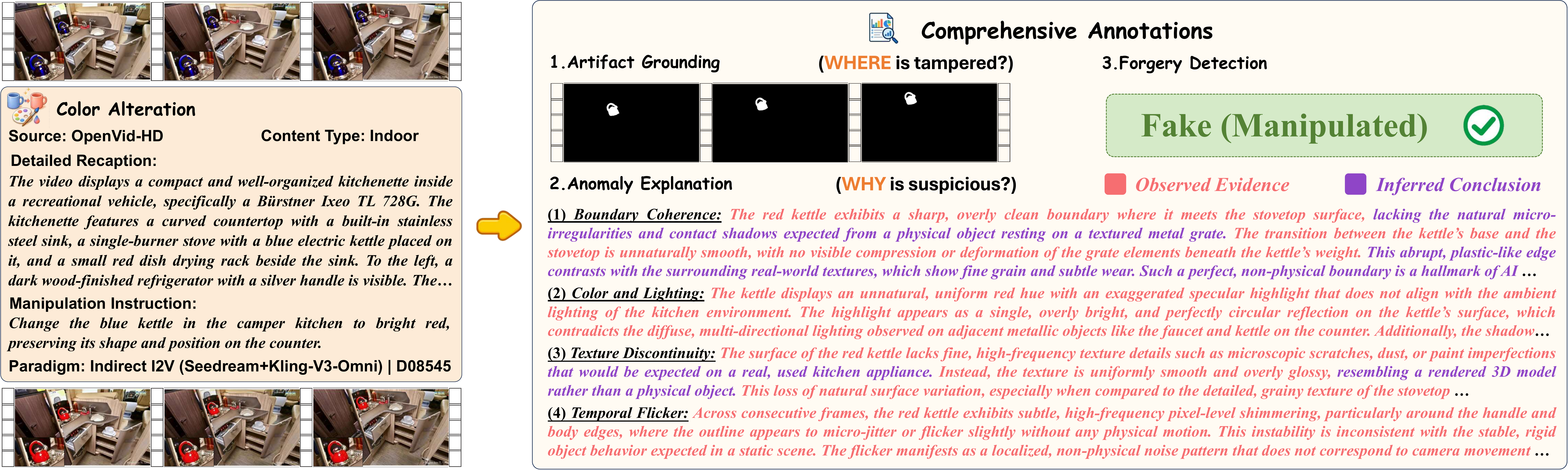}
    \caption{\textbf{Examples of \textit{Replacement} and \textit{Color Alteration} in ManiVid-38K.} Zoom in for a better view.}
    \vspace{3mm}
    \vspace{-4.5mm}
    
    \label{fig:dataset_per_task2}
\end{figure*}

\begin{figure*}[t]
    \centering
    \includegraphics[width=0.94\textwidth]{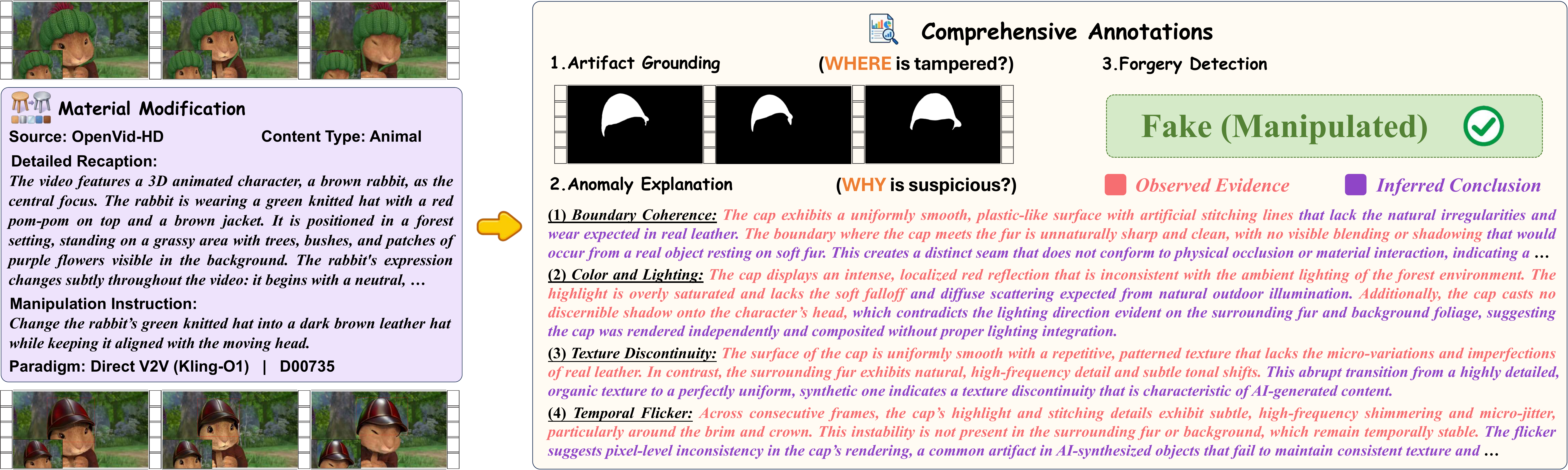}

    \vspace{2mm}


    \includegraphics[width=0.94\textwidth]{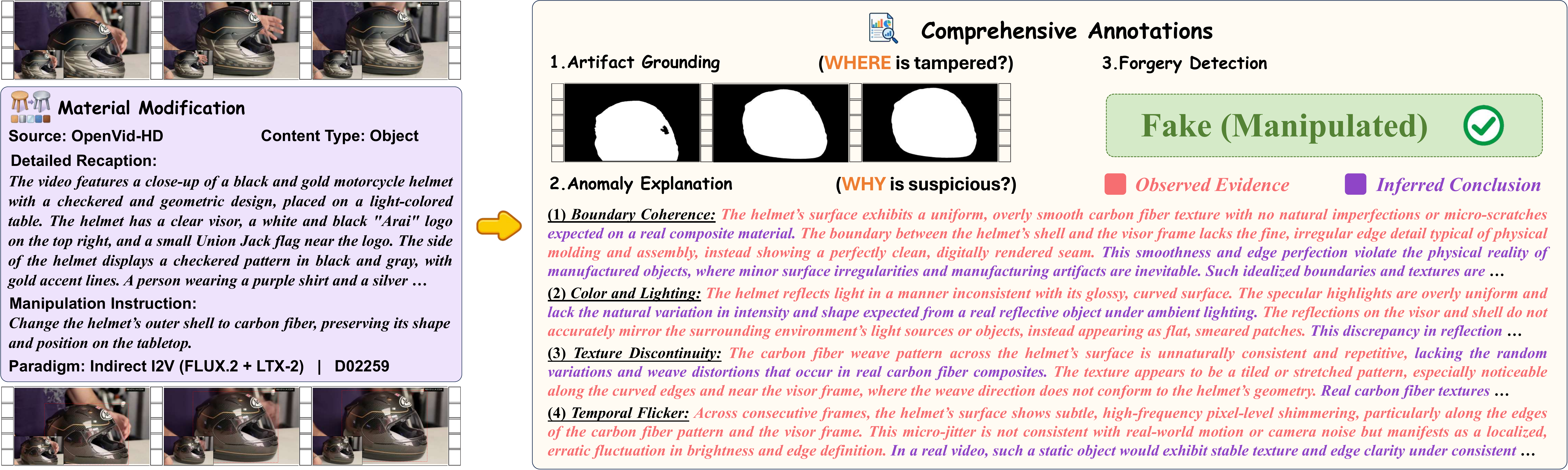}
    
    \parbox[c][1mm][c]{\textwidth}{%
    \centering
    \raisebox{1mm}{%
        \hdashrule{0.94\textwidth}{0.6pt}{3pt 2pt}%
    }%
    }
    \vspace{-3mm}

    \includegraphics[width=0.94\textwidth]{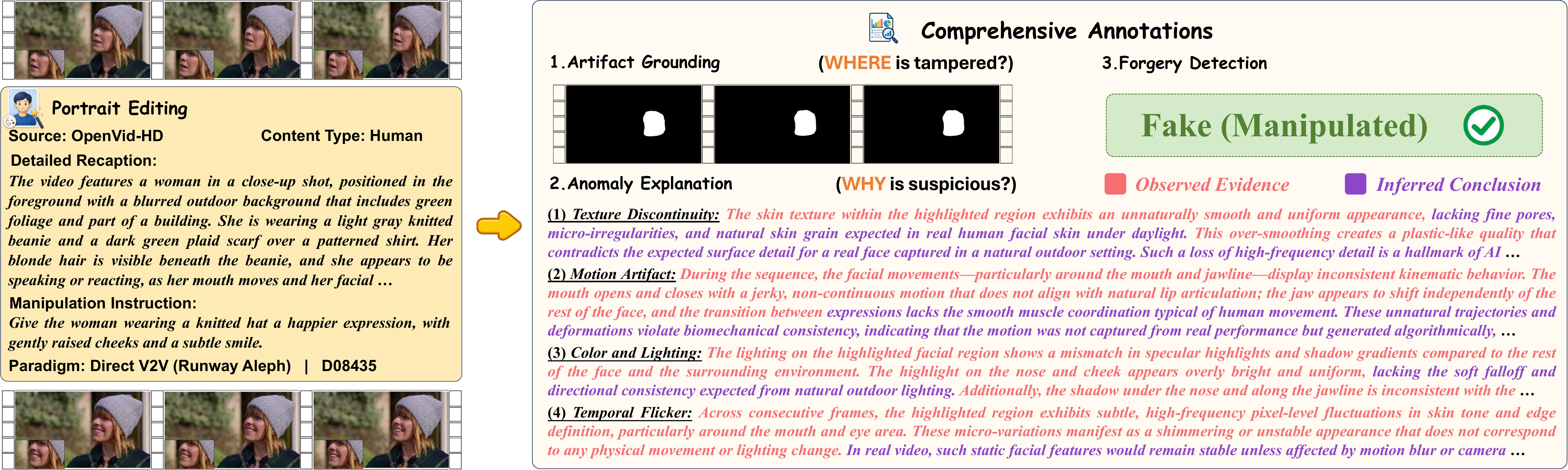}

    \vspace{2mm}


    \includegraphics[width=0.94\textwidth]{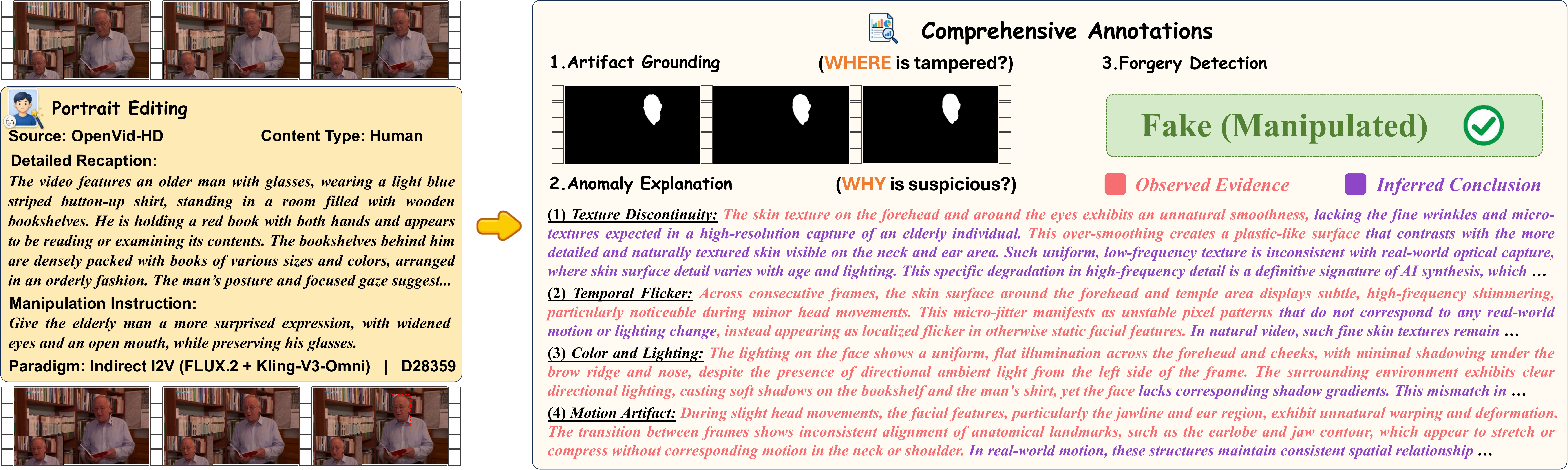}
    \caption{\textbf{Examples of \textit{Material Modification} and \textit{Portrait Editing} in ManiVid-38K.} Zoom in for a better view.}
    \vspace{3mm}
    \vspace{-2mm}
    
    \label{fig:dataset_per_task3}
\end{figure*}

Figs.~\ref{fig:dataset_per_task1}, \ref{fig:dataset_per_task2}, and \ref{fig:dataset_per_task3} presents qualitative examples of different manipulation tasks in the ManiVid-38K.

\FloatBarrier
\subsection{ManiVidBench}
\begin{figure*}[htb]
    \centering
    \vspace{-4mm}
    \includegraphics[width=\textwidth]{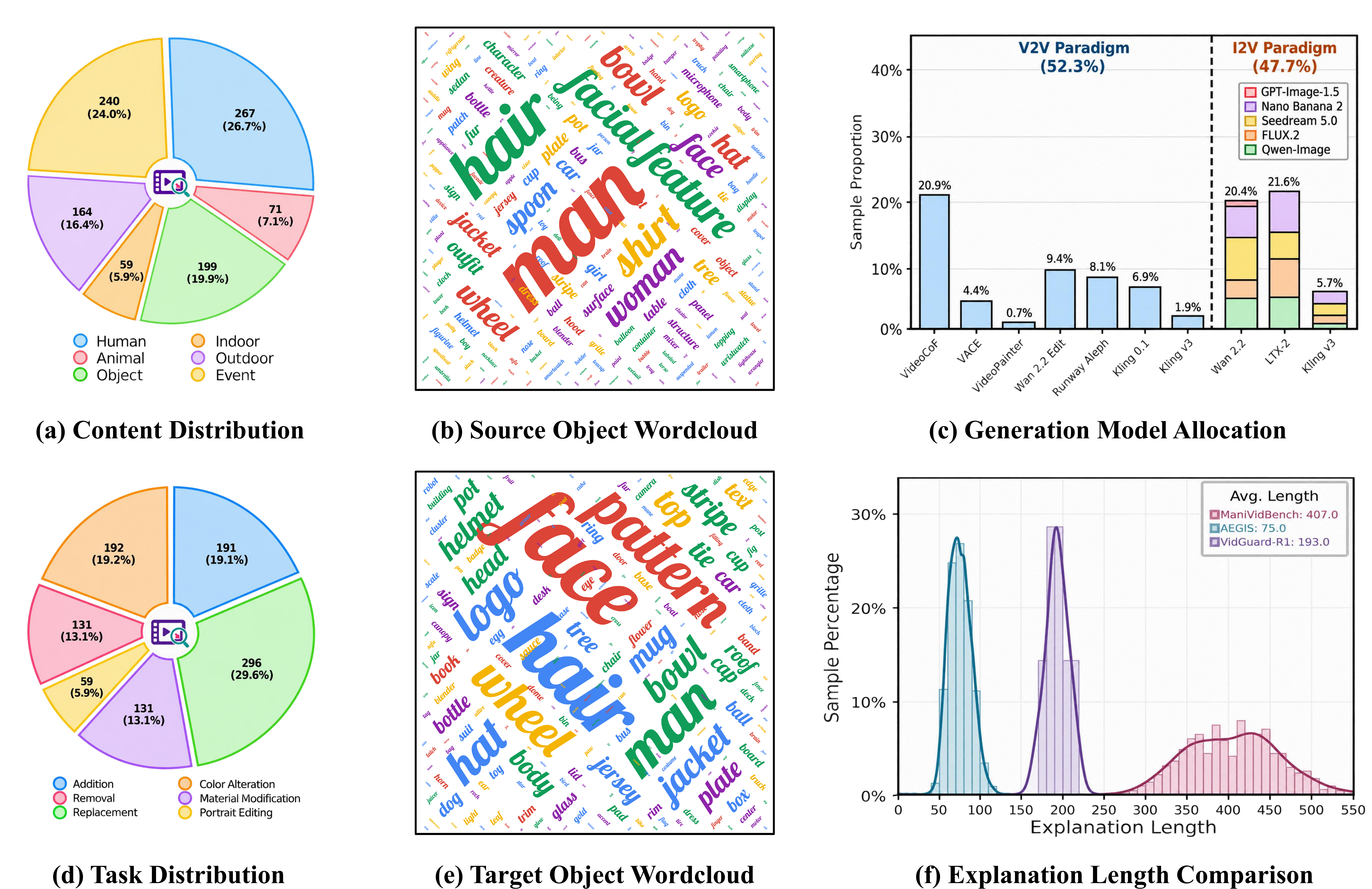}
    \vspace{-5mm}
    \caption{\textbf{Statistical Analysis of Our ManiVidBench.}}
    \vspace{3mm}
    \parbox[t]{0.96\textwidth}{\small\textbf{Reading the benchmark statistics.} ManiVidBench contains 1K real--fake pairs covering the six manipulation categories and preserves the annotation structure needed for detection, grounding, and explanation. The distributions summarize the benchmark composition used for the three-task evaluation.}
    
    \label{fig:manividbench}
\end{figure*}

We randomly sample 1K real--fake video pairs (2K videos) from ManiVid-38K dataset to construct ManiVidBench for evaluating ManiVid.
Since source-level deduplication is performed before dataset splitting, each original source video contributes only one clip to the entire dataset, ensuring source-video-level disjointness between ManiVidBench and the training split.
Detailed statistics are presented in Fig.~\ref{fig:manividbench}. We believe that this dedicated benchmark for manipulated videos, equipped with comprehensive annotations, can provide a solid foundation for future research on manipulated video detection and analysis.

\FloatBarrier
\section{ManiVidLens Framework}
\label{app:manividlens}

\FloatBarrier
\subsection{Task-Specific Input and Output Protocol}
\label{app:manividlens_protocol}

ManiVidLens uses one shared multimodal large language model (MLLM) with separate protocols for manipulation classification, anomaly explanation, and artifact grounding through SAM2~\citep{ravi2025sam}.
Each protocol receives the same 5 temporally ordered conditioning frames and their frame-aligned forensic evidence.
Under the full Forensic Evidence Router (FER) configuration, every conditioning frame contributes one Qwen3-VL visual block~\citep{bai2025qwen3} followed by three $[\mathrm{AIDE}]$ positions.
These positions contain the projected global token, semantic token, and aggregated forensic token from the shared evidence bank.
Task-specific instructions isolate the three output spaces while preserving shared visual evidence and model parameters.

Throughout this appendix, bracketed uppercase names such as $[\mathrm{SEG}]$ denote literal special tokens, while unbracketed names denote tasks, modules, or mathematical indices.

\begin{table}[!htbp]
    \centering
    \vspace{-4mm}
    \caption{\textbf{Task-specific generation protocols used by ManiVidLens.} Each row summarizes one instruction objective and its exact assistant response structure.}
    \vspace{1mm}
    \label{tab:manividlens_task_protocol}
    \small
    \setlength{\tabcolsep}{3pt}
    \begin{tabularx}{\columnwidth}{@{}lX@{}}
        \toprule[1.5pt]
        Task protocol & Instruction objective and supervised assistant response format \\
        \midrule
        Classification & Determine the manipulation status and one applicable manipulation type. The response contains \texttt{Manipulation status:} followed by \texttt{Manipulation type:}. \\
        Explanation & Describe the relevant visual, physical, and temporal anomalies. The response contains exactly one \texttt{<think>...\allowbreak</think>} reasoning block. \\
        Segmentation & Locate the manipulated region and describe its target briefly. A positive response contains the target fields and exactly one $[\mathrm{SEG}]$ marker. \\
        \bottomrule[1.5pt]
    \end{tabularx}
    \vspace{-5mm}
\end{table}

For a manipulated sample, the segmentation target first reports \texttt{Target object:} followed by the annotated object name.
It then places a concise grounding phrase inside \texttt{<p>...\allowbreak</p>} and appends $[\mathrm{SEM}][\mathrm{BBOX1}][\mathrm{BBOX2}]$ immediately before the closing tag.
The grounding phrase provides compact localization context and remains separate from the longer anomaly explanation.
Special token literals occur exclusively in assistant responses, preserving their role as generated control states.

The primary configuration trains and evaluates these three protocols separately on the shared model.
The all-in-one ablation concatenates the same explanation, classification, and segmentation components in that fixed order for direct comparison with the separate task protocol.

\begin{table*}[htb]\centering
\vspace{-2mm}
\caption{\textbf{Task Prompts Used in the ManiVidLens Framework.}
    \textit{Separate prompts} are the default; the \textit{All-in-One prompt} is used only for the 3-task~ablation.}
    \begin{minipage}{\textwidth}\vspace{0mm}
    \centering
    \begin{tcolorbox}
    \centering
    \fontsize{8pt}{10pt}\selectfont
    \begin{tabular}{p{\dimexpr\linewidth-4\tabcolsep\relax} c}

    \VarSty{{\bf \normalsize Prompts Used in ManiVidLens}} &\\

    \vspace{0.05\baselineskip}

    {\small\bfseries Separate Prompts (Default):}

    \vspace{0.3\baselineskip}

    {\bfseries Task 1: Artifact Grounding}

    Locate and segment the AI-manipulated region in this video.
    If a manipulated region is present, describe the target object and output exactly one segmentation marker for that region.
    If no manipulated region is present, state that the manipulated area is none and do \textbf{not} output a segmentation marker.
    Do \textbf{not} output classification fields or a forensic explanation.

    \vspace{0.5\baselineskip}

    {\bfseries Task 2: Anomaly Explanation}

    Analyze this video for forensic evidence of AI manipulation.
    Explain the relevant visual, physical, and temporal evidence inside exactly one reasoning block.
    Do \textbf{not} output classification fields, a target-object field, or a segmentation marker.

    \vspace{0.5\baselineskip}

    {\bfseries Task 3: Forgery Detection}

    Analyze this video for AI manipulation.
    Determine whether it is manipulated or original.
    If manipulated, identify the manipulation type as
    \textbf{\textit{Addition}}, \textbf{\textit{Replacement}}, \textbf{\textit{Material Modification}},
\textbf{\textit{Removal}}, \textbf{\textit{Color Alteration}}, or \textbf{\textit{Portrait Editing}}.
    If original, use manipulation type \textit{\textbf{Real}}.

    \vspace{0.8\baselineskip}

    {\small\bfseries All-in-One Prompt:}

    \vspace{0.2\baselineskip}

    Analyze this video for AI manipulation.
    Follow these three steps in order.
    First, explain the relevant visual, physical, and temporal forensic evidence inside exactly one reasoning block.
    Second, classify the video as manipulated or original and provide the manipulation type;
    if it is original, use type \textit{\textbf{Real}}.
    Third, localize and segment the manipulated region, or state that the manipulated area is none for an original video.
    Do \textbf{not} omit, merge, or reorder these three steps.

    \end{tabular}
    \end{tcolorbox}
    \vspace{-2mm}

    \label{app:tab:ours_prompts}

    \end{minipage}
\end{table*}

\FloatBarrier
\subsection{Forensic Evidence Router}
\label{app:manividlens_fer}

\FloatBarrier
\subsubsection{Forensic Evidence Bank}

The frozen AIDE encoder~\citep{yan2025sanity} produces one 256-dimensional semantic token $\mathbf{s}_k$ and four 2,048-dimensional forensic tokens $\{\mathbf{r}_{k,j}\}_{j=1}^{4}$ for frame $I_{t_k}$.
Separate linear adapters map the semantic and forensic tokens to $d_e=256$.
A two-layer projector with a 1,024-dimensional hidden layer constructs a global seed from $[\mathbf{s}_k;\bar{\mathbf{r}}_k]$, where $\bar{\mathbf{r}}_k$ is the mean forensic token.
With $\mathbf{T}_{\mathrm{bank}}$ denoting the learned slot embedding, the evidence bank is computed as
\begin{equation}
    \label{eq:app_forensic_bank}
    \begin{aligned}
    \mathbf{U}_k
    &= [\phi_s(\mathbf{s}_k),\{\phi_{r,j}(\mathbf{r}_{k,j})\}_{j=1}^{4},
    \phi_g([\mathbf{s}_k;\bar{\mathbf{r}}_k])] + \mathbf{T}_{\mathrm{bank}},\\
    \mathbf{U}'_k
    &= \mathbf{U}_k + \operatorname{MHA}(\operatorname{LN}(\mathbf{U}_k)),\\
    \mathbf{E}_k
    &= \mathbf{U}'_k + \operatorname{FFN}(\operatorname{LN}(\mathbf{U}'_k))
    \in \mathbb{R}^{6\times256}.
    \end{aligned}
\end{equation}
The resulting order is semantic, four forensic, and global.
This fixed ordering is retained throughout the Forensic Evidence-to-VLM (FE2VLM) adapter, the auxiliary estimator, and the Forensic Evidence-to-SAM (FE2SAM) adapter across both training stages.

\FloatBarrier
\subsubsection{Auxiliary Authenticity Estimator}

The auxiliary estimator uses only the contextualized global and semantic tokens to predict whether a video is manipulated.
For each frame, their normalized features are concatenated and passed through a two-layer MLP with a 256-dimensional hidden layer to obtain a frame logit $u_{i,k}$.
Frame logits are aggregated into a video-level manipulation logit $u_i$ by a normalized smooth maximum, and the corresponding routing score $a_i$ is
\begin{equation}
    \label{eq:app_auth_router}
    \begin{aligned}
    u_i
    &= \tau\!\left[\log\!\sum_{k=1}^{K_i}
    \exp(u_{i,k}/\tau)-\log K_i\right],\\
    a_i
    &= \rho+(1-\rho)\sigma(u_i),
    \qquad \tau=0.5,\;\rho=0.25.
    \end{aligned}
\end{equation}
A binary cross-entropy loss supervises $u_i$, with manipulated videos as the positive class.
Detached gates from this score modulate FE2VLM for every task and FE2SAM for grounding.
\mbox{Gradient detachment} lets the classification objective determine the routing score directly.
The routing floor $\rho$ retains a nonzero forensic contribution throughout both optimization stages.

\FloatBarrier
\subsubsection{Evidence Injection into the MLLM}

FE2VLM selects the global token, semantic token, and mean of the four forensic tokens from $\mathbf{E}_k$.
After adding three learned type embeddings, layer normalization and a two-layer $256\!\rightarrow\!1024\!\rightarrow\!d_{\mathrm{MLLM}}$ projector map them to the MLLM hidden space.
Let $\mathbf{v}_{i,k}$ denote these projected tokens and $\mathbf{e}_{\mathrm{AIDE}}$ the learned embedding of the $[\mathrm{AIDE}]$ token.
The injected evidence is softly routed as
\begin{equation}
    \label{eq:app_vlm_route}
    \widetilde{\mathbf{v}}_{i,k}
    = \mathbf{e}_{\mathrm{AIDE}}
    + a_i(\mathbf{v}_{i,k}-\mathbf{e}_{\mathrm{AIDE}}).
\end{equation}
The three routed tokens replace the three $[\mathrm{AIDE}]$ positions following the corresponding frame's visual block.
The MLLM therefore receives routed forensic evidence directly inside its multimodal token sequence for every conditioning frame.

\FloatBarrier
\subsubsection{Conditional SAM2 Feature Enhancement}

FE2SAM uses five evidence tokens per frame: the global token followed by all four forensic tokens.
The semantic token is excluded because semantic object guidance is already supplied by the Prompt Distill Module (PDM) prompts and SAM2 image features.
Learned type embeddings, layer normalization and a $256\!\rightarrow\!256\!\rightarrow\!256$ MLP map these tokens to the SAM2 space.

For each conditioning frame, let $\mathbf{F}_{i,k}\in\mathbb{R}^{d_s\times H_s\times W_s}$ denote the SAM2 image feature map and $\mathbf{A}_{i,k}\in\mathbb{R}^{5\times d_s}$ the projected global and 4 forensic tokens, where $d_s=256$. The $H_sW_s$ spatial positions of $\mathbf{F}_{i,k}$ serve as queries, while $\mathbf{A}_{i,k}$ provides keys and values for 4-head cross-attention.
A convolutional source adapter further refines the attended features, and the detached manipulation score interpolates between the original SAM2 features and their evidence-enhanced counterparts:
\begin{equation}
    \label{eq:app_sam_route}
    \begin{aligned}
    \mathbf{H}_{i,k}
    &= \mathbf{F}_{i,k}
    + \psi\!\left(\operatorname{MHA}(\mathbf{F}_{i,k},\mathbf{A}_{i,k},\mathbf{A}_{i,k})\right),\\
    \widetilde{\mathbf{F}}_{i,k}
    &= \mathbf{F}_{i,k}
    + a_i\!\left(\operatorname{Adapter}(\mathbf{H}_{i,k})-\mathbf{F}_{i,k}\right).
    \end{aligned}
\end{equation}
Here, $\psi$ reshapes the attended tokens and applies a $1\!\times\!1$ convolution, layer normalization, GELU, and another $1\!\times\!1$ convolution, while $\operatorname{Adapter}$ denotes the subsequent residual convolutional refinement block.
Positional encodings and a learned query bias are added to the SAM2 queries before attention.
During Stage~2 training, FE2SAM processes the 5 sampled frames of manipulated segmentation instances carrying mask supervision.
During validation and inference, FE2SAM enhances the 5 conditioning frames whenever generated segmentation states activate mask decoding; SAM2 then propagates the initialized object state across all remaining physical video frames.

\FloatBarrier
\subsection{Prompt Distill Module}
\label{app:manividlens_pdm}

\FloatBarrier
\subsubsection{Semantic Dense Prompt Fusion}

PDM reads the final-layer MLLM state associated with each grounding token.
An MLP with a ReLU activation maps the $[\mathrm{SEM}]$ state from $d_{\mathrm{MLLM}}$ into the 256-dimensional SAM2 prompt space.
The projected semantic state is spatially broadcast and added to the frozen SAM2 no-mask embedding.
A lightweight fusion block applies a $1\!\times\!1$ convolution, two-dimensional layer normalization, GELU, and another $1\!\times\!1$ convolution to produce the dense semantic prompt:
\begin{equation}
    \label{eq:app_sem_dense}
    \mathbf{D}_i
    = \operatorname{ConvFuse}\!\left(
    \mathbf{E}_{\mathrm{no\text{-}mask}}
    + \operatorname{Tile}(\phi_{\mathrm{sem}}(\mathbf{h}^{\mathrm{SEM}}_i))
    \right).
\end{equation}
The resulting dense prompt supplies video-level semantic context to every grounding frame.

\FloatBarrier
\subsubsection{Geometric Sparse Prompt Distillation}

Independent 2-layer MLPs with ReLU activations project the $[\mathrm{BBOX1}]$, $[\mathrm{BBOX2}]$, and $[\mathrm{SEG}]$ states from $d_{\mathrm{MLLM}}$ through a hidden layer of dimension $d_{\mathrm{MLLM}}$ into the 256-dimensional SAM2 prompt space.
Each sampled-frame annotation stores an exclusive-coordinate box $(x_0,y_0,x_1,y_1)$ at the original frame resolution.
We resize each valid box to the $1024\!\times\!1024$ SAM2 input space and convert its bottom-right boundary into inclusive pixel coordinates for consistent geometric supervision.
For sample $i$, $\mathcal{F}_i$ indexes the conditioning frames with valid boxes, and $\mathbf{B}_{i,k}$ denotes the corresponding resized box after this coordinate conversion.
We take the minimum bounding rectangle $\bar{\mathbf{B}}_i$ of these boxes, and the frozen SAM2 prompt encoder maps its top-left and bottom-right corners to the corner embeddings $\bar{\mathbf{b}}_{i,1}$ and $\bar{\mathbf{b}}_{i,2}$:
\begin{equation}
    \label{eq:app_box_target}
    \bar{\mathbf{B}}_i
    = \Bigl(\min_{k\in\mathcal{F}_i} x^{0}_{i,k},\ \min_{k\in\mathcal{F}_i} y^{0}_{i,k},\ \max_{k\in\mathcal{F}_i} x^{1}_{i,k},\ \max_{k\in\mathcal{F}_i} y^{1}_{i,k}\Bigr),
    \qquad
    \bar{\mathbf{b}}_{i,j}
    = \operatorname{BoxEnc}(\bar{\mathbf{B}}_i)_j,
    \quad j\in\{1,2\},
\end{equation}
where $(x^{0}_{i,k},y^{0}_{i,k})$ and $(x^{1}_{i,k},y^{1}_{i,k})$ are the top-left and bottom-right corners of $\mathbf{B}_{i,k}$.
Samples with a complete latent span and at least one valid box contribute to Eq.~\ref{eq:prompt-distill-loss}.
The projected $[\mathrm{BBOX1}]$ and $[\mathrm{BBOX2}]$ states are concatenated with the projected $[\mathrm{SEG}]$ state in the order $[\mathrm{BBOX1}]\!\rightarrow\![\mathrm{BBOX2}]\!\rightarrow\![\mathrm{SEG}]$.
Repeating this sparse prompt across all grounding frames preserves consistent geometric guidance from the shared video-level prior during mask decoding.

\FloatBarrier
\subsection{Training and Inference Pipelines}
\label{app:manividlens_train_val}

\begin{figure}[H]
    \centering
    \includegraphics[width=\linewidth]{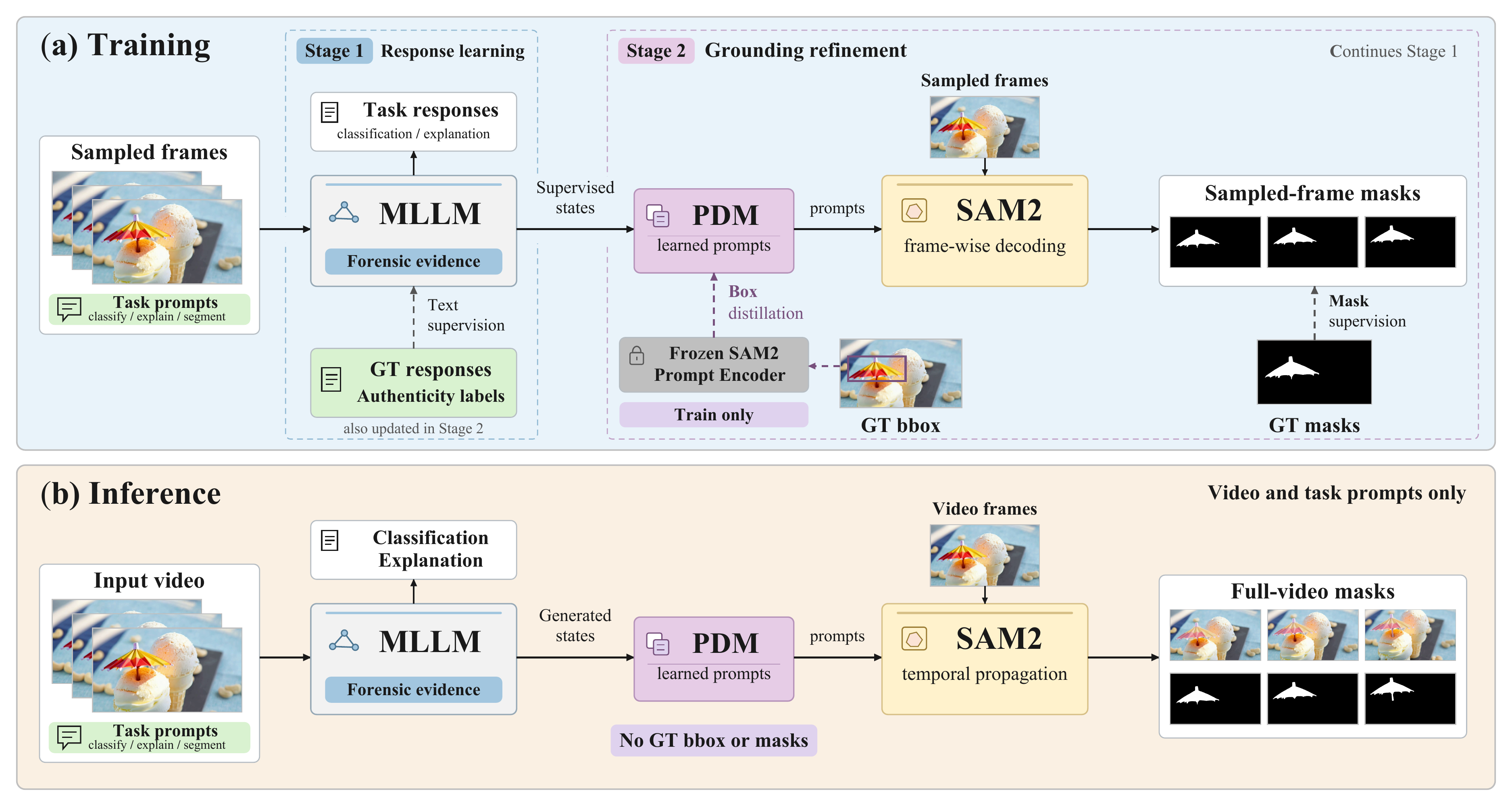}
    \vspace{-0.5cm}
    \caption{\textbf{Training and Inference Pipelines of ManiVidLens.}
    (a) Stage~1 learns task-specific responses and authenticity routing.
    Stage~2 adds box-prompt distillation and sampled-frame mask supervision, with PDM reading teacher-forced token states.
    (b) Inference constructs SAM2 prompts from generated token states and propagates masks across the full video using only video and task inputs.
    GT denotes ground-truth responses, boxes, and masks used for supervision.}
    \label{fig:training_inference_pipeline}
\end{figure}

\FloatBarrier
\subsubsection{Training Pipeline}
\label{app:manividlens_training}

\begin{table}[H]
    \centering
    \vspace{-4mm}
    \caption{\textbf{Two-stage training implementation for the current full profile.} The table reports the launcher defaults and stage-specific optimization settings.}
    \vspace{1mm}
    \label{tab:two_stage_training}
    \small
    \setlength{\tabcolsep}{3pt}
    \begin{tabularx}{\columnwidth}{@{}l>{\raggedright\arraybackslash}X>{\raggedright\arraybackslash}X@{}}
        \toprule[1.5pt]
        Training aspect & Stage~1 implementation details & Stage~2 implementation details \\
        \midrule
        Stage length & One complete training epoch & One complete training epoch \\
        Task setup & Three separate task instances with $1/3$ shared-loss scaling & Three separate task instances with $1/3$ shared-loss scaling \\
        Batch ratio & Balanced original-to-manipulated ratio of $1{:}1$ & Manipulation-focused original-to-manipulated ratio of $1{:}3$ \\
        Shared weights & Language and classification weights are $\lambda_{\mathrm{lm}}=1.0$ and $\lambda_{\mathrm{cls}}=1.0$ & Language and classification weights are $\lambda_{\mathrm{lm}}=0.1$ and $\lambda_{\mathrm{cls}}=0.5$ \\
        Grounding weights & Every grounding-loss weight remains zero & $\lambda_m=2.0$, $\lambda_d=1.0$, and $\lambda_c=1.0$ \\
        Class weight & The classifier uses unit positive-class weight & The classifier uses one-third positive-class weight \\
        Trainable set & PDM, FE2SAM, the $[\mathrm{SEG}]$ projector, and the SAM2 mask decoder remain frozen & PDM, FE2SAM, the $[\mathrm{SEG}]$ projector, and the SAM2 mask decoder are trainable \\
        \bottomrule[1.5pt]
    \end{tabularx}
    \vspace{-3mm}
\end{table}

The Qwen3-VL visual encoder and base language-model weights remain frozen, while rank-$128$ low-rank adaptation (LoRA) modules update the language layers~\citep{hu2021lora}.
The token embeddings and language-model head also remain trainable throughout both stages.
Both stages train the AIDE evidence bank, FE2VLM projector, and auxiliary authenticity estimator.

Stage~1 learns the three response protocols and video-level authenticity routing before mask optimization begins (Figure~\ref{fig:training_inference_pipeline}(a)).
Its frozen grounding components comprise the SAM2 mask decoder, the $[\mathrm{SEG}]$ projector, the FE2SAM projector and enhancer, and every PDM projection and fusion block.
Stage~2 loads the Stage~1 checkpoint and activates these grounding components.
Within SAM2, only the mask decoder is trainable during Stage~2; the image encoder, prompt encoder, memory encoder, and memory attention remain frozen.

Each sampled video view expands into explanation, classification, and segmentation instances under the separate protocol.
The three instances share identical frame indices because their sampling seed derives from the same base view.
Language and auxiliary classification losses receive a factor of $1/3$ on every task instance.
This normalization preserves the aggregate shared-loss weight assigned to each base video view.
Grounding losses receive unit task scaling only for Stage~2 segmentation instances.
Mask availability further restricts these losses to manipulated samples carrying valid spatial annotations for the sampled video view.
Teacher-forced assistant targets supervise every textual response and provide the special token states used by PDM.
Box annotations construct the cosine-distillation targets, while mask annotations supervise the mask and Dice objectives.
During Stage~2 training, mask decoding operates independently on the five sampled frames for manipulated segmentation instances without temporal video-memory propagation.

The grouped sampler constructs device-local batches that are homogeneous in concrete task and video type.
For manipulated samples, let $r_i$ denote the largest visible mask-area ratio among the sampled frames.
The per-sample auxiliary classification weight follows
\begin{equation}
    \label{eq:app_small_region_weight}
    w_i =
    \begin{cases}
        1+(0.05-r_i)/0.05, & \text{if } 0<r_i<0.05,\\
        1, & \text{for every remaining training sample}.
    \end{cases}
\end{equation}
This weighting assigns additional classification emphasis to small visible manipulated regions.

\FloatBarrier
\subsubsection{Inference Pipeline}
\label{app:manividlens_inference}

Validation and inference perform separate deterministic generations for explanation, classification, and segmentation (Figure~\ref{fig:training_inference_pipeline}(b)).
Every generation receives the same ordered conditioning frames and their corresponding routed AIDE evidence.
Forgery-detection metrics are computed from the generated classification response, while the auxiliary estimator is logged separately and supplies the routing gate. Stage~1 validation omits mask decoding altogether.
During Stage~2 validation and inference, a segmentation response containing $[\mathrm{SEG}]$ activates PDM and SAM2.
PDM reads the aligned, autoregressively generated $[\mathrm{SEM}]$, $[\mathrm{BBOX1}]$, $[\mathrm{BBOX2}]$, and $[\mathrm{SEG}]$ states.
The projected $[\mathrm{SEM}]$ state forms the dense semantic prompt through Eq.~\ref{eq:app_sem_dense}, while the projected $[\mathrm{BBOX1}]$, $[\mathrm{BBOX2}]$, and $[\mathrm{SEG}]$ states form the sparse prompt embeddings for every conditioning frame.
Validation annotations remain outside the model inputs and support metric computation exclusively.
Stage~2 validation permits at most one additional generation of segmentation responses per sample.

The 5 conditioning frames follow the policy in Appendix~\ref{app:frame_sampling_scope}, while the evaluation sequence contains every physical frame of the complete video.
FE2SAM injects frame-aligned forensic evidence into the SAM2 image features of the 5 conditioning frames.
The PDM prompt embeddings initialize the object state on the same conditioning frames before temporal propagation begins.
With the earliest conditioning frame at local index zero, SAM2 propagates the initialized object state once in the forward temporal direction through the complete physical-frame sequence.
The implementation resizes propagated mask logits bilinearly to each original frame resolution and thresholds sigmoid probabilities at $0.5$.
If the response omits $[\mathrm{SEG}]$, the pipeline skips mask decoding.

\FloatBarrier
\section{Experiments}

\FloatBarrier
\subsection{Implementation Details}
\label{app:experiment:details}

\FloatBarrier
\subsubsection{Model Setup \& Dataset}
\label{app:benchmark_baselines}

\noindent\textbf{Model Setup.}
ManiVidLens combines Qwen3-VL-4B-Instruct~\citep{bai2025qwen3} for multimodal reasoning, a frozen AIDE encoder~\citep{yan2025sanity} for forensic cues, and SAM2 Hiera-L~\citep{ravi2025sam} for video segmentation.
All three components are initialized from their respective released pretrained checkpoints before any task-specific training is performed.

\noindent\textbf{Dataset.}
Training uses the ManiVid-38K dataset training split of 18K real--fake video pairs spanning six content categories and six manipulation tasks.
Each pair provides a video-level authenticity label, frame-level pixel masks, and a textual anomaly explanation.
ManiVidBench contains 2K videos sampled from ManiVid-38K dataset: 1K manipulated videos and their authentic videos across the same six manipulation categories, with task-specific evaluation scopes defined below.

\FloatBarrier
\subsubsection{Training and Inference Settings}
\label{app:frame_sampling_scope}
\label{app:optimization_setup}

\noindent\textbf{Training.}
In the reported full configuration, Stage~1 learns the three task protocols for one epoch with grounding frozen; Stage~2 loads that checkpoint and trains grounding for one epoch with lower language and classification weights.
Each video yields five views per epoch, and every view expands into three task instances sharing five ordered frames sampled with random temporal patterns and a stride from $\{4,6,8,10\}$.
Invalid windows fall back to five randomly selected frames from the complete video, sorted temporally.
Stage~1 samples original and manipulated videos at $1{:}1$, whereas Stage~2 uses $1{:}3$.
Both stages use AdamW with a base learning rate of $10^{-4}$, weight decay of $0.05$, a $5\%$ linear warmup, and cosine decay.
With bfloat16 precision on eight NVIDIA RTX PRO 5000 GPUs, a per-device microbatch of two and four-step accumulation yield an effective batch size of $64$; sequences are capped at $8{,}192$ tokens, and LoRA uses rank $128$, scaling $256$, and dropout $0.05$~\citep{hu2021lora}.
Appendix~\ref{app:manividlens_training} specifies the stage-wise loss weights and trainable modules.
For a consistent budget, trainable image and video expert comparisons receive two epochs on the ManiVid-38K dataset training split for each supported task; training-free comparisons and general MLLMs use released checkpoints without fine-tuning in the reported evaluation protocol.

\noindent\textbf{Inference.}
ManiVidLens generates each task separately and deterministically; only grounding activates SAM2 through $[\mathrm{SEG}]$.
Forgery detection is evaluated on all 2,000 ManiVidBench videos, whereas artifact grounding and anomaly explanation are evaluated on the 1,000 manipulated videos.
For detection, image experts majority-vote $0.5$-threshold predictions over thirty uniformly sampled frames; other methods use native video-level outputs.
For explanation, one response is evaluated per manipulated video; ManiVidLens and general MLLMs use five conditioning frames.
These are the first five decodable candidates from a uniform thirty-frame grid.
ManiVidLens reuses these frames across all three task generations.
For grounding, image experts predict and are scored on the thirty sampled frames.
General MLLMs and Omni-Fake-R1 output frame-indexed boxes, which prompt shared SAM2 Hiera-L propagation and evaluation over every physical frame.
Video segmentation models directly process the complete video, while ManiVidLens initializes full-video mask propagation from its five conditioning frames; both are evaluated on every physical frame of each test video under this complete-video evaluation protocol.

\FloatBarrier
\subsubsection{Evaluation Metrics}
\label{app:evaluation_metrics}

\noindent\textbf{Artifact Grounding.}
Mean IoU (mIoU) computes binary-mask intersection over union for every evaluated frame, averages those scores within each video, and then macro-averages across videos.
This region-overlap metric directly measures how completely each predicted mask covers the reference manipulation area.
J\&F averages region similarity ($J$), which equals the frame IoU, and contour accuracy ($F$) with a boundary tolerance of $0.008$ times the image diagonal under the same two-level aggregation~\citep{yuan2025sa2va}.
This combined score complements regional overlap with boundary fidelity, which is essential for precise pixel-level artifact localization.
Jointly empty mask pairs score one, whereas pairs containing exactly one empty mask score zero.

\noindent\textbf{Anomaly Explanation.}
ROUGE-L is the token-level longest-common-subsequence F1 between the first explicitly closed, nonempty \texttt{<think>} blocks after lowercase alphanumeric tokenization without stemming~\citep{lin2004rouge}.
It measures whether a generated explanation recovers the reference evidence and its ordered lexical content.
Cosine Similarity Score (CSS) computes cosine similarity between attention-mask mean-pooled paraphrase-MiniLM-L6-v2 embeddings that are subsequently L2-normalized~\citep{sentence_transformers_minilm,reimers2019sentence,xu2024fakeshield}.
It captures semantic agreement when a valid explanation paraphrases the reference with different wording.
Both scores are computed per video and then averaged across the manipulated subset; a missing or invalid generated block scores zero, while a missing reference is excluded.

\noindent\textbf{Forgery Detection.}
Accuracy is the proportion of correct video-level authenticity predictions among all 2,000 test videos.
The balanced set of 1,000 authentic and 1,000 manipulated videos makes it a direct measure of overall correctness.
F1 is the harmonic mean of precision and recall with the manipulated class treated as positive.
It complements Accuracy by exposing the trade-off between missed manipulations and false alarms.
Any unparseable generated authenticity label counts as an incorrect prediction during metric computation.

Across the main and ablation tables, \textbf{\texttt{bold}} and \underline{\texttt{underlined}} values indicate the best and second-best results within each table column, respectively.

\FloatBarrier
\subsection{Baseline Model Details}
\label{app:model_details}

\noindent\textbf{CNNSpot}~\citep{wang2020cnn}\quad
CNNSpot is a convolutional image detector that uses a ResNet-50 classifier to distinguish real images from synthetic images.
Training on one source generator with blur and JPEG augmentation helps the detector generalize to unseen generators.

\noindent\textbf{UnivFD}~\citep{ojha2023towards}\quad
UnivFD detects synthetic images using the feature space of a frozen pretrained vision-language model.
A nearest-neighbor rule or linear classifier predicts authenticity from these representations, which are learned from diverse image-text pairs.

\noindent\textbf{AIDE}~\citep{yan2025sanity}\quad
AIDE combines high-level semantic representations with low-level forensic features for AI-generated image detection.
Its local branch samples high- and low-frequency patches to capture noise patterns and other synthesis artifacts.

\noindent\textbf{LEGION}~\citep{kang2025legion}\quad
LEGION is an MLLM-based framework for synthetic-image detection, artifact segmentation, and textual explanation.
It pairs pixel-level artifact masks with natural-language descriptions of the visual inconsistencies in the identified regions.

\noindent\textbf{FakeShield}~\citep{xu2024fakeshield}\quad
FakeShield is an explainable image-forgery framework that predicts authenticity, describes tampering evidence, and localizes manipulated regions.
Its domain-tag-guided detection module handles different forgery types, while a multimodal localization module uses the generated descriptions to guide mask prediction.

\noindent\textbf{SIDA}~\citep{huang2025sida}\quad
SIDA is a large multimodal model for detecting, localizing, and explaining deepfakes in social-media images.
It distinguishes authentic, fully synthetic, and tampered images, producing manipulation masks and textual explanations for its decisions.

\noindent\textbf{D3}~\citep{zheng2025d3}\quad
D3 is a training-free video-detection method that identifies temporal discrepancies through second-order features.
Statistics of changes in inter-frame feature relationships, extracted with a pretrained visual encoder, distinguish real from generated videos.

\noindent\textbf{VidGuard-R1}~\citep{park2025vidguard}\quad
VidGuard-R1 adapts a multimodal language model for video-authenticity detection and explanatory reasoning through group relative policy optimization.
Its reward design accounts for temporal artifacts and generation complexity, encouraging the model to produce authenticity judgments with supporting natural-language rationales.

\noindent\textbf{ReStraV}~\citep{interno2026ai}\quad
ReStraV detects generated videos by examining the geometry of frame trajectories in a pretrained visual representation space.
It summarizes temporal curvature and stepwise distances between frame embeddings, then uses a lightweight classifier to identify deviations from the trajectory statistics of natural videos.

\noindent\textbf{NSG-VD}~\citep{zhang2026physics}\quad
NSG-VD uses normalized spatiotemporal gradients to characterize deviations from natural video dynamics under a probability-flow conservation formulation.
It estimates these features with pretrained diffusion models and measures their maximum mean discrepancy from real-video reference features to obtain a detection score.

\noindent\textbf{DeMamba}~\citep{chen2026demamba}\quad
DeMamba introduces a Detail Mamba module for detecting AI-generated videos through spatial and temporal inconsistencies.
The module augments video representations with Mamba-based detail modeling, providing a plug-in component for authenticity classification across different generators and video degradations.

\noindent\textbf{VideoLISA}~\citep{bai2024one}\quad
VideoLISA combines a video-language model with a segmentation model to generate video masks from language instructions that require contextual reasoning.
Its sparse-dense frame sampling balances temporal coverage with spatial detail, while a shared tracking token supports segmentation of the target across multiple frames.

\noindent\textbf{VISA}~\citep{yan2024visa}\quad
VISA is a video segmentation assistant that resolves implicit object queries using world knowledge and video context.
A mask decoder segments and tracks the inferred target across video frames.

\noindent\textbf{Sa2VA}~\citep{yuan2025sa2va}\quad
Sa2VA integrates multimodal language understanding with SAM2 for dense grounding in images and videos.
The language model generates instruction tokens that guide mask prediction and video propagation, supporting both grounded conversation and referring segmentation within a shared framework.

\noindent\textbf{Omni-Fake-R1}~\citep{li2026omni}\quad
Omni-Fake-R1 is a reinforcement-learning-based detector for deepfakes across image, audio, video, and audiovisual inputs.
It combines visual and auditory evidence when available and generates structured authenticity decisions, localization outputs, and natural-language explanations; its visual localization is expressed as bounding boxes.

\noindent\textbf{InternVL3.5-30B-A3B}~\citep{wang2025internvl3}\quad
InternVL3.5-30B-A3B is an open multimodal model that connects an InternViT visual encoder to a mixture-of-experts language model through a multilayer perceptron.
It retains dynamic image-resolution processing and supports image and video understanding, with approximately three billion active language-model parameters per token.

\noindent\textbf{Qwen3.5-35B-A3B}~\citep{qwen3.5}\quad
Qwen3.5-35B-A3B is a multimodal model with a vision encoder and a sparse mixture-of-experts language backbone containing 35 billion parameters, of which three billion are active per token.
Its hybrid architecture combines gated delta networks with attention layers, supporting visual understanding and language reasoning with sparse computation.

\noindent\textbf{GPT-5.6-Sol}~\citep{openai2026gpt56}\quad
GPT-5.6-Sol is a proprietary general-purpose model that accepts text and image inputs and produces textual responses for reasoning and visual-understanding tasks.
In our evaluation, it analyzes sampled video frames to provide authenticity judgments, anomaly explanations, and frame-indexed boxes without task-specific fine-tuning.

\FloatBarrier
\FloatBarrier
\subsection{More Performance Analysis}
\label{app:more_performance_analysis}
\subsubsection{By Manipulation Task}
\label{app:by_manipulation_task}
The category breakdown in Tables~\ref{tab:grounding}--\ref{tab:detection} compares forgery detection, artifact grounding, and anomaly explanation across the six manipulation types. Following the metrics and aggregation in Appendix~\ref{app:evaluation_metrics}, we examine how authenticity decisions, mask recovery, and explanation quality vary for the same manipulation type, extending the overall results in Sec.~\ref{sec:main_results}.

\noindent\textbf{Addition and Replacement.}
On Addition, ManiVidLens reaches 0.444 mIoU and 0.467 J\&F, above Sa2VA's 0.414 and 0.432, together with 0.969 detection accuracy and 0.813 CSS.
On Replacement, it reaches 0.421 mIoU and 0.439 J\&F, improving over LEGION's 0.385 and 0.392, while detection accuracy is 0.885 vs. DeMamba's 0.890.
These results separate precise recovery of the altered object's spatial extent from video-level authenticity discrimination, showing why the benchmark evaluates spatial localization and authenticity decisions as distinct, complementary outputs.

\noindent\textbf{Color and Material.}
These edits preserve object identity while changing visual attributes, placing emphasis on the altered appearance within an otherwise recognizable object.
ManiVidLens reaches 0.487 mIoU on Color and 0.526 on Material, exceeding FakeShield's 0.429 and Sa2VA's 0.399, respectively.
Its J\&F scores of 0.507 and 0.532 accompany CSS scores of 0.772 and 0.795 and detection accuracies of 0.919 and 0.939.
Table~\ref{tab:ablation_grounding} connects this pattern to the architectural design: combining FER and PDM improves both categories over either module alone, supporting complementary roles for forensic evidence and structured prompts when grounding visual attributes.

\noindent\textbf{Portrait Editing.}
ManiVidLens leads grounding at 0.290 mIoU and 0.306 J\&F, as well as detection accuracy at 0.839 vs. DeMamba's 0.831.
Its absolute grounding scores remain lower than those for Addition, Color, and Material.
Explanation exhibits a different ranking: ManiVidLens leads ROUGE-L at 0.545 vs. SIDA's 0.251, but its CSS of 0.699 trails SIDA's 0.746 and LEGION's 0.718.
The differing rankings separate lexical overlap from semantic agreement with reference explanations, motivating complementary evaluation of the words used to describe an anomaly and the meaning conveyed by the generated forensic description.

\noindent\textbf{Removal.}
ManiVidLens reaches 0.947 detection accuracy and 0.773 CSS, alongside 0.135 mIoU vs. Qwen3.5's 0.221.
The reference mask describes the removed object's original extent, whereas the observed region contains reconstructed background with little object semantics.
Detecting traces of background reconstruction and recovering the removed object's original extent are distinct requirements for this task.
The detection--grounding gap motivates localization cues that describe the reconstructed background and its spatial boundaries, complementing the object semantics used to localize additions or replacements where target objects remain visible.

\subsubsection{By Generation Model}
\label{app:by_generation_model}

Table~\ref{tab:results_per_model} extends the main results with generation-model breakdowns and the overall scores on all 1,000 ManiVidBench pairs.
Indirect I2V groups follow the final video generator, pooling over first-frame editors, with task metrics and aggregation defined in Appendix~\ref{app:evaluation_metrics}.

\noindent\textbf{Generation Paradigms.}
Detection accuracy is lower for Direct V2V than Indirect I2V (0.889 vs. 0.954), accompanied by lower ROUGE-L (0.582 vs. 0.605) and CSS (0.774 vs. 0.820).
Grounding favors Direct V2V, with 0.415 mIoU and 0.430 J\&F vs. 0.398 and 0.413.
These complementary task-level differences motivate evaluating authenticity, explanation quality, and spatial recovery separately across generation paradigms.

\noindent\textbf{Individual Generation Models.}
Within Direct V2V, VideoCoF yields the lowest mIoU (0.335), whereas Kling-O1 is hardest to detect (0.688 accuracy) despite stronger grounding (0.435 mIoU).
Wan 2.2-Video-Edit reaches 0.553 mIoU and 0.979 accuracy, showing that the difficulty of locating and recognizing manipulations varies across video editing models.
Within Indirect I2V, LTX-2 leads grounding at 0.440 mIoU, compared with Wan 2.2's 0.383 and Kling-V3-Omni's 0.292.
Detection accuracy ranges from 0.781 to 0.984 across these three I2V models, while their mIoU spans 0.148.
The generation-model breakdown therefore shows variation in both authenticity discrimination and spatial recovery.

\begin{table}[!t]
\vspace{-4mm}
\centering
\caption{\textbf{ManiVidLens Performance by Generation Model.} ($\cdot$) indicate each generation paradigm's share of ManiVidBench.
Indirect I2V groups follow the final video generator, pooling over first-frame editors.
Avg. aggregates each paradigm, while Overall matches the main tables under the evaluation protocol in Appendix~\ref{app:evaluation_metrics}.}
\setlength{\tabcolsep}{2pt}
\renewcommand{\arraystretch}{1.5}
\belowrulesep=-0.25pt
\aboverulesep=-0.25pt
\resizebox{\textwidth}{!}{%
\begin{tabular}{lc|*{8}{c}|*{4}{c}|c}
\toprule[1.5pt]
\multirow{2}{*}{\textbf{Task}} & \multirow{2}{*}{\textbf{Metric}}
& \multicolumn{8}{c|}{\textbf{Direct V2V (52.3\%)}}
& \multicolumn{4}{c|}{\textbf{Indirect I2V (47.7\%)}}
& \multirow{2}{*}{\textbf{Overall}} \\
\cmidrule(l{3pt}r{3pt}){3-10} \cmidrule(l{3pt}r{3pt}){11-14}
& & \textbf{Kling-O1} & \makecell{\textbf{Kling-V3}\\\textbf{Omni}}
& \makecell{\textbf{Runway}\\\textbf{Aleph}} & \makecell{\textbf{Wan 2.2}\\\textbf{Video-Edit}}
& \textbf{VideoCoF} & \textbf{VACE} & \textbf{VideoPainter} & \textbf{Avg.}
& \makecell{\textbf{Kling-V3}\\\textbf{Omni}} & \textbf{Wan 2.2} & \textbf{LTX-2} & \textbf{Avg.} & \\
\midrule
\multicolumn{2}{l|}{$N$ (pairs)} & 69 & 19 & 81 & 94 & 209 & 44 & 7 & 523 & 57 & 204 & 216 & 477 & 1,000 \\
\midrule
\multirow{2}{*}{\textbf{Grounding}} & mIoU & 0.435 & 0.419 & 0.452 & \textbf{0.553} & 0.335 & 0.396 & \underline{0.465} & 0.415 & 0.292 & 0.383 & 0.440 & 0.398 & 0.407 \\
 & J\&F & 0.450 & 0.436 & 0.466 & \textbf{0.574} & 0.347 & 0.414 & \underline{0.469} & 0.430 & 0.303 & 0.402 & 0.452 & 0.413 & 0.422 \\
\midrule
\multirow{2}{*}{\textbf{Explanation}} & R-L & 0.533 & 0.557 & 0.605 & \textbf{0.621} & 0.570 & 0.596 & 0.607 & 0.582 & \underline{0.615} & 0.608 & 0.600 & 0.605 & 0.593 \\
 & CSS & 0.686 & 0.740 & 0.828 & \textbf{0.855} & 0.743 & 0.796 & 0.789 & 0.774 & \underline{0.835} & 0.826 & 0.811 & 0.820 & 0.796 \\
\midrule
\multirow{2}{*}{\textbf{Detection}} & Acc & 0.688 & 0.947 & 0.975 & 0.979 & 0.854 & 0.977 & \textbf{1.000} & 0.889 & 0.781 & 0.971 & \underline{0.984} & 0.954 & 0.920 \\
 & F1 & 0.751 & 0.944 & 0.976 & 0.978 & 0.850 & 0.978 & \textbf{1.000} & 0.892 & 0.725 & 0.970 & \underline{0.984} & 0.952 & 0.919 \\
\bottomrule[1.5pt]
\end{tabular}%
}
\vspace{-4mm}
\label{tab:results_per_model}
\end{table}

\subsubsection{From Classification to Attribution}
\label{app:attribution}
\begin{figure*}[htb]
    \centering
    \vspace{-4mm}
    \includegraphics[width=0.8\textwidth]{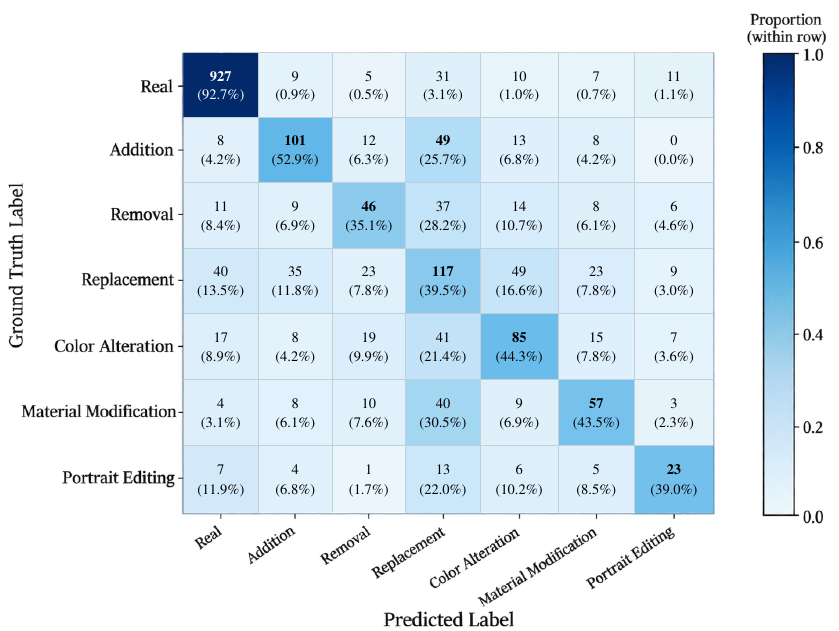}
    \vspace{-1mm}
    \caption{\textbf{Confusion Matrix for Seven-Class Manipulation Attribution.}}
    \vspace{-3mm}
    
    \label{fig:attribution}
\end{figure*}

Given the strong performance of ManiVidLens on binary forgery detection (0.920 Acc and 0.919 F1), we further extend the evaluation to the more challenging \textbf{manipulation attribution task}. Instead of simply distinguishing between real and fake videos, this task further requires identifying the specific manipulation type for fake videos. Our classification prompt already asks the model to predict the most likely manipulation type in addition to the authenticity label. We therefore directly evaluate these predictions under a 7-class setting, comprising Real and the six manipulation types in ManiVid. On ManiVidBench, ManiVidLens achieves an overall 7-class accuracy of 0.678 and a Macro-F1 of 0.501. The corresponding confusion matrix is shown in Fig.~\ref{fig:attribution}.
A closer inspection of the confusion matrix reveals a clear distinction between authenticity recognition and fine-grained manipulation attribution. ManiVidLens correctly classifies 927 out of 1,000 real videos, corresponding to a 92.7\% recall for the Real class. In contrast, the errors among manipulated videos are predominantly concentrated across different manipulation types rather than between real and fake, indicating that distinguishing the specific editing operation remains substantially more challenging than determining authenticity.
Among the manipulation categories, Addition achieves the highest class-wise recall (101/191, 52.9\%), followed by Color Alteration (85/192, 44.3\%) and Material Modification (57/131, 43.5\%). A notable pattern is that Replacement acts as the dominant source of cross-category confusion. For example, 49 Addition, 37 Removal, 41 Color Alteration, 40 Material Modification, and 13 Portrait Editing samples are predicted as Replacement. Conversely, true Replacement samples are also distributed across several related categories, most prominently Color Alteration (49 cases), Real (40 cases), and Addition (35 cases). This confusion pattern suggests that the model can often identify the presence of localized manipulation evidence while remaining uncertain about the exact operation that produced it. In particular, manipulation types that modify an existing object, such as Replacement, Color Alteration, and Material Modification, may share similar spatial and visual forensic cues, making fine-grained attribution inherently more ambiguous. These results further motivate manipulation attribution as a more stringent diagnostic setting beyond binary forgery detection.

\subsubsection{Human Evaluation for Anomaly Explanation}
\label{app:human_evaluation}

Although ROUGE-L and Cosine Similarity Score (CSS) provide scalable and reproducible evaluation for the anomaly explanation task, both metrics have inherent limitations. In particular, ROUGE-L relies heavily on lexical and structural overlap with the reference explanation and may therefore favor responses that follow similar wording or structured templates, even when semantically equivalent explanations are expressed differently. CSS alleviates this issue by measuring semantic similarity in the embedding space, but it still cannot directly determine whether the described forensic evidence is actually supported by the visual content. We therefore conduct an additional human evaluation to complement the automatic metrics.

Due to the substantial cost of manually inspecting video content and reviewing long-form forensic explanations, it is impractical to conduct human evaluation on all manipulated samples in ManiVidBench. We instead randomly sample 50 manipulated videos from each of the six manipulation categories, resulting in 300 videos in total. We evaluate ManiVidLens together with three representative comparison methods: the image-based expert LEGION, the video-based expert Omni-Fake-R1, and the general MLLM GPT-5.6 Sol. This yields 1,200 model-generated explanations for manual inspection. Four authors independently review the corresponding videos, manipulated regions, and generated explanations, while the model identities are hidden during evaluation to reduce potential bias. Each explanation is scored from 1 to 5 along the following three dimensions:

\begin{itemize}[leftmargin=10pt, topsep=0pt, itemsep=3pt, partopsep=1pt, parsep=1pt]
    \item \textbf{Evidence Faithfulness.} Whether the described visual or temporal anomalies are actually observable in the video and localized manipulation region, without hallucinating unsupported artifacts. A score of 1 indicates that the explanation is largely inconsistent with the observable evidence, while a score of 5 indicates that nearly all described evidence is visually well supported.
    
    \item \textbf{Forensic Relevance.} Whether the identified evidence is diagnostically relevant to video manipulation, rather than consisting of generic scene descriptions or weakly related observations. A score of 1 denotes largely irrelevant or non-forensic descriptions, whereas a score of 5 indicates that the explanation consistently identifies meaningful manipulation-sensitive cues, such as boundary, texture, lighting, or temporal inconsistencies.
    
    \item \textbf{Explanation Quality.} Whether the explanation provides a clear, specific, and coherent connection between the observed anomaly and the manipulation hypothesis. A score of 1 corresponds to vague, repetitive, or poorly justified reasoning, while a score of 5 represents precise and logically well-supported forensic reasoning with minimal redundancy.
\end{itemize}

The four authors achieve an intraclass correlation coefficient (ICC) of $0.86$, indicating substantial agreement between them.
For each explanation, we first average the scores from the four evaluators over the three evaluation dimensions to obtain a single human evaluation score. Specifically, for video $i$ and model $m$, the human score is computed as
\begin{equation}
H_{i,m}
=
\frac{1}{4\times3}
\sum_{e=1}^{4}
\sum_{d=1}^{3}
s_{i,m}^{(e,d)},
\end{equation}
where $s_{i,m}^{(e,d)}$ denotes the score assigned by evaluator $e$ on evaluation dimension $d$. This produces one human score for each model-generated explanation.

Since our primary interest is whether an automatic metric can correctly reflect the \emph{relative quality of different models on the same video}, rather than merely correlating with their aggregated scores, we perform correlation analysis at the individual-video level. For each video $i$, we rank the four evaluated models according to their human evaluation scores, yielding a human ranking vector $\mathbf{r}^{H}_{i}$. We independently rank the same four models according to their CSS or ROUGE-L scores, denoted as $\mathbf{r}^{Q}_{i}$, where $Q\in\{\mathrm{CSS},\mathrm{R\text{-}L}\}$. We then compute Spearman's rank correlation coefficient between the two rankings:
\begin{equation}
\rho_{i}^{Q}
=
\operatorname{Spearman}
\left(
\mathbf{r}^{H}_{i},
\mathbf{r}^{Q}_{i}
\right).
\end{equation}
When no ties occur, this is equivalent to
\begin{equation}
\rho_{i}^{Q}
=
1-
\frac{6\sum_{m=1}^{4}
\left(r^{H}_{i,m}-r^{Q}_{i,m}\right)^2}
{4(4^2-1)}.
\end{equation}
For tied scores, average ranks are assigned before computing the rank correlation.

For each manipulation category $c$, we average the per-video correlations over its 50 sampled videos:
\begin{equation}
\bar{\rho}_{c}^{Q}
=
\frac{1}{50}
\sum_{i\in c}
\rho_{i}^{Q}.
\end{equation}
Finally, we report the mean correlation across all six manipulation categories as
\begin{equation}
\bar{\rho}^{Q}
=
\frac{1}{6}
\sum_{c=1}^{6}
\bar{\rho}_{c}^{Q}.
\end{equation}

CSS exhibits consistently strong agreement with human judgments across all six manipulation categories. The category-wise mean Spearman correlations for Addition, Removal, Replacement, Color Alteration, Material Modification, and Portrait Editing are $0.804$, $0.768$, $0.812$, $0.784$, $0.796$, and $0.756$, respectively, resulting in an overall average correlation of $\bar{\rho}=0.787$. In contrast, the corresponding correlations for ROUGE-L are $0.472$, $0.436$, $0.488$, $0.412$, $0.448$, and $0.396$, with a substantially lower average of $\bar{\rho}=0.442$. These results indicate that CSS more consistently preserves the relative quality ordering perceived by human evaluators at the individual-video level, whereas ROUGE-L shows only moderate agreement and is more sensitive to lexical overlap and template-level similarity with the reference explanations. Therefore, although neither automatic metric can replace direct human inspection, the substantially stronger agreement between CSS and human judgments supports the use of CSS in Table~\ref{tab:explanation} as a meaningful indicator of the relative anomaly explanation capability of different models.

Table~\ref{tab:human_scores_absolute} reports the category-wise absolute human scores, averaged over four evaluators and three evaluation dimensions. ManiVidLens obtains the highest mean score in five of the six categories; LEGION leads on Portrait Editing.

\begin{table*}[t]
    \centering
    \small
    \setlength{\tabcolsep}{7pt}
    \caption{\textbf{Absolute Human Evaluation Scores (1--5) for Anomaly Explanations by Manipulation Task.}}
    \label{tab:human_scores_absolute}
    \begin{tabular}{lcccccc}
        \toprule
        Model & Addition & Removal & Replacement & Color & Material & Portrait \\
        \midrule
        LEGION & 3.77 & \underline{3.72} & \underline{3.91} & \underline{3.84} & \underline{3.88} & \textbf{3.87} \\
        Omni-Fake-R1 & \underline{3.81} & 3.68 & 3.75 & 3.76 & 3.77 & 3.63 \\
        GPT-5.6 Sol & 2.74 & 2.80 & 2.78 & 2.76 & 2.82 & 2.78 \\
        ManiVidLens & \textbf{4.25} & \textbf{4.09} & \textbf{4.11} & \textbf{4.09} & \textbf{4.18} & \underline{3.80} \\
        \bottomrule
    \end{tabular}
\end{table*}

\FloatBarrier
\subsection{Ablation Study}
\label{app:ablation}

The ablations examine how frame count, architectural components, and task-specific prompts affect performance.
Tables~\ref{tab:ablation_grounding}, \ref{tab:ablation_explanation}, and~\ref{tab:ablation_detection} expand the aggregate results in Tables~\ref{tab:ablation_frames} and~\ref{tab:ablation_module}.
Each table compares 3, 5, 7, and 9 sampled frames, followed by Base, + PDM, + FER, and + Both.
Base omits PDM and FER, and + Both denotes the full ManiVidLens architecture.
The default model uses five frames.
Avg. follows the benchmark-level aggregation in Appendix~\ref{app:evaluation_metrics}.

\FloatBarrier
\subsubsection{Artifact Grounding}
\begin{table}[!htbp]
    \centering
    \vspace{-4mm}
    \caption{\textbf{Category-wise Ablation for Artifact Grounding.}
    Column groups follow the six manipulation categories, with benchmark-level scores in Avg.
    Metrics are mIoU and J\&F ($\uparrow$).}
    \definecolor{mygray}{gray}{.92}
    \setlength{\tabcolsep}{3pt}
    \renewcommand{\arraystretch}{1.3}
    \belowrulesep=-0.25pt
    \aboverulesep=-0.25pt
    \resizebox{\linewidth}{!}{
        \begin{tabular}{l|*{12}{>{\centering\arraybackslash}p{0.85cm}}|*{2}{>{\centering\arraybackslash}p{0.85cm}}}
        \toprule[1.5pt]
        \multirow{3}{*}{\textbf{Setting}} & \multicolumn{12}{c|}{\textbf{ManiVidBench}} & \multicolumn{2}{c}{\multirow{2}{*}{\textbf{Avg.}}} \\
        & \multicolumn{2}{c}{\textbf{Addition}} & \multicolumn{2}{c}{\textbf{Removal}} & \multicolumn{2}{c}{\textbf{Replacement}} & \multicolumn{2}{c}{\textbf{Color}} & \multicolumn{2}{c}{\textbf{Material}} & \multicolumn{2}{c|}{\textbf{Portrait}} & \multicolumn{2}{c}{} \\
        \cmidrule(l{3pt}r{3pt}){2-3} \cmidrule(l{3pt}r{3pt}){4-5} \cmidrule(l{3pt}r{3pt}){6-7} \cmidrule(l{3pt}r{3pt}){8-9} \cmidrule(l{3pt}r{3pt}){10-11} \cmidrule(l{3pt}r{3pt}){12-13} \cmidrule(l{3pt}r{3pt}){14-15}
         & mIoU & J\&F & mIoU & J\&F & mIoU & J\&F & mIoU & J\&F & mIoU & J\&F & mIoU & J\&F & mIoU & J\&F \\
        \hline
        \rowcolor{mygray}
        \multicolumn{15}{c}{\textit{Ablation 1: Number of Sampled Frames}} \\
        3 Frames & 0.412 & 0.432 & 0.119 & 0.118 & 0.367 & 0.382 & 0.396 & 0.417 & 0.451 & 0.456 & 0.279 & 0.286 & 0.354 & 0.368 \\
        5 Frames & 0.444 & 0.467 & \underline{0.135} & \underline{0.137} & \underline{0.421} & \underline{0.439} & \textbf{0.487} & \textbf{0.507} & \textbf{0.526} & \textbf{0.532} & 0.290 & \underline{0.306} & 0.407 & \underline{0.422} \\
        7 Frames & \textbf{0.503} & \textbf{0.533} & 0.109 & 0.120 & \textbf{0.435} & \textbf{0.460} & 0.447 & 0.474 & 0.478 & 0.494 & \textbf{0.348} & \textbf{0.370} & \underline{0.408} & \textbf{0.431} \\
        9 Frames & \underline{0.470} & \underline{0.475} & \textbf{0.161} & \textbf{0.155} & 0.418 & 0.425 & \underline{0.485} & \underline{0.496} & \underline{0.514} & \underline{0.509} & \underline{0.310} & 0.301 & \textbf{0.413} & 0.416 \\
        \hline
        \rowcolor{mygray}
        \multicolumn{15}{c}{\textit{Ablation 2: Architectural Components}} \\
        Base & 0.414 & 0.432 & 0.132 & 0.135 & 0.343 & 0.356 & 0.390 & 0.406 & 0.399 & 0.406 & 0.182 & 0.186 & 0.336 & 0.348 \\
        + PDM & \textbf{0.450} & \underline{0.466} & \textbf{0.158} & \textbf{0.155} & 0.393 & 0.406 & \underline{0.427} & \underline{0.446} & 0.401 & 0.416 & 0.170 & 0.178 & 0.368 & 0.380 \\
        + FER & 0.433 & 0.447 & 0.112 & 0.116 & \textbf{0.434} & \textbf{0.449} & 0.422 & 0.441 & \underline{0.449} & \underline{0.449} & \underline{0.202} & \underline{0.208} & \underline{0.378} & \underline{0.389} \\
        + Both~(Ours) & \underline{0.444} & \textbf{0.467} & \underline{0.135} & \underline{0.137} & \underline{0.421} & \underline{0.439} & \textbf{0.487} & \textbf{0.507} & \textbf{0.526} & \textbf{0.532} & \textbf{0.290} & \textbf{0.306} & \textbf{0.407} & \textbf{0.422} \\
        \bottomrule[1.5pt]
        \end{tabular}
    }
    \vspace{-3mm}
    \label{tab:ablation_grounding}
\end{table}

\noindent\textbf{Frame Count.}
Increasing from three to five frames improves mIoU in all six categories (Table~\ref{tab:ablation_grounding}), with the largest gains in Color (+0.092), Material (+0.076), and Replacement (+0.055).
Color and Material attain their highest grounding scores at five frames.
Seven frames improve Addition mIoU from 0.444 to 0.503 and Portrait from 0.290 to 0.348, while Color and Material decrease by 0.040 and 0.048.
These opposing category-level changes produce the small aggregate increase from 0.407 to 0.408 mIoU.
Nine frames give the highest Removal mIoU of 0.161 and the highest overall mIoU of 0.413. Overall J\&F decreases from 0.431 at seven frames to 0.416 at nine frames, while inference time rises to 19.20\,s (Table~\ref{tab:ablation_frames}).

\noindent\textbf{Architectural Components.}
The full model improves mIoU over Base in every category.
The largest mIoU gains occur in Material (\textbf{from 0.399 to 0.526}), Portrait (from 0.182 to 0.290), and Color (from 0.390 to 0.487).
Adding PDM to FER improves these categories by a further 0.077, 0.088, and 0.065, respectively.
The best individual configurations differ across categories: PDM alone achieves the highest Addition and Removal mIoU of 0.450 and 0.158, while FER alone leads Replacement at 0.434.
The full model's Removal gain over Base is 0.003.
The joint architecture is strongest on attribute and portrait edits, while Removal remains its lowest-scoring category.

We also measure conditioning-frame coverage directly. For each manipulated video in ManiVid-38K, we check whether at least one of the five inference frames has a nonempty ground-truth manipulation mask. The five sampled frames cover all manipulated videos (100\% video-level coverage). The first manipulated frame appears, on average, 0.72\% of the way through a video; the latest onset is 12.8\%, and 98.53\% of videos show the manipulation within the first 10\%. This audit describes coverage on ManiVid-38K and does not establish coverage for brief or late edits in other distributions.

As a reference for the geometric prompts, we additionally run SAM2 Hiera-L with bounding boxes derived from the ground-truth masks on the 1,000 manipulated ManiVidBench videos. This oracle-box setting reaches 84.55\% mIoU. It serves as a prompt-conditioned reference and is not directly comparable to the end-to-end ManiVidLens result, which predicts its prompts.

\FloatBarrier
\subsubsection{Anomaly Explanation}
\begin{table}[!htbp]
    \centering
    \vspace{-2mm}
    \caption{\textbf{Category-wise Ablation for Anomaly Explanation.}
    Column groups follow the six manipulation categories in the main results, with benchmark-level scores in Avg.
    Metrics are \mbox{ROUGE-L} (R-L) and Cosine Similarity Score (CSS) ($\uparrow$).}
    \definecolor{mygray}{gray}{.92}
    \setlength{\tabcolsep}{3pt}
    \renewcommand{\arraystretch}{1.3}
    \belowrulesep=-0.25pt
    \aboverulesep=-0.25pt
    \resizebox{\linewidth}{!}{
        \begin{tabular}{l|*{12}{>{\centering\arraybackslash}p{0.85cm}}|*{2}{>{\centering\arraybackslash}p{0.85cm}}}
        \toprule[1.5pt]
        \multirow{3}{*}{\textbf{Setting}} & \multicolumn{12}{c|}{\textbf{ManiVidBench}} & \multicolumn{2}{c}{\multirow{2}{*}{\textbf{Avg.}}} \\
        & \multicolumn{2}{c}{\textbf{Addition}} & \multicolumn{2}{c}{\textbf{Removal}} & \multicolumn{2}{c}{\textbf{Replacement}} & \multicolumn{2}{c}{\textbf{Color}} & \multicolumn{2}{c}{\textbf{Material}} & \multicolumn{2}{c|}{\textbf{Portrait}} & \multicolumn{2}{c}{} \\
        \cmidrule(l{3pt}r{3pt}){2-3} \cmidrule(l{3pt}r{3pt}){4-5} \cmidrule(l{3pt}r{3pt}){6-7} \cmidrule(l{3pt}r{3pt}){8-9} \cmidrule(l{3pt}r{3pt}){10-11} \cmidrule(l{3pt}r{3pt}){12-13} \cmidrule(l{3pt}r{3pt}){14-15}
         & R-L & CSS & R-L & CSS & R-L & CSS & R-L & CSS & R-L & CSS & R-L & CSS & R-L & CSS \\
        \hline
        \rowcolor{mygray}
        \multicolumn{15}{c}{\textit{Ablation 1: Number of Sampled Frames}} \\
        3 Frames & 0.587 & 0.791 & 0.574 & 0.761 & 0.571 & 0.756 & 0.575 & 0.756 & 0.578 & 0.781 & 0.544 & 0.695 & 0.575 & 0.763 \\
        5 Frames & 0.603 & 0.813 & \underline{0.576} & 0.773 & \textbf{0.613} & \textbf{0.831} & 0.582 & 0.772 & 0.587 & 0.795 & \underline{0.545} & \underline{0.699} & \underline{0.593} & 0.796 \\
        7 Frames & \underline{0.614} & \underline{0.832} & \textbf{0.592} & \textbf{0.797} & \underline{0.594} & \underline{0.798} & \textbf{0.593} & \textbf{0.802} & \textbf{0.604} & \textbf{0.818} & \textbf{0.556} & \textbf{0.722} & \textbf{0.596} & \textbf{0.803} \\
        9 Frames & \textbf{0.618} & \textbf{0.836} & \textbf{0.592} & \underline{0.796} & 0.587 & 0.790 & \underline{0.589} & \underline{0.794} & \underline{0.600} & \underline{0.810} & 0.519 & 0.685 & 0.592 & \underline{0.797} \\
        \hline
        \rowcolor{mygray}
        \multicolumn{15}{c}{\textit{Ablation 2: Architectural Components}} \\
        Base & 0.580 & 0.785 & 0.553 & 0.743 & 0.528 & 0.755 & 0.562 & 0.761 & 0.563 & 0.772 & 0.521 & 0.676 & 0.552 & 0.758 \\
        + PDM & 0.581 & 0.800 & 0.560 & 0.764 & 0.564 & 0.768 & 0.568 & \underline{0.773} & 0.567 & 0.783 & 0.518 & 0.672 & 0.565 & 0.771 \\
        + FER & \textbf{0.621} & \textbf{0.846} & \textbf{0.592} & \textbf{0.796} & \underline{0.598} & \underline{0.807} & \textbf{0.597} & \textbf{0.805} & \textbf{0.605} & \textbf{0.824} & \underline{0.532} & \underline{0.694} & \textbf{0.598} & \textbf{0.808} \\
        + Both~(Ours) & \underline{0.603} & \underline{0.813} & \underline{0.576} & \underline{0.773} & \textbf{0.613} & \textbf{0.831} & \underline{0.582} & 0.772 & \underline{0.587} & \underline{0.795} & \textbf{0.545} & \textbf{0.699} & \underline{0.593} & \underline{0.796} \\
        \bottomrule[1.5pt]
        \end{tabular}
    }
    \vspace{-3mm}
    \label{tab:ablation_explanation}
\end{table}

\noindent\textbf{Frame Count.}
Seven frames improve both explanation metrics over five frames in Addition, Removal, Color, Material, and Portrait, while Replacement decreases from 0.613 to 0.594 ROUGE-L and from 0.831 to 0.798 CSS (Table~\ref{tab:ablation_explanation}).
Overall ROUGE-L rises from 0.593 to 0.596, and CSS from 0.796 to 0.803.
Color shows the largest positive CSS increase, from 0.772 to 0.802.
At nine frames, Addition reaches its highest ROUGE-L of 0.618 and CSS of 0.836, while Removal reaches 0.592 ROUGE-L and 0.796 CSS.
Moving from seven to nine frames reduces Portrait ROUGE-L from 0.556 to 0.519 and CSS from 0.722 to 0.685, alongside smaller decreases in Replacement, Color, and Material.
The aggregate explanation scores peak at seven frames.

\noindent\textbf{Architectural Components.}
FER alone achieves the highest ROUGE-L and CSS in Addition, Removal, Color, and Material, while the full model leads Replacement with 0.613 ROUGE-L and 0.831 CSS.
Base trails FER on Portrait with 0.521 ROUGE-L and 0.676 CSS, compared with FER's 0.532 ROUGE-L and 0.694 CSS.
Adding PDM to FER reduces CSS by 0.033 in both Addition and Color, while raising Replacement ROUGE-L by 0.015 and CSS by 0.024, and raising Portrait ROUGE-L by 0.013 and CSS by 0.005.
FER alone yields the strongest overall explanation scores of 0.598 ROUGE-L and 0.808 CSS; the full architecture yields the strongest grounding scores while retaining 0.593 ROUGE-L and 0.796 CSS at the benchmark level.

\FloatBarrier
\subsubsection{Forgery Detection}
\begin{table}[!htbp]
    \centering
    \vspace{-2mm}
    \caption{\textbf{Category-wise Ablation for Forgery Detection.}
    Column groups follow the six manipulation categories in the main results, with benchmark-level scores in Avg.
    Metrics are video-level Accuracy (Acc) and F1 ($\uparrow$) for original and manipulated videos.}
    \definecolor{mygray}{gray}{.92}
    \setlength{\tabcolsep}{3pt}
    \renewcommand{\arraystretch}{1.3}
    \belowrulesep=-0.25pt
    \aboverulesep=-0.25pt
    \resizebox{\linewidth}{!}{
        \begin{tabular}{l|*{12}{>{\centering\arraybackslash}p{0.85cm}}|*{2}{>{\centering\arraybackslash}p{0.85cm}}}
        \toprule[1.5pt]
        \multirow{3}{*}{\textbf{Setting}} & \multicolumn{12}{c|}{\textbf{ManiVidBench}} & \multicolumn{2}{c}{\multirow{2}{*}{\textbf{Avg.}}} \\
        & \multicolumn{2}{c}{\textbf{Addition}} & \multicolumn{2}{c}{\textbf{Removal}} & \multicolumn{2}{c}{\textbf{Replacement}} & \multicolumn{2}{c}{\textbf{Color}} & \multicolumn{2}{c}{\textbf{Material}} & \multicolumn{2}{c|}{\textbf{Portrait}} & \multicolumn{2}{c}{} \\
        \cmidrule(l{3pt}r{3pt}){2-3} \cmidrule(l{3pt}r{3pt}){4-5} \cmidrule(l{3pt}r{3pt}){6-7} \cmidrule(l{3pt}r{3pt}){8-9} \cmidrule(l{3pt}r{3pt}){10-11} \cmidrule(l{3pt}r{3pt}){12-13} \cmidrule(l{3pt}r{3pt}){14-15}
         & Acc & F1 & Acc & F1 & Acc & F1 & Acc & F1 & Acc & F1 & Acc & F1 & Acc & F1 \\
        \hline
        \rowcolor{mygray}
        \multicolumn{15}{c}{\textit{Ablation 1: Number of Sampled Frames}} \\
        3 Frames & 0.953 & 0.953 & 0.920 & 0.920 & 0.887 & 0.885 & 0.906 & 0.906 & 0.912 & 0.912 & 0.805 & 0.803 & 0.906 & 0.905 \\
        5 Frames & \underline{0.969} & \underline{0.968} & 0.947 & 0.945 & 0.885 & 0.883 & \underline{0.919} & \underline{0.919} & \textbf{0.939} & \textbf{0.941} & \textbf{0.839} & \textbf{0.846} & 0.920 & \underline{0.919} \\
        7 Frames & \textbf{0.971} & \textbf{0.971} & \textbf{0.958} & \textbf{0.958} & \underline{0.899} & \underline{0.899} & \textbf{0.930} & \textbf{0.930} & \underline{0.931} & 0.931 & \underline{0.822} & \underline{0.821} & \underline{0.926} & \textbf{0.926} \\
        9 Frames & 0.966 & 0.966 & \underline{0.950} & \underline{0.949} & \textbf{0.907} & \textbf{0.905} & \textbf{0.930} & \textbf{0.930} & \textbf{0.939} & \underline{0.938} & 0.814 & 0.810 & \textbf{0.927} & \textbf{0.926} \\
        \hline
        \rowcolor{mygray}
        \multicolumn{15}{c}{\textit{Ablation 2: Architectural Components}} \\
        Base & 0.935 & 0.939 & 0.912 & 0.919 & 0.856 & 0.871 & 0.878 & 0.891 & 0.885 & 0.897 & 0.780 & 0.787 & 0.882 & 0.893 \\
        + PDM & 0.958 & 0.960 & 0.935 & 0.936 & 0.858 & 0.861 & \underline{0.909} & \underline{0.909} & 0.931 & 0.932 & \underline{0.797} & \underline{0.803} & 0.903 & 0.905 \\
        + FER & \textbf{0.971} & \textbf{0.971} & \underline{0.943} & \underline{0.941} & \textbf{0.899} & \textbf{0.896} & \underline{0.909} & 0.905 & \underline{0.935} & \underline{0.934} & 0.780 & 0.776 & \underline{0.918} & \underline{0.916} \\
        + Both~(Ours) & \underline{0.969} & \underline{0.968} & \textbf{0.947} & \textbf{0.945} & \underline{0.885} & \underline{0.883} & \textbf{0.919} & \textbf{0.919} & \textbf{0.939} & \textbf{0.941} & \textbf{0.839} & \textbf{0.846} & \textbf{0.920} & \textbf{0.919} \\
        \bottomrule[1.5pt]
        \end{tabular}
    }
    \vspace{-3mm}
    \label{tab:ablation_detection}
\end{table}

\noindent\textbf{Frame Count.}
Additional frames improve overall detection while changing which categories perform best (Table~\ref{tab:ablation_detection}).
Moving from five to seven frames increases Accuracy in Addition, Removal, Replacement, and Color, including gains of 0.014 in Replacement and 0.011 in Removal.
Material and Portrait decrease by 0.008 and 0.017.
Portrait reaches its highest Accuracy of 0.839 at five frames, and Material peaks at 0.939 with either five or nine.
Seven and nine frames both achieve 0.926 overall F1.
The nine-frame Accuracy of 0.927 exceeds seven frames by 0.001, with inference time increasing from 17.72\,s to 19.20\,s.

\noindent\textbf{Architectural Components.}
The full model's detection scores reflect Accuracy gains in all six categories.
Relative to Base, Accuracy increases across all six categories; the displayed gains are 0.034 in Addition, 0.035 in Removal, 0.029 in Replacement, 0.041 in Color, 0.054 in Material, and 0.059 in Portrait.
Removal improves from 0.912 to 0.947, while Replacement improves from 0.856 to 0.885.
The full model reaches 0.920 Accuracy and 0.919 F1, gains of 0.038 and 0.026 over Base.
Combining the modules also improves on either module alone, whose overall Accuracy values are 0.903 for PDM and 0.918 for FER.
The largest Accuracy recovery occurs in Portrait, where the full model reaches 0.839 vs. 0.797 for PDM and 0.780 for FER.

\subsubsection{Prompt Design}
\label{app:ablation_prompt}

\noindent\textbf{Prompt Settings.}
The separate setting performs explanation, detection, and grounding through three task-specific queries to one shared ManiVidLens model.
The all-in-one setting combines the same tasks into a single response in the fixed order of explanation, detection, and grounding (Table~\ref{app:tab:ours_prompts}).
During Stage~1, separate prompting expands each sampled video view into three task-specific instances over one epoch, while all-in-one prompting uses one joint instance over two epochs; both settings then use one Stage~2 epoch.

\begin{table}[H]
    \centering
    \vspace{-2mm}
    \caption{\textbf{Category-wise Ablation on Prompt Design.}
    All-in-one and separate prompts are compared across six manipulation categories and three tasks.
    Avg.\ reports benchmark-level scores.
    Higher metric values ($\uparrow$) indicate better performance on all three tasks.}
    \definecolor{mygray}{gray}{.92}
    \setlength{\tabcolsep}{3pt}
    \renewcommand{\arraystretch}{1.3}
    \belowrulesep=-0.25pt
    \aboverulesep=-0.25pt
    \resizebox{\linewidth}{!}{
        \begin{tabular}{l|*{12}{>{\centering\arraybackslash}p{0.85cm}}|*{2}{>{\centering\arraybackslash}p{0.85cm}}}
        \toprule[1.5pt]
        \multirow{2}{*}{\textbf{Prompt Design}} & \multicolumn{12}{c|}{\textbf{ManiVidBench}} & \multicolumn{2}{c}{\multirow{2}{*}{\textbf{Avg.}}} \\
        & \multicolumn{2}{c}{\textbf{Addition}} & \multicolumn{2}{c}{\textbf{Removal}} & \multicolumn{2}{c}{\textbf{Replacement}} & \multicolumn{2}{c}{\textbf{Color}} & \multicolumn{2}{c}{\textbf{Material}} & \multicolumn{2}{c|}{\textbf{Portrait}} & \multicolumn{2}{c}{} \\
        \midrule
        \rowcolor{mygray}
        \multicolumn{15}{c}{\textit{Task Mode: Classification}} \\
        \textbf{Metrics} & Acc & F1 & Acc & F1 & Acc & F1 & Acc & F1 & Acc & F1 & Acc & F1 & Acc & F1 \\
        \cmidrule(l{3pt}r{3pt}){1-1} \cmidrule(l{3pt}r{3pt}){2-3} \cmidrule(l{3pt}r{3pt}){4-5} \cmidrule(l{3pt}r{3pt}){6-7} \cmidrule(l{3pt}r{3pt}){8-9} \cmidrule(l{3pt}r{3pt}){10-11} \cmidrule(l{3pt}r{3pt}){12-13} \cmidrule(l{3pt}r{3pt}){14-15}
        All-in-one & \underline{0.838} & \underline{0.833} & \underline{0.779} & \underline{0.768} & \underline{0.777} & \underline{0.766} & \underline{0.758} & \underline{0.744} & \underline{0.790} & \underline{0.783} & \underline{0.627} & \underline{0.569} & \underline{0.778} & \underline{0.767} \\
        Separate~(Ours) & \textbf{0.969} & \textbf{0.968} & \textbf{0.947} & \textbf{0.945} & \textbf{0.885} & \textbf{0.883} & \textbf{0.919} & \textbf{0.919} & \textbf{0.939} & \textbf{0.941} & \textbf{0.839} & \textbf{0.846} & \textbf{0.920} & \textbf{0.919} \\
        \hline
        \rowcolor{mygray}
        \multicolumn{15}{c}{\textit{Task Mode: Explanation}} \\
        \textbf{Metrics} & R-L & CSS & R-L & CSS & R-L & CSS & R-L & CSS & R-L & CSS & R-L & CSS & R-L & CSS \\
        \cmidrule(l{3pt}r{3pt}){1-1} \cmidrule(l{3pt}r{3pt}){2-3} \cmidrule(l{3pt}r{3pt}){4-5} \cmidrule(l{3pt}r{3pt}){6-7} \cmidrule(l{3pt}r{3pt}){8-9} \cmidrule(l{3pt}r{3pt}){10-11} \cmidrule(l{3pt}r{3pt}){12-13} \cmidrule(l{3pt}r{3pt}){14-15}
        All-in-one & \underline{0.598} & \underline{0.800} & \textbf{0.576} & \underline{0.766} & \underline{0.580} & \underline{0.771} & \underline{0.574} & \underline{0.757} & \underline{0.579} & \underline{0.774} & \underline{0.541} & \underline{0.690} & \underline{0.579} & \underline{0.769} \\
        Separate~(Ours) & \textbf{0.603} & \textbf{0.813} & \textbf{0.576} & \textbf{0.773} & \textbf{0.613} & \textbf{0.831} & \textbf{0.582} & \textbf{0.772} & \textbf{0.587} & \textbf{0.795} & \textbf{0.545} & \textbf{0.699} & \textbf{0.593} & \textbf{0.796} \\
        \hline
        \rowcolor{mygray}
        \multicolumn{15}{c}{\textit{Task Mode: Segmentation}} \\
        \textbf{Metrics} & mIoU & J\&F & mIoU & J\&F & mIoU & J\&F & mIoU & J\&F & mIoU & J\&F & mIoU & J\&F & mIoU & J\&F \\
        \cmidrule(l{3pt}r{3pt}){1-1} \cmidrule(l{3pt}r{3pt}){2-3} \cmidrule(l{3pt}r{3pt}){4-5} \cmidrule(l{3pt}r{3pt}){6-7} \cmidrule(l{3pt}r{3pt}){8-9} \cmidrule(l{3pt}r{3pt}){10-11} \cmidrule(l{3pt}r{3pt}){12-13} \cmidrule(l{3pt}r{3pt}){14-15}
        All-in-one & \underline{0.381} & \underline{0.395} & \underline{0.113} & \underline{0.119} & \underline{0.295} & \underline{0.308} & \underline{0.275} & \underline{0.288} & \underline{0.348} & \underline{0.359} & \underline{0.028} & \underline{0.032} & \underline{0.275} & \underline{0.286} \\
        Separate~(Ours) & \textbf{0.444} & \textbf{0.467} & \textbf{0.135} & \textbf{0.137} & \textbf{0.421} & \textbf{0.439} & \textbf{0.487} & \textbf{0.507} & \textbf{0.526} & \textbf{0.532} & \textbf{0.290} & \textbf{0.306} & \textbf{0.407} & \textbf{0.422} \\
        \bottomrule[1.5pt]
        \end{tabular}
    }
    \vspace{-3mm}
    \label{tab:ablation_prompt}
\end{table}

\noindent\textbf{Task Performance.}
Separate prompting improves all six metrics (Table~\ref{tab:ablation_prompt}).
Grounding mIoU increases \textbf{from 0.275 to 0.407}, and J\&F from 0.286 to 0.422.
Detection Accuracy rises \textbf{from 0.778 to 0.920}, and F1 from 0.767 to 0.919.
Explanation improves by 0.014 ROUGE-L and 0.027 CSS, reaching 0.593 ROUGE-L and 0.796 CSS.
The larger gains in detection and grounding show that task separation chiefly benefits authenticity decisions and spatial localization.

\noindent\textbf{Category-wise Effects.}
Separate prompting matches or improves every metric across all six manipulation categories.
The largest grounding and detection gains occur on Portrait: mIoU increases \textbf{from 0.028 to 0.290}, and detection F1 rises from 0.569 to 0.846.
Color mIoU also rises from 0.275 to 0.487, while Material CSS improves from 0.774 to 0.795.

\noindent\textbf{Task-Specific Generation.}
In the all-in-one sequence, detection is conditioned on the generated explanation, and grounding is conditioned on both preceding outputs.
Separate prompts give each task its own generation context: detection directly predicts authenticity, and grounding directly produces the localization phrase and segmentation tokens.
The three tasks retain shared model parameters and visual evidence.
The largest gains occur in the two tasks that follow explanation in the all-in-one sequence, supporting separate prompts as the default task protocol.

\FloatBarrier
\subsubsection{\texorpdfstring{Backbone and FER Pathway Ablations}{Backbone and FER Pathway Ablations}}
\label{app:backbone_fer_ablations}
We evaluate the sensitivity of ManiVidLens to the MLLM and segmentation-backbone scales, then isolate the two evidence-injection pathways in FER. Unless otherwise stated, all configurations use the same training split, budget, and optimization settings.

\noindent\textbf{Backbone Scale.} We compare Qwen3-VL-2B and Qwen3-VL-4B with SAM2 Hiera-Small and Hiera-Large while retaining both FER and PDM.

\begin{table*}[t]
    \centering
    \small
    \setlength{\tabcolsep}{7pt}
    \caption{\textbf{Backbone-scale Ablation on ManiVidBench.}}
    \label{tab:backbone_scale_ablation}
    \begin{tabular}{cc|cc|cc|cc}
        \toprule
        Qwen3-VL & SAM2 & mIoU & J\&F & ROUGE-L & CSS & Accuracy & F1 \\
        \midrule
        4B & Large & \textbf{0.407} & \textbf{0.422} & \textbf{0.593} & \textbf{0.796} & \textbf{0.920} & \textbf{0.919} \\
        2B & Large & \underline{0.404} & \underline{0.421} & 0.576 & 0.764 & 0.905 & 0.902 \\
        4B & Small & 0.401 & 0.415 & \underline{0.580} & \underline{0.775} & \underline{0.911} & \underline{0.909} \\
        2B & Small & 0.403 & 0.418 & 0.574 & 0.761 & 0.908 & 0.905 \\
        \bottomrule
    \end{tabular}
\end{table*}

The 4B + Large configuration achieves the highest scores across all six metrics. SAM2-Large improves both grounding metrics over SAM2-Small with either Qwen3-VL backbone. The 4B backbone improves explanation and detection with either SAM2 variant; the differences in grounding across Qwen3-VL scales are smaller.

\noindent\textbf{FER Pathways.} FE2SAM injects forensic features into SAM2 spatial features, whereas FE2VLM supplies forensic evidence to the MLLM as visual tokens. We disable PDM to isolate these pathways and compare each pathway separately and jointly with the Base model.

\begin{table*}[t]
    \centering
    \small
    \setlength{\tabcolsep}{7pt}
    \caption{\textbf{Ablation of the Two FER Evidence-injection Pathways.}}
    \label{tab:fer_pathway_ablation}
    \begin{tabular}{l|cc|cc|cc}
        \toprule
        Active pathways & mIoU & J\&F & ROUGE-L & CSS & Accuracy & F1 \\
        \midrule
        Neither (Base) & 0.336 & 0.348 & 0.552 & 0.758 & 0.882 & 0.893 \\
        FE2SAM only & \underline{0.372} & \underline{0.383} & 0.592 & 0.798 & 0.903 & 0.902 \\
        FE2VLM only & 0.367 & 0.379 & \underline{0.597} & \underline{0.807} & \underline{0.906} & \underline{0.905} \\
        Both (full FER) & \textbf{0.378} & \textbf{0.389} & \textbf{0.598} & \textbf{0.808} & \textbf{0.918} & \textbf{0.916} \\
        \bottomrule
    \end{tabular}
\end{table*}

FE2SAM alone improves grounding over Base from 0.336/0.348 to 0.372/0.383 mIoU/J\&F, while FE2VLM alone improves explanation from 0.552/0.758 to 0.597/0.807 ROUGE-L/CSS. Combining both pathways yields the strongest grounding and explanation results among these configurations (0.378/0.389 mIoU/J\&F and 0.598/0.808 ROUGE-L/CSS).

\begin{table*}[t]
    \centering
    \small
    \setlength{\tabcolsep}{6pt}
    \caption{\textbf{VideoSham Cross-dataset Detection Results Without Fine-tuning.} All values are percentages.}
    \label{tab:cross_dataset_videosham}
    \begin{tabular}{lcc}
        \toprule
        Metric & AIDE & ManiVidLens \\
        \midrule
        Real recall & \textbf{97.16} & \underline{66.48} \\
        Add an entity or subject & \underline{2.90} & \textbf{62.32} \\
        Remove an entity or subject & \underline{5.06} & \textbf{45.57} \\
        Modify background or color & \underline{11.11} & \textbf{62.96} \\
        Replace or add text & \underline{0.00} & \textbf{42.22} \\
        Repeat, drop, or reverse frames / change speed & \underline{1.69} & \textbf{32.20} \\
        Replace or add audio & \underline{0.00} & \textbf{36.96} \\
        Mean recall across six attack types & \underline{3.46} & \textbf{47.04} \\
        Fake recall & \underline{3.69} & \textbf{47.73} \\
        Overall accuracy & \underline{50.43} & \textbf{57.10} \\
        \bottomrule
    \end{tabular}
\end{table*}

\begin{table*}[t]
    \centering
    \small
    \setlength{\tabcolsep}{7pt}
    \caption{\textbf{FaceForensics++ Cross-dataset Detection Results Without Fine-tuning.} All values are percentages.}
    \label{tab:cross_dataset_ffpp}
    \begin{tabular}{lcc}
        \toprule
        Metric & AIDE & ManiVidLens \\
        \midrule
        Deepfakes recall & \underline{15.71} & \textbf{83.57} \\
        Face2Face recall & \underline{18.57} & \textbf{75.00} \\
        FaceSwap recall & \underline{12.86} & \textbf{75.00} \\
        NeuralTextures recall & \underline{14.29} & \textbf{77.86} \\
        Real recall & \textbf{92.14} & \underline{41.43} \\
        Overall accuracy & \underline{30.71} & \textbf{70.57} \\
        Binary balanced accuracy & \underline{53.75} & \textbf{59.64} \\
        \bottomrule
    \end{tabular}
\end{table*}

\FloatBarrier
\subsubsection{\texorpdfstring{Out-of-distribution~(OOD) Forgery Detection}{Out-of-distribution~(OOD) Forgery Detection}}
\label{app:cross_dataset_detection}
To assess transfer beyond ManiVidBench, we evaluate the same ManiVidLens checkpoint on VideoSham~\citep{mittal2023video} and FaceForensics++~\citep{rossler2019faceforensics++} without target-dataset fine-tuning as our OOD experiments. We compare it with the frozen AIDE encoder evaluated on the same test videos.

On VideoSham, we evaluate all 352 real--fake pairs (704 videos) in the public release, covering all six attack types. Table~\ref{tab:cross_dataset_videosham} reports video-level percentages; attack recalls are computed over manipulated videos of the corresponding type.

AIDE identifies 97.16\% of real videos but detects only 3.69\% of manipulated videos. ManiVidLens raises fake recall to 47.73\% while retaining 66.48\% real recall, narrowing the real--fake recall gap from 93.47 to 18.75 percentage points and improving overall accuracy from 50.43\% to 57.10\%. Both methods receive RGB input only; therefore, the audio-edit row reflects visual predictions on videos whose audio was manipulated.

For FaceForensics++, we use the official c23 (HQ) test split with 140 real videos and 140 videos from each of its four forgery categories (700 videos total). Because the test set contains four times as many fake as real videos, we report binary balanced accuracy in addition to overall accuracy.

ManiVidLens reaches 70.57\% overall accuracy and 59.64\% binary balanced accuracy, compared with 30.71\% and 53.75\% for AIDE. Its fake recall ranges from 75.00\% to 83.57\% across the four forgery categories, whereas AIDE ranges from 12.86\% to 18.57\%. The lower real recall of ManiVidLens (41.43\% vs. 92.14\%) shows that its higher overall accuracy should be interpreted together with the class-balanced result. These experiments provide evidence of transfer for binary forgery detection; neither dataset supplies the region masks and anomaly explanations needed to test external grounding or explanation.

\FloatBarrier
\subsubsection{Robustness Study}
\label{app:robustness}
\begin{table}[!htbp]
    \centering
    \vspace{-2mm}
    \caption{\textbf{Robustness Evaluation under Different Video Degradations.}
    Grounding, explanation, and detection scores ($\uparrow$) are evaluated under four types of video degradation.
    Parenthesized values show signed percentage changes relative to the undegraded ManiVidLens baseline, with negative values indicating performance degradation for the corresponding metric.}
    \vspace{1mm}
    \setlength{\tabcolsep}{3pt}
    \renewcommand{\arraystretch}{1.3}
    \scriptsize
    \belowrulesep=-0.25pt
    \aboverulesep=-0.25pt
    
    \resizebox{\linewidth}{!}{
        \begin{tabular}{lc|cc|cc|cc}
        \toprule[1pt]
        \multicolumn{2}{l|}{\multirow{2}{*}{\textbf{Setting}}}
        & \multicolumn{2}{c|}{\textbf{Grounding}}
        & \multicolumn{2}{c|}{\textbf{Explanation}}
        & \multicolumn{2}{c}{\textbf{Detection}} \\
    
        \cmidrule(l{3pt}r{3pt}){3-4}
        \cmidrule(l{3pt}r{3pt}){5-6}
        \cmidrule(l{3pt}r{3pt}){7-8}
    
        \multicolumn{2}{l|}{}
        & mIoU & J\&F & R-L & CSS & Acc & F1 \\
        \hline
    
        \multicolumn{2}{l|}{ManiVidLens}
        & 0.407 & 0.422 & 0.593 & 0.796 & 0.920 & 0.919 \\
    
        \hline
    
        \multirow{2}{*}{H.264 Compression}
        & CRF 23
        & 0.384\,{\tiny\textbf{(-5.7\%)}} & 0.400\,{\tiny\textbf{(-5.2\%)}}
    & 0.562\,{\tiny\textbf{(-5.2\%)}} & 0.735\,{\tiny\textbf{(-7.7\%)}}
    & 0.882\,{\tiny\textbf{(-4.1\%)}} & 0.882\,{\tiny\textbf{(-4.0\%)}} \\
        
        & CRF 40
        & 0.246\,{\tiny\textbf{(-39.6\%)}} & 0.255\,{\tiny\textbf{(-39.6\%)}}
    & 0.514\,{\tiny\textbf{(-13.3\%)}} & 0.638\,{\tiny\textbf{(-19.8\%)}}
    & 0.591\,{\tiny\textbf{(-35.8\%)}} & 0.535\,{\tiny\textbf{(-41.8\%)}} \\
    
        \hline
    
        \multirow{2}{*}{Downsampling}
        & $\times$ 0.75
        & 0.399\,{\tiny\textbf{(-2.0\%)}} & 0.416\,{\tiny\textbf{(-1.4\%)}}
    & 0.577\,{\tiny\textbf{(-2.7\%)}} & 0.775\,{\tiny\textbf{(-2.6\%)}}
    & 0.895\,{\tiny\textbf{(-2.7\%)}} & 0.895\,{\tiny\textbf{(-2.6\%)}} \\
        
        & $\times$ 0.50
        & 0.400\,{\tiny\textbf{(-1.7\%)}} & 0.415\,{\tiny\textbf{(-1.7\%)}}
    & 0.567\,{\tiny\textbf{(-4.4\%)}} & 0.764\,{\tiny\textbf{(-4.0\%)}}
    & 0.859\,{\tiny\textbf{(-6.6\%)}} & 0.859\,{\tiny\textbf{(-6.5\%)}} \\
    
        \hline
    
        \multirow{2}{*}{FPS Reduction}
        & 16 FPS
        & 0.387\,{\tiny\textbf{(-4.9\%)}} & 0.401\,{\tiny\textbf{(-5.0\%)}}
    & 0.575\,{\tiny\textbf{(-3.0\%)}} & 0.761\,{\tiny\textbf{(-4.4\%)}}
    & 0.897\,{\tiny\textbf{(-2.5\%)}} & 0.896\,{\tiny\textbf{(-2.5\%)}} \\
        
        & 8 FPS
        & 0.393\,{\tiny\textbf{(-3.4\%)}} & 0.407\,{\tiny\textbf{(-3.6\%)}}
    & 0.574\,{\tiny\textbf{(-3.2\%)}} & 0.778\,{\tiny\textbf{(-2.3\%)}}
    & 0.906\,{\tiny\textbf{(-1.5\%)}} & 0.906\,{\tiny\textbf{(-1.4\%)}} \\
    
        \hline
    
        \multirow{2}{*}{Random Frame Drop}
        & 10\%
        & 0.380\,{\tiny\textbf{(-6.6\%)}} & 0.395\,{\tiny\textbf{(-6.4\%)}}
    & 0.574\,{\tiny\textbf{(-3.2\%)}} & 0.760\,{\tiny\textbf{(-4.5\%)}}
    & 0.900\,{\tiny\textbf{(-2.2\%)}} & 0.899\,{\tiny\textbf{(-2.2\%)}} \\
        
        & 30\%
        & 0.383\,{\tiny\textbf{(-5.9\%)}} & 0.398\,{\tiny\textbf{(-5.7\%)}}
    & 0.573\,{\tiny\textbf{(-3.4\%)}} & 0.758\,{\tiny\textbf{(-4.8\%)}}
    & 0.890\,{\tiny\textbf{(-3.3\%)}} & 0.889\,{\tiny\textbf{(-3.3\%)}} \\
    
        \bottomrule[1pt]
        \end{tabular}
    }
    \vspace{-1mm}
    \label{tab:ablation_robustness}
\end{table}

\noindent\textbf{Evaluation Protocol.}
We apply each degradation independently to the original test videos using the same trained checkpoint, prompts, frame-sampling policy, and evaluation implementation.
The settings comprise H.264 compression at CRF~23 and~40, spatial downsampling by factors of $0.75$ and $0.50$, frame-rate reduction to 16 and 8\,FPS, and random removal of $10\%$ and $30\%$ of frames.
Grounding is evaluated on every decoded frame of each degraded video: source-frame correspondences select the original masks, and predictions are mapped back to the original ground-truth coordinate system and resolution.
Explanation and detection retain the original text references and authenticity labels, respectively, with metric aggregation following Appendix~\ref{app:evaluation_metrics}.
For baseline and degraded scores $s_{\mathrm{base}}$ and $s_{\mathrm{degraded}}$, the signed relative change is
\begin{equation}
    \Delta_s = \frac{s_{\mathrm{degraded}}-s_{\mathrm{base}}}{s_{\mathrm{base}}}\times100\%,
    \label{eq:robustness_relative_change}
\end{equation}
where negative values indicate a decrease from the corresponding baseline score.

\noindent\textbf{Performance under Degradation.}
Severe H.264 compression produces the largest losses across all six metrics (Table~\ref{tab:ablation_robustness}).
Increasing compression from CRF~23 to CRF~40 reduces mIoU from 0.384 to 0.246, detection F1 from 0.882 to 0.535, and CSS from 0.735 to 0.638.
Spatial downsampling preserves grounding more strongly: at half resolution, mIoU changes \textbf{from 0.407 to 0.400} and J\&F \textbf{from 0.422 to 0.415}, while detection F1 decreases from 0.919 to 0.859.
Temporal degradations produce smaller changes than severe compression.
At 8\,FPS, ManiVidLens retains 0.393 mIoU, 0.778 CSS, and 0.906 detection F1.
Randomly removing $30\%$ of frames reduces mIoU from 0.407 to 0.383 and detection F1 from 0.919 to 0.889.
Grounding is non-monotonic across the temporal severity levels: 8\,FPS yields slightly higher mIoU than 16\,FPS, as does $30\%$ frame removal compared with $10\%$.
Together, these results show stronger sensitivity to severe compression than to spatial resizing or temporal frame loss.

\FloatBarrier

\subsection{ManiVidLens Case Visualization}
\begin{figure*}[p]
    \centering
    \includegraphics[width=\textwidth,height=0.93\textheight,keepaspectratio]{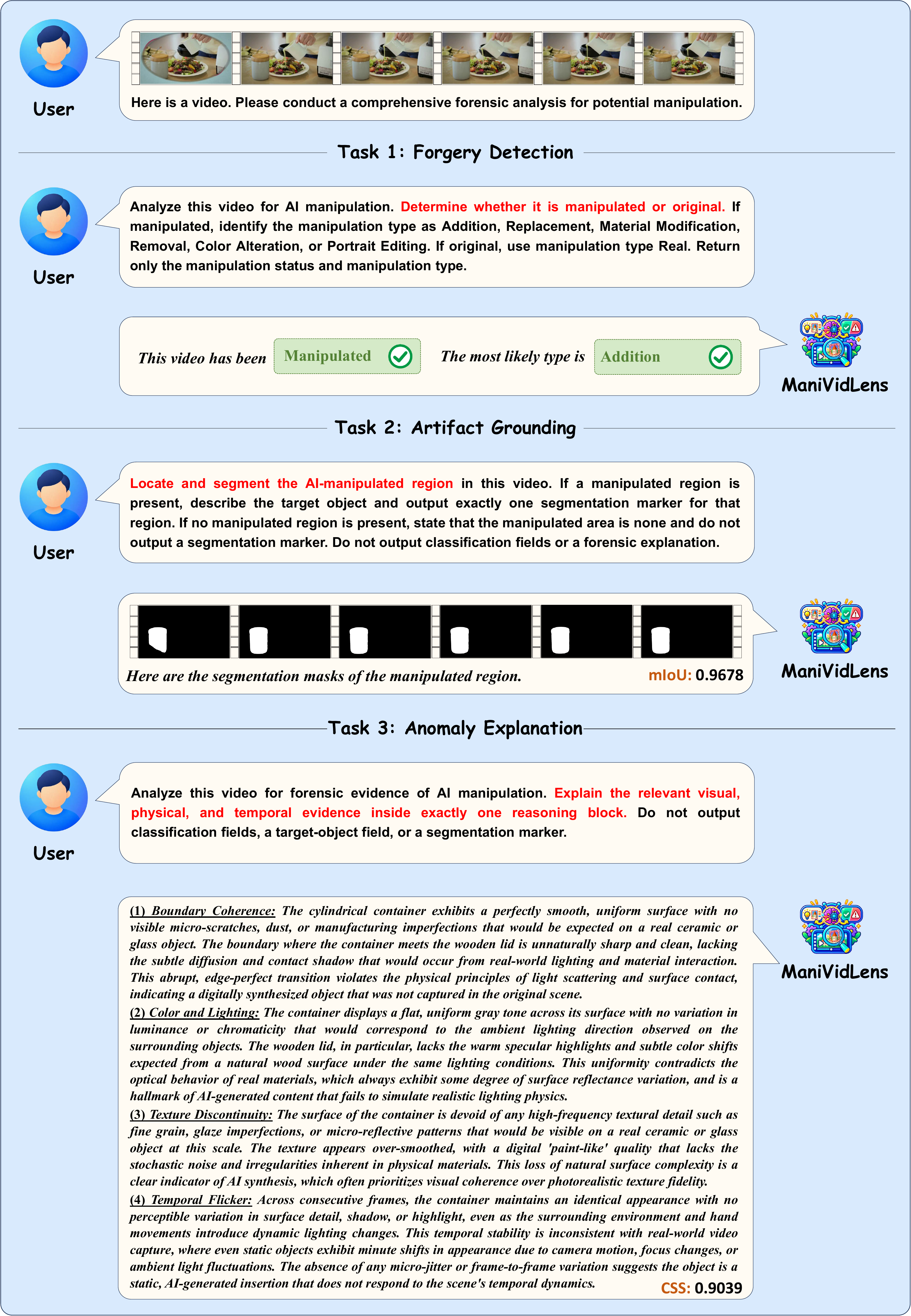}
    \vspace{1mm}
    \caption{\textbf{A ManiVidLens Case on an \textit{Addition} Sample.}}\label{fig:addition_case_1}
\end{figure*}

\begin{figure*}[p]
    \centering
    \includegraphics[width=\textwidth,height=0.93\textheight,keepaspectratio]{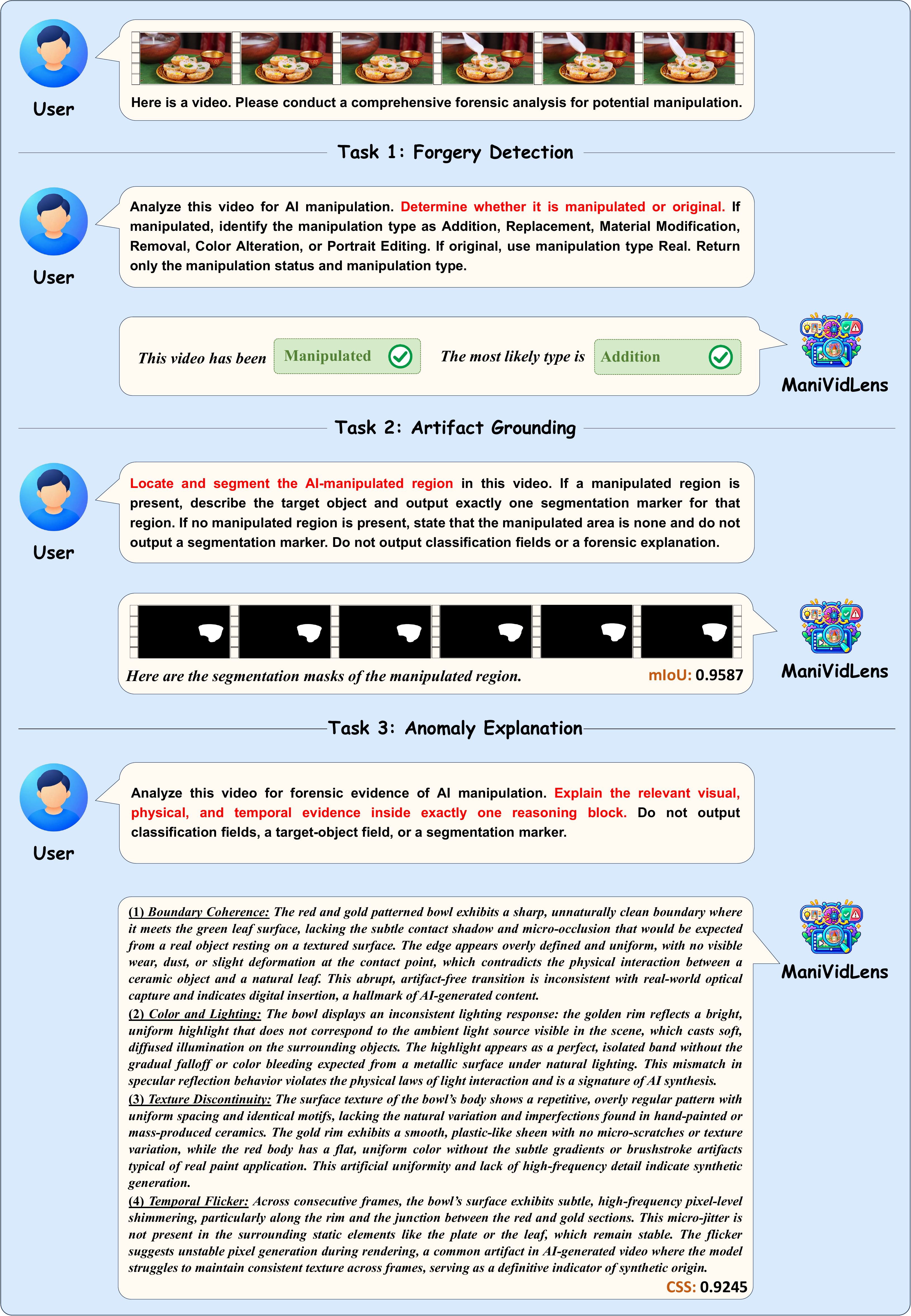}
    \vspace{1mm}
    \caption{\textbf{A ManiVidLens Case on an \textit{Addition} Sample.}}\label{fig:addition_case_2}
\end{figure*}

\begin{figure*}[p]
    \centering
    \includegraphics[width=\textwidth,height=0.93\textheight,keepaspectratio]{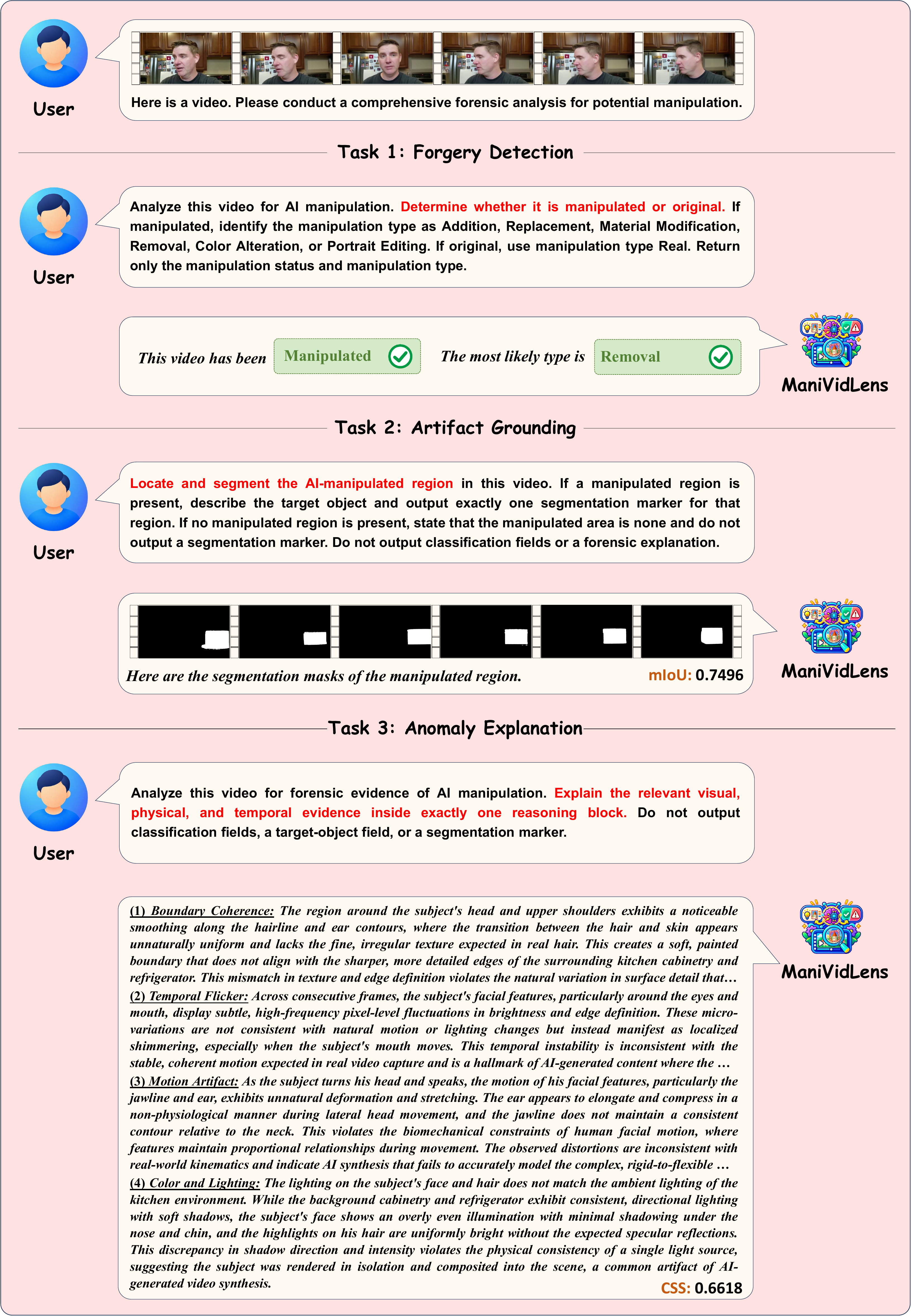}
    \vspace{1mm}
    \caption{\textbf{A ManiVidLens Case on a \textit{Removal} Sample.}}\label{fig:removal_case_1}
\end{figure*}

\begin{figure*}[p]
    \centering
    \includegraphics[width=\textwidth,height=0.93\textheight,keepaspectratio]{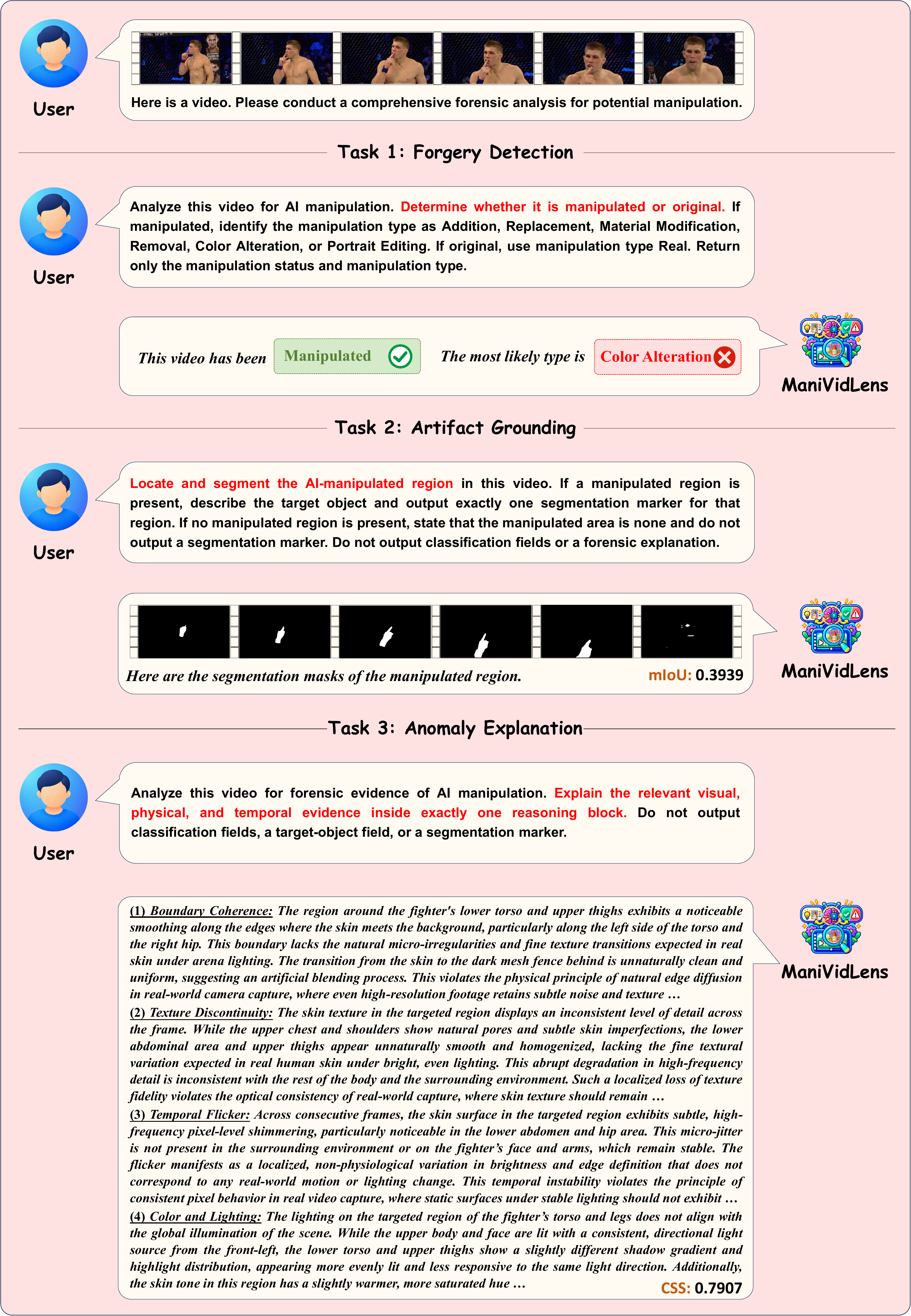}
    \vspace{1mm}
    \caption{\textbf{A ManiVidLens Case on a \textit{Removal} Sample.}}\label{fig:removal_case_2}
\end{figure*}

\begin{figure*}[p]
    \centering
    \includegraphics[width=\textwidth,height=0.93\textheight,keepaspectratio]{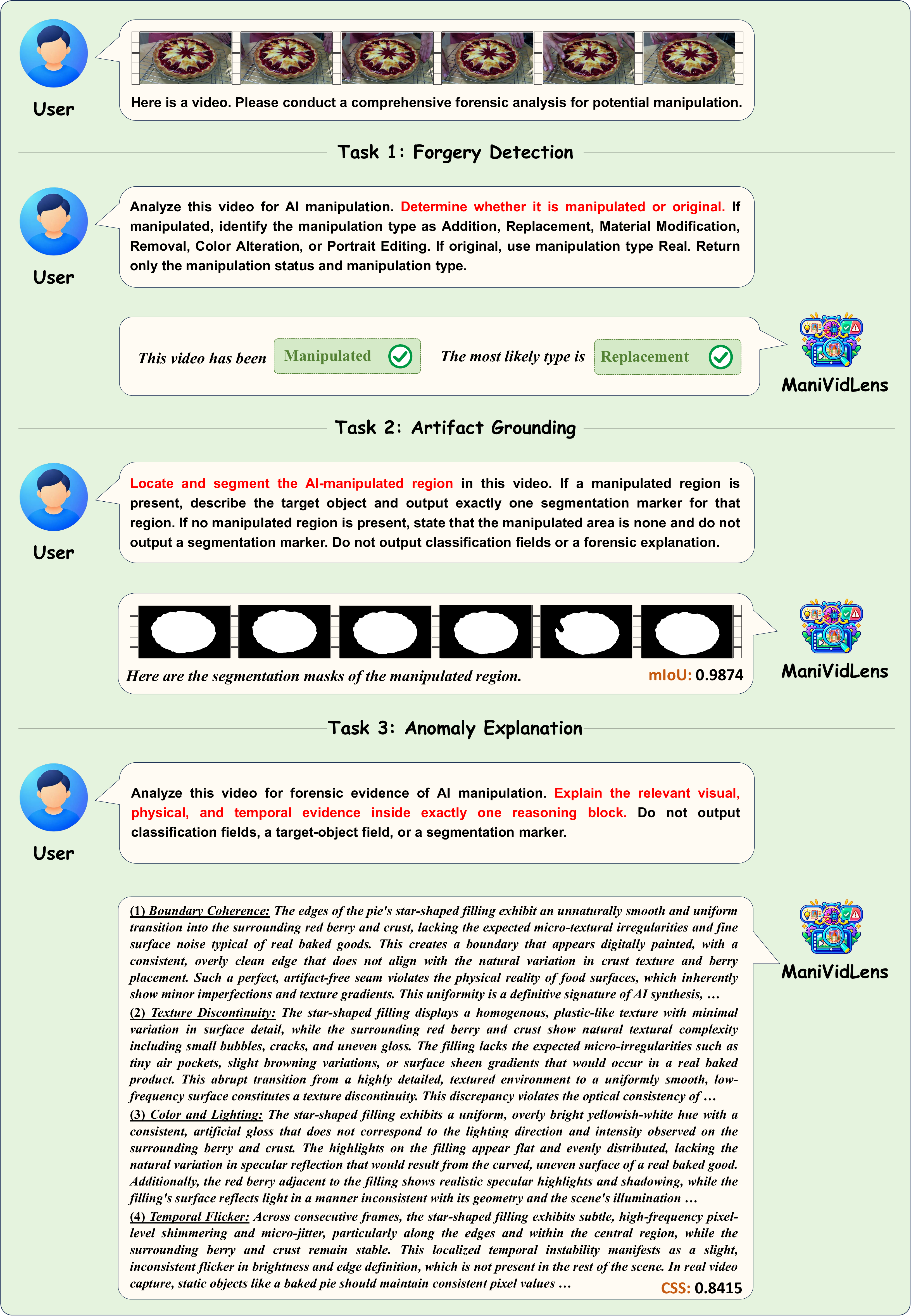}
    \vspace{1mm}
    \caption{\textbf{A ManiVidLens Case on a \textit{Replacement} Sample.}}\label{fig:replacement_case_1}
\end{figure*}

\begin{figure*}[p]
    \centering
    \includegraphics[width=\textwidth,height=0.93\textheight,keepaspectratio]{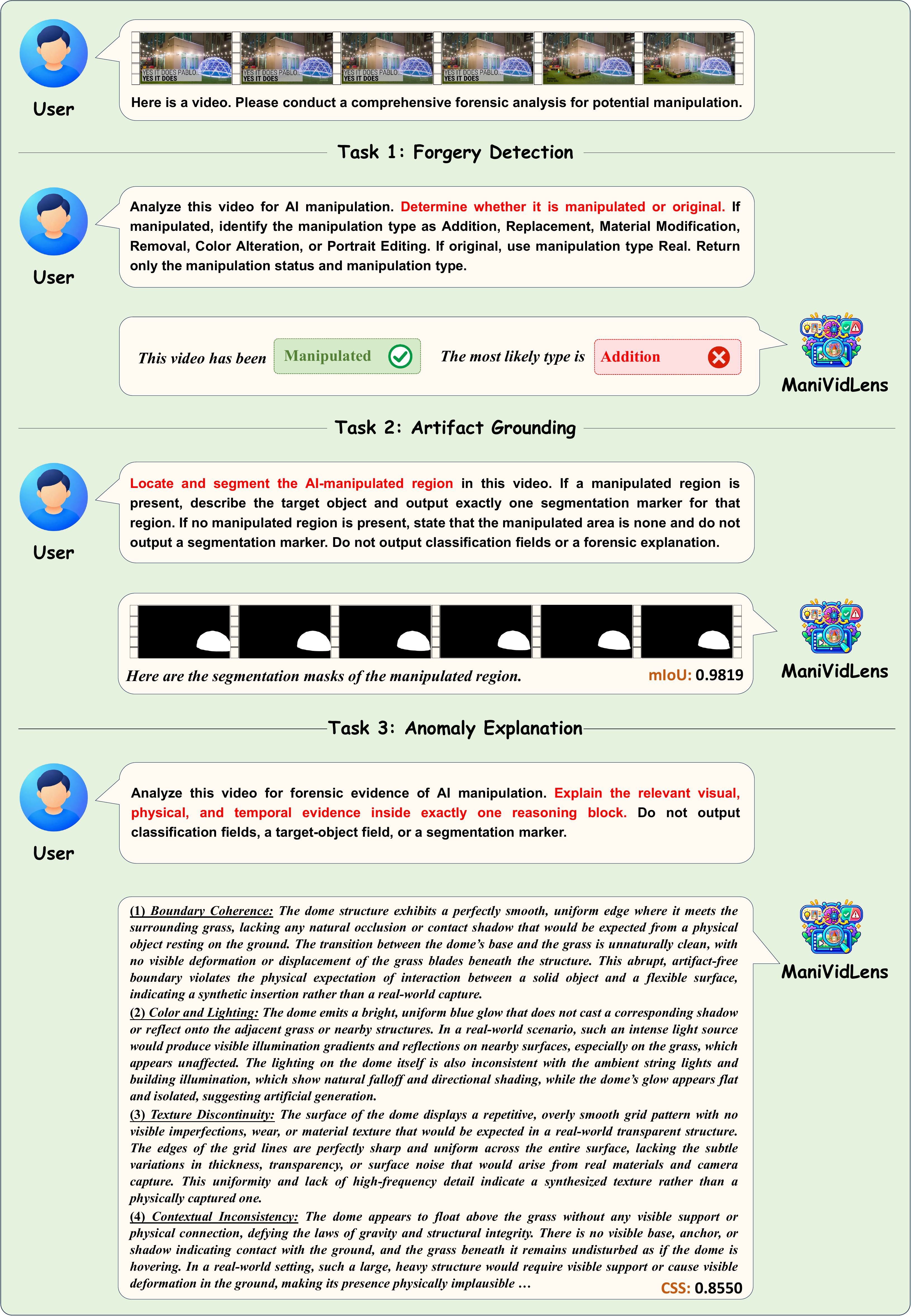}
    \vspace{1mm}
    \caption{\textbf{A ManiVidLens Case on a \textit{Replacement} Sample.}}\label{fig:replacement_case_2}
\end{figure*}

\begin{figure*}[p]
    \centering
    \includegraphics[width=\textwidth,height=0.93\textheight,keepaspectratio]{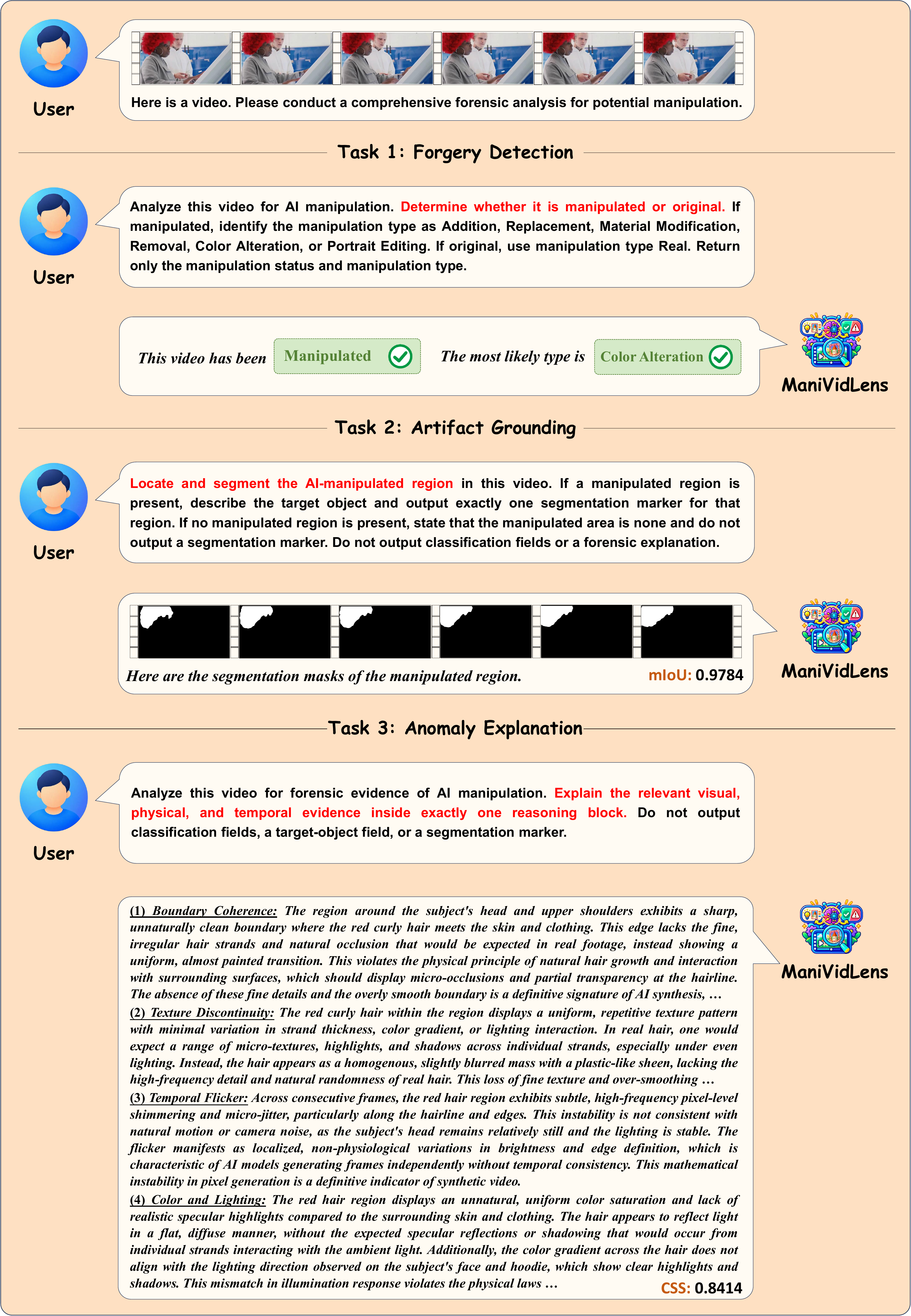}
    \vspace{1mm}
    \caption{\textbf{A ManiVidLens Case on a \textit{Color Alteration} Sample.}}\label{fig:color_case_1}
\end{figure*}

\begin{figure*}[p]
    \centering
    \includegraphics[width=\textwidth,height=0.93\textheight,keepaspectratio]{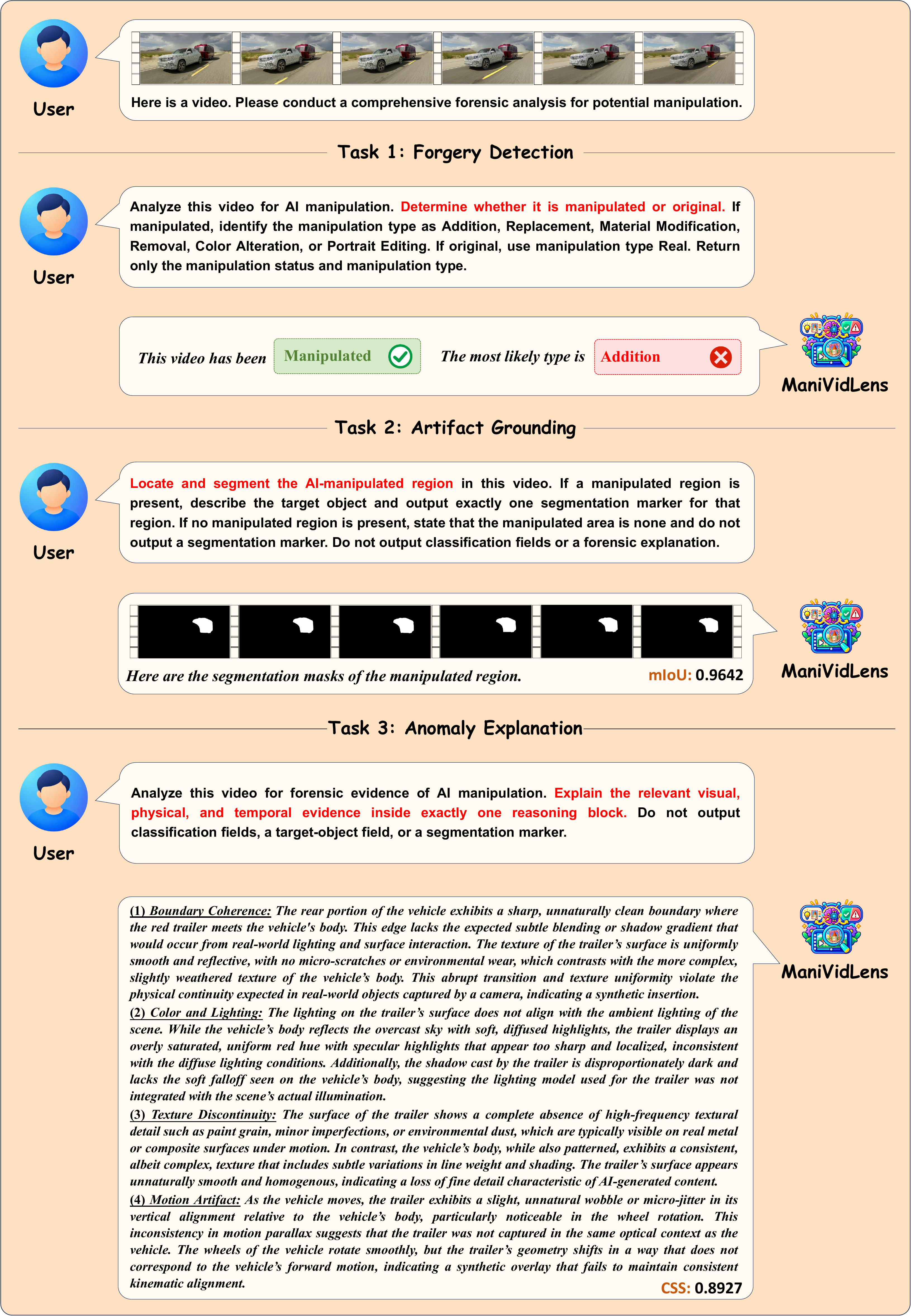}
    \vspace{1mm}
    \caption{\textbf{A ManiVidLens Case on a \textit{Color Alteration} Sample.}}\label{fig:color_case_2}
\end{figure*}

\begin{figure*}[p]
    \centering
    \includegraphics[width=\textwidth,height=0.93\textheight,keepaspectratio]{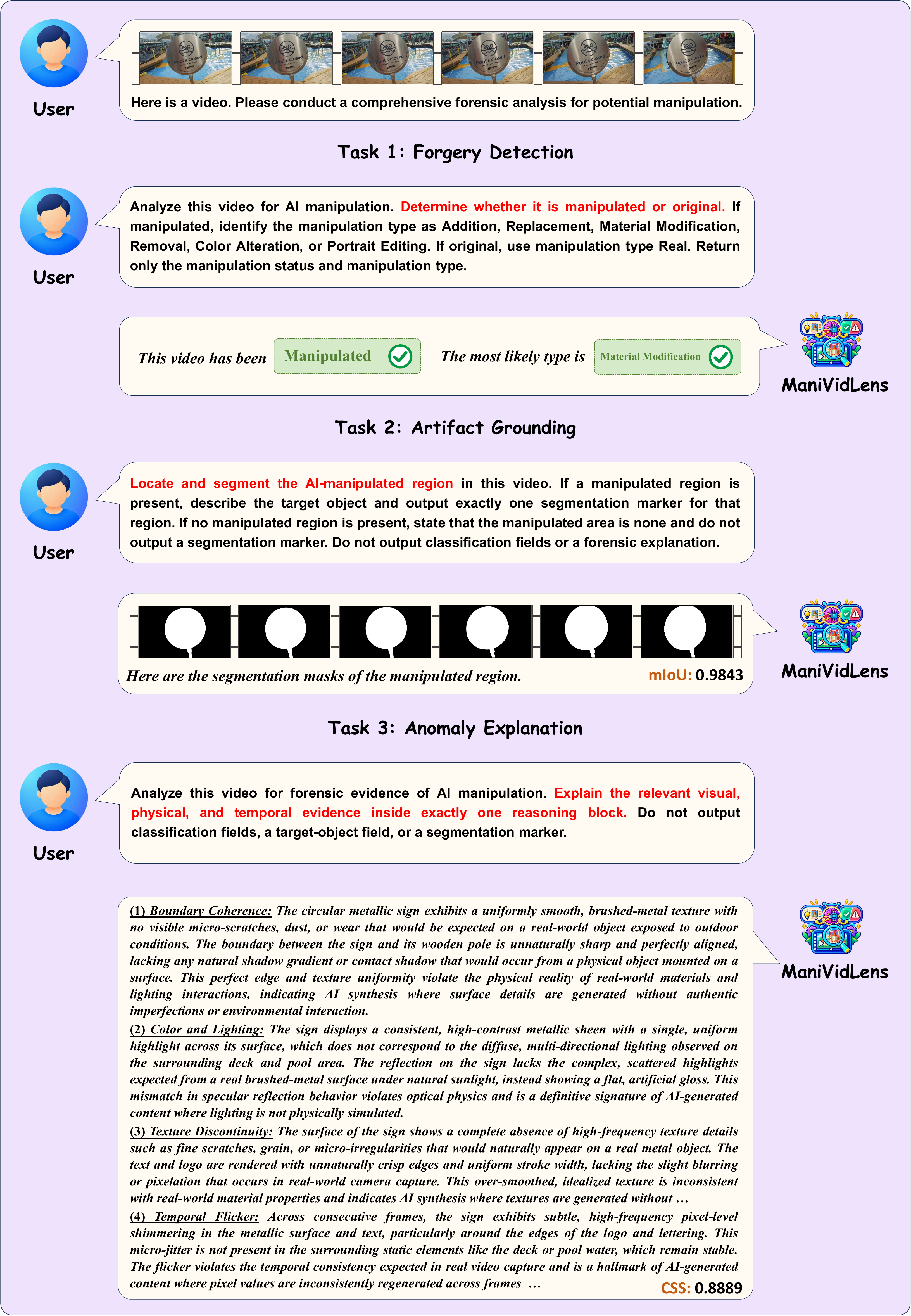}
    \vspace{1mm}
    \caption{\textbf{A ManiVidLens Case on a \textit{Material Modification} Sample.}}\label{fig:material_case_1}
\end{figure*}

\begin{figure*}[p]
    \centering
    \includegraphics[width=\textwidth,height=0.93\textheight,keepaspectratio]{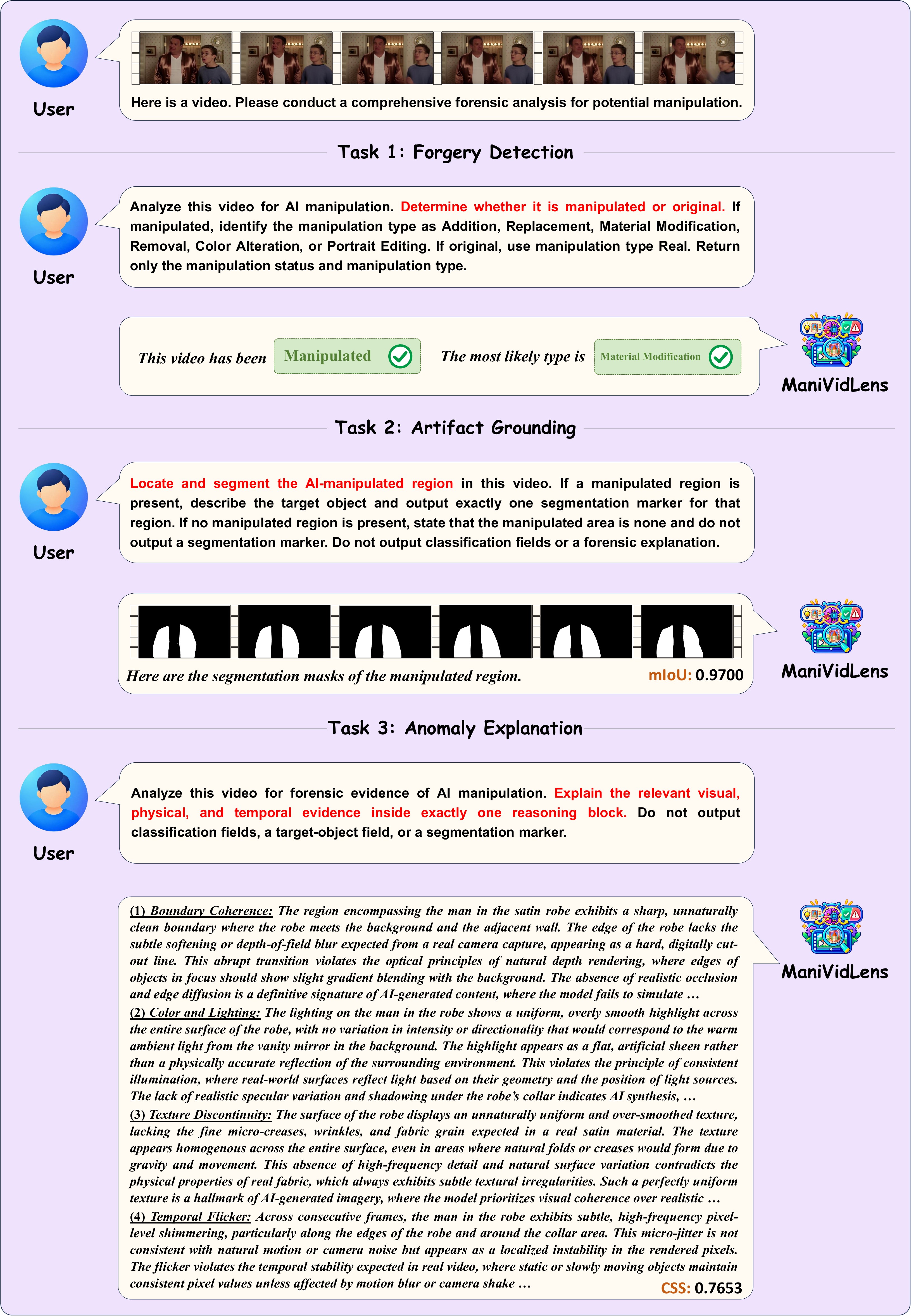}
    \vspace{1mm}
    \caption{\textbf{A ManiVidLens Case on a \textit{Material Modification} Sample.}}\label{fig:material_case_2}
\end{figure*}

\begin{figure*}[p]
    \centering
    \includegraphics[width=\textwidth,height=0.93\textheight,keepaspectratio]{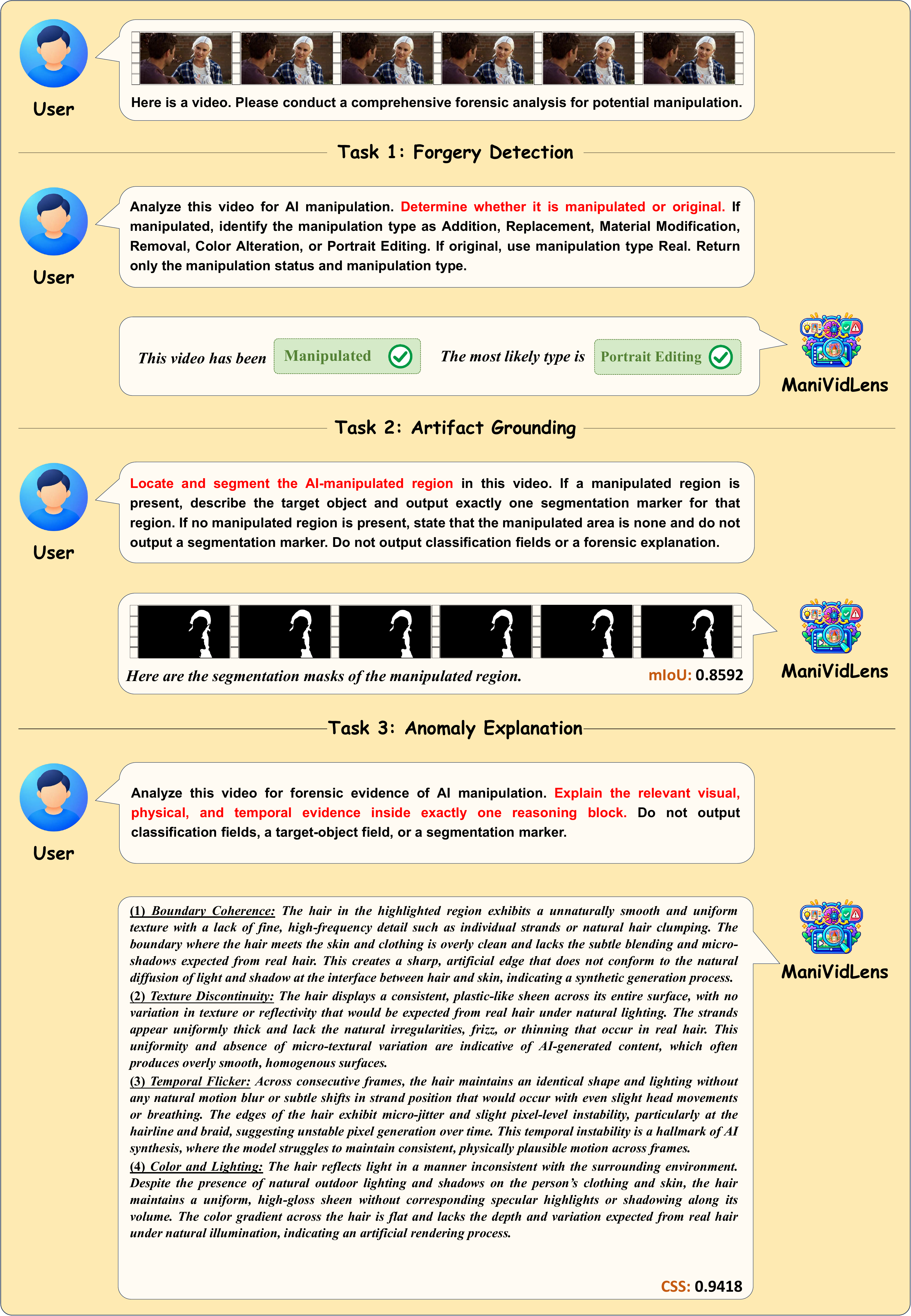}
    \vspace{1mm}
    \caption{\textbf{A ManiVidLens Case on a \textit{Portrait Editing} Sample.}}\label{fig:portrait_case_1}
\end{figure*}

\begin{figure*}[p]
    \centering
    \includegraphics[width=\textwidth,height=0.93\textheight,keepaspectratio]{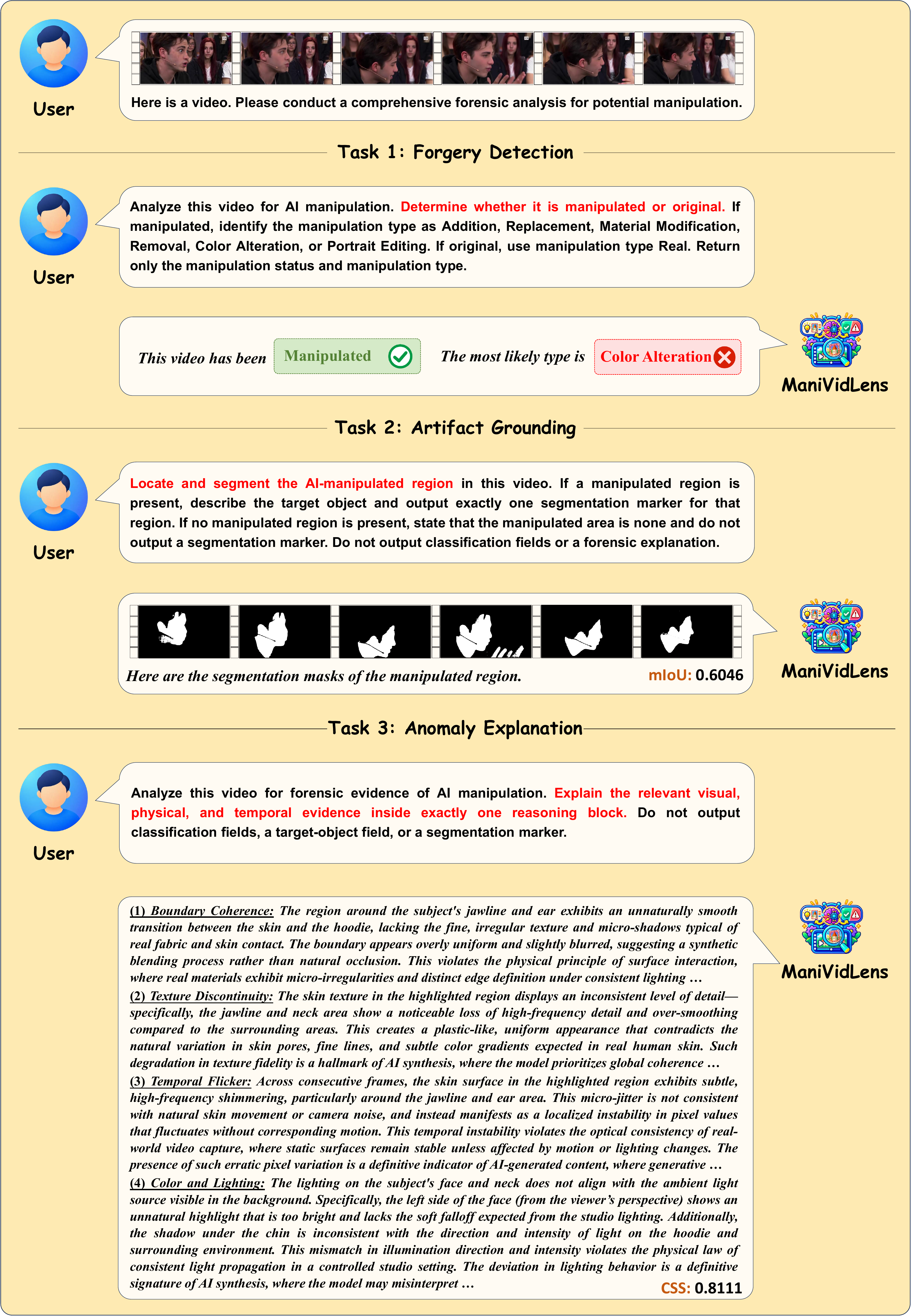}
    \vspace{1mm}
    \caption{\textbf{A ManiVidLens Case on a \textit{Portrait Editing} Sample.}}\label{fig:portrait_case_2}
\end{figure*}

Fig.~\ref{fig:addition_case_1}, ~\ref{fig:addition_case_2}~(\textbf{\textit{Addition}}), ~\ref{fig:removal_case_1}, ~\ref{fig:removal_case_2}~(\textbf{\textit{Removal}}), ~\ref{fig:replacement_case_1}, ~\ref{fig:replacement_case_2}~(\textbf{\textit{Replacement}}), ~\ref{fig:color_case_1}, ~\ref{fig:color_case_2}~(\textbf{\textit{Color Alteration}}),~\ref{fig:material_case_1}, ~\ref{fig:material_case_2}~(\textbf{\textit{Material Modification}}), ~\ref{fig:portrait_case_1}, ~\ref{fig:portrait_case_2}~(\textbf{\textit{Portrait Editing}}) present multiple ManiVidLens cases across different manipulation samples.

\end{document}